# Learning with the Nash-Sutcliffe loss

Hristos Tyralis[1*], Georgia Papacharalampous[2]

[1]Support Command, Hellenic Air Force, Elefsina Air Base, 19 200, Elefsina, Greece (montchrister@gmail.com, hristos@itia.ntua.gr, https://orcid.org/0000-0002-8932-4997)

[2]Department of Land, Environment, Agriculture and Forestry, University of Padova, Viale dell'Università 16, 35020, Legnaro, Italy (papacharalampous.georgia@gmail.com, georgia.papacharalampous@unipd.it, https://orcid.org/0000-0001-5446-954X)

*Corresponding author

**Abstract**: The Nash-Sutcliffe efficiency (NSE) is a widely used, positively oriented relative measure for evaluating forecasts across multiple time series. However, it lacks a decision-theoretic foundation for this purpose. To address this, we examine its negatively oriented counterpart, which we refer to as Nash-Sutcliffe loss, defined as $L_{\mathrm{NS}} = 1 - \mathrm{NSE}$. We prove that $L_{\mathrm{NS}}$ is strictly consistent for an elicitable and identifiable multi-dimensional functional, which we name the Nash-Sutcliffe functional. This functional is a data-weighted component-wise mean. The common practice of maximizing the average NSE across multiple series is the sample analog of minimizing the expected $L_{\mathrm{NS}}$. Consequently, this operation implicitly assumes that all series originate from a single non-stationary, stochastic process. We introduce Nash-Sutcliffe linear regression, a multi-dimensional model estimated by minimizing the average $L_{\mathrm{NS}}$, which reduces to a data-weighted least squares formulation. By reorienting the sample average loss function, we extend the previously proposed evaluation and estimation framework to forecasting multiple stationary dependent time series with differing stochastic properties. This constitutes a more natural empirical implementation of the NSE than the earlier formulation. Our results establish a decision-theoretic foundation for NSE-based model estimation and forecast evaluation in large datasets, while further clarifying the benefits of global over local machine learning models.

## 1. Introduction

A primary goal in machine learning, statistical modeling and forecasting is to make accurate predictions. The variables we aim to predict are random and often continuous; consequently, regression models are designed to predict statistical functionals (properties) of the variables' probability distributions, such as the mean or median (Dimitriadis et al. 2024a; Gneiting 2011). Models that target such functionals without specifying the distribution are called semiparametric (Dimitriadis et al. 2024a). They assume only that the distribution belongs to a broad family, making them more flexible than fully parametric models. In these models, the procedures for estimation and prediction evaluation are intrinsically linked. The functional we target dictates the loss function used for both estimation and evaluation (Dimitriadis et al. 2024a).

## 1.1 A practical example: Predicting the conditional mean

Consider a one-dimensional dependent random variable $\underline{y}$ and a one-dimensional predictor random variable $\underline{x}$. Assume a correctly specified semiparametric model $M$ predicts the conditional expectation $\mathbb{E}_F[\underline{y}|\underline{x}]$, such that $\mathbb{E}_F[\underline{y}|\underline{x}] = M(\underline{x}|\theta_0)$ for a unique, unknown parameter $\theta_0$.

Since $\theta_0$ is unknown, we write the prediction for a given $x$ and parameter $\theta$ as $z = M(x|\theta)$. To ensure the model predicts the conditional mean, the parameter $\theta$ must be estimated by minimizing the Mean Squared Error (MSE) between the prediction vector $\boldsymbol{z}_n = (z_1, \dots, z_n)^{\mathrm{T}} = (M(x_1|\theta), \dots, M(x_n|\theta))^{\mathrm{T}}$ and the observed vector $\boldsymbol{y}_n = (y_1, \dots, y_n)^{\mathrm{T}}$ (Dimitriadis et al. 2024a). In what follows, vectors are treated as column vectors. Formal notational definitions are detailed in Appendix A, eq. (A.1). The MSE is defined as:

$$\mathrm{MSE}(\boldsymbol{z}_n, \boldsymbol{y}_n) := (1/n) \sum_{i=1}^{n} L_{\mathrm{SE}}(z_i, y_i) \tag{1.1}$$

where $L_{\mathrm{SE}}$ is the squared error loss function:

$$L_{\mathrm{SE}}(z, y) := (z - y)^2 \tag{1.2}$$

The estimator $\underset{\theta \in \Theta}{\arg\min}\, \mathrm{MSE}(\boldsymbol{M}(\underline{\boldsymbol{x}}_n|\theta), \underline{\boldsymbol{y}}_n)$, where $\boldsymbol{M}(\boldsymbol{x}_n|\theta) = (M(x_1|\theta), \dots, M(x_n|\theta))^{\mathrm{T}}$, is a consistent estimator of $\theta_0$ (Dimitriadis et al. 2024a). This approach does not require knowledge of the exact conditional cumulative distribution function (CDF) $F_{\underline{y}|\underline{x}}$. It only assumes this CDF belongs to a family $\mathcal{F}$ for which the conditional expectation exists and is finite.

## 1.2 Model selection, strictly consistent loss functions and realized losses

In practice, the true data-generating process is unknown, and the semiparametric models we use may be misspecified. This necessitates comparing multiple models to select the most accurate one (Patton 2020).

When the goal is to predict the conditional mean, a natural approach is to compare models using strictly consistent loss functions for the mean. A loss function is strictly consistent for a functional if its expected value is uniquely minimized by the true value of that functional (Gneiting 2011). This property ensures that, on average, the model is incentivized to report the true functional value. For example, the squared error loss $L_{\text{SE}}$ is strictly consistent for the mean, making it the natural choice for evaluating mean predictions. In practice, the realized (average) loss (e.g. MSE) is computed on a test set and the model with the lowest realized loss is selected (Gneiting 2011). A functional that uniquely minimizes a strictly consistent loss function is called elicitable (Gneiting 2011).

It is important to distinguish between the consistency of statistical estimators and the strict consistency of loss functions; they refer to different concepts. However, strictly consistent loss functions do lead to consistent $M$-estimators, connecting the two procedures (Dimitriadis et al. 2024a).

## 1.3 The Nash-Sutcliffe efficiency (NSE) and the realized NSE ($\overline{\text{NSE}}$)

In forecasting, practitioners often work with a single time series. One widely adopted metric in this setting is the Nash-Sutcliffe efficiency (NSE), a positively oriented score (higher values are better) defined as a transformation of the MSE (Nash and Sutcliffe 1970):

$$\text{NSE}(\boldsymbol{z}_d, \boldsymbol{y}_d) := 1 - \text{MSE}(\boldsymbol{z}_d, \boldsymbol{y}_d)/\text{MSE}(\boldsymbol{1}_d \mu(\boldsymbol{y}_d), \boldsymbol{y}_d), d \geq 2 \tag{1.3}$$

Here $\boldsymbol{1}_d$ is a $d$-dimensional vector of ones (see eq. (A.3)) and $\mu(\boldsymbol{y}_d)$ represents the sample mean of $\boldsymbol{y}_d$ (see eq. (A.30)). The constraint $d \geq 2$ is necessary because for $d = 1$, the denominator MSE becomes 0, rendering the NSE undefined. The NSE has achieved broad adoption in geosciences and environmental sciences (especially in hydrology), where it often functions as a core measure of predictive performance (Melsen et al. 2025). Its application has since expanded to a variety of other disciplines, including statistical forecasting, machine learning and statistical learning.

NSE values range from $-\infty$ to 1. A value greater than 0 indicates the model's predictions are more accurate than a naïve benchmark that predicts the mean of the observation ($\mu(\boldsymbol{y}_d)$). As a skill score, it quantifies predictive skill relative to this benchmark (Gneiting and Resin 2023). For a single time series, NSE retains the model rankings of MSE but is often preferred for its interpretability.

With the increasing availability of larger datasets, it became common to test models across multiple time series. NSE has been widely adopted for such multi-series comparisons series (Perrin et al. 2001), largely due to its homogeneity of degree zero:

$$\mathrm{NSE}(c\boldsymbol{z}_d, c\boldsymbol{y}_d) = \mathrm{NSE}(\boldsymbol{z}_d, \boldsymbol{y}_d) \forall c \in \mathbb{R}\backslash\{0\} \quad (1.4)$$

This property was assumed to facilitate comparisons across datasets with different scales. Hyndman and Koehler (2006) refer to skill scores of this type (when expressed in negative orientation) as relative measures and discuss the interpretation of their values, much as we described earlier for the NSE.

Consider $n$ time series, each of length $d$, represented by a $d \times n$ matrix $\boldsymbol{Y}_{d\times n} = [y_{ij}]$ (for matrix notation see eq. (A.5)), where each column $\boldsymbol{Y}_{\cdot,j} = (y_{1j}, \dots, y_{dj})^{\mathrm{T}}$ represents the $j^{\text{th}}$ series (for notation of rows and columns of matrices see eqs. (A.6) and (A.7)). The model predictions are in a separate $d \times n$ matrix $\boldsymbol{Z}_{d\times n} = [z_{ij}]$. To rank predictions across all series, the NSE is calculated for each series and then averaged across all series (Perrin et al. 2001):

$$\overline{\mathrm{NSE}}(\boldsymbol{Z}_{d\times n}, \boldsymbol{Y}_{d\times n}) = (1/n)\sum_{j=1}^{n} \mathrm{NSE}(\boldsymbol{Z}_{\cdot,j}, \boldsymbol{Y}_{\cdot,j}), d \geq 2 \quad (1.5)$$

This can be expanded as:

$$\overline{\mathrm{NSE}}(\boldsymbol{Z}_{d\times n}, \boldsymbol{Y}_{d\times n}) = 1 - (1/n)\sum_{j=1}^{n}\left(\sum_{i=1}^{d}(z_{ij} - y_{ij})^2 \,/\, \sum_{i=1}^{d}(\mu(\boldsymbol{Y}_{\cdot,j}) - y_{ij})^2\right) \quad (1.6)$$

## 1.4 Problem formulation and contributions

For our analysis, it is convenient to work with a negatively oriented version of NSE, which we refer to as Nash-Sutcliffe loss function:

$$L_{\mathrm{NS}}(\boldsymbol{z}_d, \boldsymbol{y}_d) = 1 - \mathrm{NSE}(\boldsymbol{z}_d, \boldsymbol{y}_d) = w(\boldsymbol{y}_d) L_{\mathrm{EN}}(\boldsymbol{z}_d, \boldsymbol{y}_d), d \geq 2 \quad (1.7)$$

where

$$L_{\mathrm{EN}}(\boldsymbol{z}_d, \boldsymbol{y}_d) = \|\boldsymbol{z}_d - \boldsymbol{y}_d\|_2^2 \quad (1.8)$$

(EN stands for Euclidean norm; see eq. (A.12)) and:

$$w(\boldsymbol{y}_d) = 1/L_{\mathrm{EN}}(\mu(\boldsymbol{y}_d)\boldsymbol{1}_d, \boldsymbol{y}_d) \quad (1.9)$$

The realized NSE ($\overline{\mathrm{NSE}}$) lacks a firm theoretical foundation, as the strict consistency properties of the $L_{\mathrm{NS}}$ loss have not been established. In contrast, the $L_{\mathrm{SE}}$ loss is known to be strictly consistent for the mean. Specifically, the expectation $\mathbb{E}_F[L_{\mathrm{SE}}(z, \underline{y})]$, which is a function of $z$, after integrating out $y$, is minimized at $z = \mathbb{E}_F[\underline{y}]$. The functional $\mathbb{E}_F[\underline{y}]$ is one-dimensional, as are the realizations of $\underline{y}$. The only assumption required to use its empirical counterpart, the MSE, is that the realizations $y_1, \ldots, y_n$ in eq. (1.1) were generated by a common distribution $F$ with a finite expectation.

The core idea here is to consider $L_{\mathrm{NS}}(\boldsymbol{z}_d, \boldsymbol{y}_d)$ as evaluating a prediction $\boldsymbol{z}_d$ of a $d$-dimensional functional against a realization $\boldsymbol{y}_d$ from the random vector $\underline{\boldsymbol{y}}_d$ with joint distribution $F$. We then investigate whether and where the expectation $\mathbb{E}_F[L_{\mathrm{NS}}(\boldsymbol{z}_d, \underline{\boldsymbol{y}}_d)]$ is minimized. Just as MSE is the empirical counterpart of $L_{\mathrm{SE}}$, the empirical counterpart of $L_{\mathrm{NS}}$ is the realized Nash-Sutcliffe loss:

$$\bar{L}_{\mathrm{NS}}(\boldsymbol{Z}_{d\times n}, \boldsymbol{Y}_{d\times n}) = (1/n) \sum_{j=1}^{n} L_{\mathrm{NS}}\left(\boldsymbol{Z}_{\cdot,j}, \boldsymbol{Y}_{\cdot,j}\right), d \geq 2 \tag{1.10}$$

This is also the negatively oriented counterpart of $\overline{\mathrm{NSE}}$, since

$$\bar{L}_{\mathrm{NS}}(\boldsymbol{Z}_{d\times n}, \boldsymbol{Y}_{d\times n}) = 1 - \overline{\mathrm{NSE}}(\boldsymbol{Z}_{d\times n}, \boldsymbol{Y}_{d\times n}) \tag{1.11}$$

The aim of this manuscript is to characterize the theoretical properties of the Nash-Sutcliffe loss function, thereby establishing a framework for interpreting empirical findings from prior studies on multi-series forecast evaluation. Our contributions are as follows:

(i) In Section 4.1, we prove that $L_{\mathrm{NS}}$ is a strictly consistent loss function that elicits a $d$-dimensional functional, which we call the Nash-Sutcliffe functional. This functional is a data-weighted component-wise mean.

(ii) In Section 4.2, we characterize the identifiability of the Nash-Sutcliffe functional. Section 4.3 demonstrates examples of $M$-estimation and Section 4.4 offers an intuitive explanation of its practical use.

(iii) The denominator of the Nash-Sutcliffe loss can become 0 on the set $\{\boldsymbol{y}_d \in \mathbb{R}^d : \|\mu(\boldsymbol{y}_d)\mathbf{1}_d - \boldsymbol{y}_d\|_2^2 = 0\}$, potentially causing the loss to be infinite. In Section 4.5, we show how to construct distributions, within known families, whose support excludes this set, confirming that the Nash-Sutcliffe loss properties hold for a rich class of distributions. Section 4.6 examines transformations of the denominator to exclude 0 values. While this allows support including the problematic set, the resulting loss functions no longer elicit the Nash-Sutcliffe functional, but rather a transformation of it.

(iv) In Section 4.7, we construct skill scores based on the Nash-Sutcliffe loss. Section 4.8 introduces Nash-Sutcliffe regression, which applies the Nash-Sutcliffe loss to linear models.

(v) The regression setup presented in Section 4.8 does not suit forecasting applications because it generates predictions across spatial replicates of existing time series. To make Nash-Sutcliffe regression applicable to forecasting, Section 5 reinterprets the results from Section 4, by shifting the framework to a temporal perspective.

(vi) Finally, Section 6 presents applications that demonstrate the practical relevance of these theoretical properties.

### 1.5 Manuscript structure

The remainder of the manuscript is structured as follows. Section 2 reviews the literature on strictly consistent loss functions in statistical science and the literature on the NSE, which has developed through advances in geosciences. Section 3 presents the necessary theoretical background. Sections 4 and 5 detail our original theoretical contributions, followed by applications in Section 6 that demonstrate their practical significance. Section 7 discusses the findings and Section 8 concludes. Appendices A-E feature supporting material on notation, distributions, proofs, a concise reference for key equations and statistical software, respectively. Supplementary information includes the data and code required to reproduce the applications in Section 6.

## 2. Literature review

We summarize the extant literature relevant to the core topics of this manuscript. First, Section 2.1 presents research on loss functions in general, specifically focusing on consistent loss functions. Section 2.2 then summarizes research on the NSE, primarily originating from the geosciences and environmental sciences.

### 2.1 Machine and statistical learning literature

Gneiting (2011) presents a comprehensive overview of strictly consistent loss (scoring) functions. Research on Bregman loss functions, which are strictly consistent for the component-wise mean of a random vector, is well-established in the literature (Banerjee et al. 2005; Gneiting 2011; Osband and Reichelstein 1985; Savage 1971). Furthermore, Frongillo and Kash (2015) and Fissler and Ziegel (2016) investigate strictly consistent loss functions for multi-dimensional functionals. Dimitriadis et al. (2024a) emphasize the intrinsic connection between model estimation and prediction evaluation, while Patton (2020) stresses the importance of pre-specifying the loss function for evaluation, thereby ensuring the modeler's estimation procedure is aligned with it.

Foundational concepts regarding loss functions, consistency and propriety are formalized by Murphy and Daan (1985). Murphy (1988) defines the NSE as a skill score and demonstrates a decomposition to facilitate a better understanding of the metric. Hyndman and Koehler (2006) review several measures of forecast accuracy; adopting their methodology, the negatively oriented variant of the NSE can be understood as a relative measure, specifically the relative MSE. For in-sample diagnostics, Gneiting and Resin (2023) establish that the NSE corresponds to the coefficient of determination, which is a special case of the universal coefficient of determination. Whether a prediction effectively targets the true functional is assessed by identification functions (Dimitriadis et al. 2024b) that measure absolute performance of forecasts, while Nolde and Ziegel (2017) examine the theoretical underpinnings of what constitutes good absolute performance.

## 2.2 Geosciences and environmental sciences literature

The NSE was originally introduced by Nash and Sutcliffe (1970) and Melsen et al. (2025) trace the history of the metric. General loss functions and error metrics used in geosciences are discussed extensively by Beven (2025), Bennett et al. (2013), Krause et al. (2005), Moriasi et al. (2007) and Jackson et al. (2019). Gupta and Kling (2009) present another detailed decomposition of the NSE, while Gupta and Kling (2011) outline the typical range of values expected when estimating hydrologic models. Clark et al. (2021) examine the uncertainties involved in evaluating the NSE.

From a statistical perspective, McCuen et al. (2006) evaluate the probability distributions of the NSE, Lamontagne et al. (2020) treat NSE variables as random variables and Duc and Sawada (2023) examine the NSE from the perspective of signal processing theory. Variants of the NSE have been developed for specialized cases, such as seasonal time series (Garrick et al. 1978; Schaefli and Gupta 2007). Perrin et al. (2001) expand the use of the metric for evaluating models across multiple time series.

Regarding the use of functionals in hydrologic modeling, Pande (2013a; 2013b) observe that hydrologic models can be trained to predict specific functionals, such as quantiles. Building on this, Tyralis and Papacharalampous (2021) demonstrate the equivalence of strictly consistent loss functions and target functionals for the estimation and evaluation of hydrologic models for quantiles and Tyralis et al. (2023) extend this work to expectiles. Vrugt (2024) refers to statistical functionals as hydrograph functionals when they possess physical significance in hydrology.

## 3. Theoretical background

This section outlines the theoretical background required for the developments in Sections 4 and 5. Section 3.1 covers the definitions and theory of loss function strict consistency and functional elicitability. Section 3.2 covers the theory of strict identification functions. Due to its central role in characterizing the Nash-Sutcliffe loss, we present a key theorem from Gneiting (2011) on the strict consistency of weighted loss functions separately in Section 3.3. Subsequent sections introduce the concepts of realized losses (Section 3.4) and skill scores (Section 3.5). We then detail model estimation via $M$-estimation in Section 3.6. Finally, Section 3.7 presents the specific strictly consistent loss functions that underpin the construction of the Nash-Sutcliffe loss.

### 3.1 Strictly consistency loss functions and elicitable functionals

Let $\underline{\boldsymbol{y}}_d$ be a $d$-dimensional random variable, with realization $\boldsymbol{y}_d$. We write $\underline{\boldsymbol{y}}_d \sim F$ to indicate that $\underline{\boldsymbol{y}}_d$ has joint CDF $F$, defined as

$$F(\boldsymbol{y}_d) := P(\underline{\boldsymbol{y}}_d \leq \boldsymbol{y}_d) \tag{3.1}$$

where $\underline{\boldsymbol{y}}_d \leq \boldsymbol{y}_d$ indicates an element-wise comparison (see eq. (A.4)).

A $k$-dimensional statistical functional (or simply a functional) $\boldsymbol{T}_k$ is a mapping (Fissler and Ziegel 2016; Gneiting 2011)

$$\boldsymbol{T}_k : \mathcal{F} \to \mathcal{P}(D), F \mapsto \boldsymbol{T}_k(F) \subseteq D \tag{3.2}$$

where $\mathcal{F}$ is a class of probability distributions. This functional assigns to each distribution $F \in \mathcal{F}$ a subset $\boldsymbol{T}_k(F)$ of the domain $D \subseteq \mathbb{R}^k$, that is an element of the power set $\mathcal{P}(D)$.

Now let $\boldsymbol{y}_d \in I \subseteq \mathbb{R}^d$ be a realization of the random variable $\underline{\boldsymbol{y}}_d$ and $\boldsymbol{z}_k \in D \subseteq \mathbb{R}^k$ be a $k$-dimensional functional prediction. A loss function $L$ is a mapping (Fissler and Ziegel 2016; Gneiting 2011)

$$L : D \times I \to \mathbb{R}, (\boldsymbol{z}_k, \boldsymbol{y}_d) \mapsto L(\boldsymbol{z}_k, \boldsymbol{y}_d) \in [0, \infty) \tag{3.3}$$

that assigns a penalty $L(\boldsymbol{z}_k, \boldsymbol{y}_d)$ to the prediction $\boldsymbol{z}_k$ when the realization is $\boldsymbol{y}_d$. The dimension of the functional need not match the dimension of the random variable. For example, the variable could be one-dimensional, while the functional is a pair of quantiles. Loss functions are assumed to be negatively oriented, meaning that a lower penalty indicates a better prediction.

The following definition describes the property of (strict) consistency, which is an appealing characteristic for a loss function. A consistent loss function is, in expectation, minimized by the true value of the functional. It is strictly consistent if this minimum is attained only at the true functional value.

**Definition 1** (Gneiting 2011): A loss function $L$ is $\mathcal{F}$-consistent for the functional $\boldsymbol{T}_k$ if

$$\mathbb{E}_F[L(\boldsymbol{t}_k, \underline{\boldsymbol{y}}_d)] \leq \mathbb{E}_F[L(\boldsymbol{z}_k, \underline{\boldsymbol{y}}_d)] \;\forall\; F \in \mathcal{F}, \boldsymbol{t}_k \in \boldsymbol{T}_k(F), \boldsymbol{z}_k \in D \tag{3.4}$$

It is strictly $\mathcal{F}$-consistent if it is $\mathcal{F}$-consistent and equality in eq. (3.4) implies that $\boldsymbol{z}_k \in \boldsymbol{T}_k(F)$.■

The importance of strict consistency is that it incentivizes a modeler to predict the true value of the functional when her/his goal is to minimize the expected loss on a test set. However, not all functionals possess a strictly consistent loss function. Those that do are called elicitable.

**Definition 2** (Gneiting 2011): The functional $\boldsymbol{T}_k$ is elicitable relative to the class $\mathcal{F}$ if there exists a loss function $L$ that is strictly $\mathcal{F}$-consistent for $\boldsymbol{T}_k$.■

## 3.2 Strictly identification functions and identifiable functionals

Loss functions are useful for comparing predictive models, but they do not indicate whether a single prediction is accurately targeting the true statistical functional. To address this, modelers use identification functions (Gneiting 2011; Nolde and Ziegel 2017). These functions act like diagnostic tools. If a prediction correctly targets a specific functional, this is revealed through the expected value of the identification function. Consider an identification function $V: D \times I \to \mathbb{R}^k$. For a prediction $\boldsymbol{z}_k$ and an observation $\boldsymbol{y}_d$, the value of $V(\boldsymbol{z}_k, \boldsymbol{y}_d)$ indicates whether $\boldsymbol{z}_k$ aligns with the target functional $\boldsymbol{T}_k(F)$.

**Definition 3** (Dimitriadis et al. 2024b; Gneiting 2011): A function $V: D \times I \to \mathbb{R}^k$ is an $\mathcal{F}$-identification function for the functional $\boldsymbol{T}_k(F)$ if it satisfies:

$$\mathbb{E}_F[V(\boldsymbol{z}_k, \underline{\boldsymbol{y}}_d)] = \boldsymbol{0}_k \Rightarrow \boldsymbol{z}_k \in \boldsymbol{T}_k(F) \;\forall\; F \in \mathcal{F}, \boldsymbol{z}_k \in D \subseteq \mathbb{R}^k \tag{3.5}$$

If the converse implication also holds, i.e.

$$\mathbb{E}_F[V(\boldsymbol{z}_k, \underline{\boldsymbol{y}}_d)] = \boldsymbol{0}_k \Leftrightarrow \boldsymbol{z}_k \in \boldsymbol{T}_k(F) \;\forall\; F \in \mathcal{F}, \boldsymbol{z}_k \in D \subseteq \mathbb{R}^k \tag{3.6}$$

then $V$ is said to strictly identify the functional $\boldsymbol{T}_k$ and the functional is referred to as identifiable.■

Identifiability allows us to check whether a predictive model's output genuinely corresponds to its target functional. Suppose a model is designed to predict the mean. If we compute the identification function on a test sample and its average is essentially 0, then the model's prediction is statistically supported. This type of reasoning is closely connected to ideas of calibration (Dawid 1982; Fissler et al. 2023; Gneiting and Resin 2023).

In summary, these concepts serve distinct purposes. Strictly consistent loss functions allow ranking models relative to each other, whereas strict identification functions allow verifying whether a model is actually predicting the true functional. They therefore address different but complementary questions (Fissler and Ziegel 2016).

### 3.3 Strictly consistent weighted loss functions

The following theorem by Gneiting (2011) allows characterizing the (strict) consistency properties of loss functions constructed by applying a weight function to existing (strictly) consistent loss functions.

**Theorem 1** (Gneiting 2011): Let the functional $\boldsymbol{T}_k$ be defined on a class $\mathcal{F}$ of probability distributions that admit a density, $f$, with respect to a dominating measure on the domain $\mathcal{D}$. Consider a measurable weight function

$$w: D \rightarrow [0, \infty) \tag{3.7}$$

Let $\mathcal{F}^{(w)} \subseteq \mathcal{F}$ be the subclass of distributions in $\mathcal{F}$ for which $w(\boldsymbol{y}_d) f(\boldsymbol{y}_d)$ has a finite integral over $D$, and the probability measure $F^{(w)}$, with density proportional to $w(\boldsymbol{y}_d) f(\boldsymbol{y}_d)$, belongs to $\mathcal{F}$. Define the functional

$$\boldsymbol{T}_k^{(w)}: \mathcal{F}^{(w)} \longrightarrow \mathcal{P}(I), F \longmapsto \boldsymbol{T}_k^{(w)}(F) = \boldsymbol{T}_k(F^{(w)}), \tag{3.8}$$

on this subclass $\mathcal{F}^{(w)}$. Then the following holds:

(i) If $\boldsymbol{T}_k$ is elicitable, then $\boldsymbol{T}_k^{(w)}$ is elicitable.

(ii) If $L$ is $\mathcal{F}$-consistent for $\boldsymbol{T}_k$, then

$$L^{(w)}(\boldsymbol{z}_k, \boldsymbol{y}_d) = w(\boldsymbol{y}_d) L(\boldsymbol{z}_k, \boldsymbol{y}_d) \tag{3.9}$$

is $\mathcal{F}^{(w)}$-consistent for $\boldsymbol{T}_k^{(w)}$.

(iii) If $L$ is strictly $\mathcal{F}$-consistent for $\boldsymbol{T}_k$, then $L^{(w)}$ is strictly $\mathcal{F}^{(w)}$-consistent for $\boldsymbol{T}_k^{(w)}$.■

This theorem not only characterizes the weighted loss function but also the resulting elicitable functionals. These functionals are defined by applying the original functional $\boldsymbol{T}_k$ to a transformed distribution, specifically the one whose density is proportional to the weighted original density.

## 3.4 Empirical identification functions and realized losses

The previous definitions address the theoretical population properties of the target random variable. In practice, modelers work with a test set of realizations and predictions from multiple models to assess absolute performance and rank them. The simplest case involves one-dimensional functionals evaluated against one-dimensional realizations. More generally, predictions may concern $k$-dimensional functionals and realizations may come from $d$-dimensional random variables. In all cases, models are diagnostically evaluated using the empirical identification function (Fissler et al. 2023), where values near 0 indicate adequate absolute performance. Models are then ranked by their realized (average) loss, with lower values indicating better performance (Fissler and Ziegel 2016; Gneiting 2011).

### *3.4.1 One-dimensional functionals and one-dimensional realizations*

To assess predictions of a one-dimensional functional $T$, consider a test set of realizations $\boldsymbol{y}_n = (y_1, \dots, y_n)^{\mathrm{T}}$ of the random variable $\underline{y}$. Given corresponding functional predictions $\boldsymbol{z}_n = (z_1, \dots, z_n)^{\mathrm{T}}$, the predictive performance is summarized by the realized loss

$$\bar{L}(\boldsymbol{z}_n, \boldsymbol{y}_n) = (1/n) \textstyle\sum_{j=1}^{n} L(z_j, y_j) \tag{3.10}$$

and the absolute performance is measured by the empirical identification function

$$\bar{V}(\boldsymbol{z}_n, \boldsymbol{y}_n) = (1/n) \textstyle\sum_{j=1}^{n} V(z_j, y_j) \tag{3.11}$$

### *3.4.2 $k$-dimensional functionals and $d$-dimensional realizations*

For $k$-dimensional functionals $\boldsymbol{T}_k = (T_1, \dots, T_k)^{\mathrm{T}}$, consider a test set with $n$ realizations arranged in a $d \times n$ matrix $\boldsymbol{Y}_{d\times n} = [y_{ij}]$, where each column vector $\boldsymbol{Y}_{\cdot,j} = (y_{1j}, \dots, y_{dj})^{\mathrm{T}}$ represents one realization of the $d$-dimensional random variable $\underline{\boldsymbol{y}}_d = (\underline{y}_1, \dots, \underline{y}_d)^{\mathrm{T}}$. The corresponding functional predictions are given by a $k \times n$ matrix $\boldsymbol{Z}_{k\times n} = [z_{ij}]$, where each column $\boldsymbol{Z}_{\cdot,j} = (z_{1j}, \dots, z_{kj})^{\mathrm{T}}$ constitutes one prediction. The predictive performance is summarized by the realized loss

$$\bar{L}(\boldsymbol{Z}_{k\times n}, \boldsymbol{Y}_{d\times n}) = (1/n) \textstyle\sum_{j=1}^{n} L(\boldsymbol{Z}_{\cdot,j}, \boldsymbol{Y}_{\cdot,j}) \tag{3.12}$$

and the absolute performance is measured by the empirical identification function

$$\bar{V}(\boldsymbol{Z}_{k\times n}, \boldsymbol{Y}_{d\times n}) = (1/n) \textstyle\sum_{j=1}^{n} V(\boldsymbol{Z}_{\cdot,j}, \boldsymbol{Y}_{\cdot,j}) \tag{3.13}$$

## 3.5 Skill scores

Skill scores offer an interpretable measure of a model's relative performance against a benchmark (Murphy and Daan 1985). As introduced earlier, the NSE is a skill score specific to the one-dimensional mean functional. The following definitions generalize this concept for both one-dimensional functionals and $k$-dimensional functionals (Gneiting 2011).

### *3.5.1 One-dimensional functionals and one-dimensional realizations*

For a one-dimensional functional, the skill score quantifies the improvement of predictions $\boldsymbol{z}_n$ over a reference $\boldsymbol{z}_{n,\text{ref}}$, given realizations $\boldsymbol{y}_n$ (Gneiting 2011). It is defined as:

$$\bar{L}_{\text{skill}}(\boldsymbol{z}_n, \boldsymbol{y}_n, \boldsymbol{z}_{n,\text{ref}}) := (\bar{L}(\boldsymbol{z}_n, \boldsymbol{y}_n) - \bar{L}(\boldsymbol{z}_{n,\text{ref}}, \boldsymbol{y}_n))/(\bar{L}(\boldsymbol{z}_{n,\text{opt}}, \boldsymbol{y}_n) - \bar{L}(\boldsymbol{z}_{n,\text{ref}}, \boldsymbol{y}_n)) \quad (3.14)$$

where $\bar{L}$ is the realized loss from eq. (3.10)

A common simplification occurs when the optimal loss $\bar{L}(\boldsymbol{z}_{n,\text{opt}}, \boldsymbol{y}_n) = 0$, reducing the expression to

$$\bar{L}_{\text{skill}}(\boldsymbol{z}_n, \boldsymbol{y}_n, \boldsymbol{z}_{n,\text{ref}}) := 1 - \bar{L}(\boldsymbol{z}_n, \boldsymbol{y}_n)/\bar{L}(\boldsymbol{z}_{n,\text{ref}}, \boldsymbol{y}_n) \quad (3.15)$$

The NSE is a special case of eq. (3.15) that arises when using the MSE as the realized loss.

### *3.5.2 $k$-dimensional functionals and $d$-dimensional realizations*

For $k$-dimensional functionals, and $d$-dimensional realizations, the skill score has an analogous definition

$$\bar{L}_{\text{skill}}(\boldsymbol{Z}_{k\times n}, \boldsymbol{Y}_{d\times n}, \boldsymbol{Z}_{d\times n,\text{ref}}) := (\bar{L}(\boldsymbol{Z}_{k\times n}, \boldsymbol{Y}_{d\times n}) - \bar{L}(\boldsymbol{Z}_{d\times n,\text{ref}}, \boldsymbol{Y}_{d\times n}))/(\bar{L}(\boldsymbol{Z}_{d\times n,\text{opt}}, \boldsymbol{Y}_{d\times n}) - \bar{L}(\boldsymbol{Z}_{d\times n,\text{ref}}, \boldsymbol{Y}_{d\times n})) \quad (3.16)$$

where $\bar{L}$ is the realized loss from eq. (3.12). When $\bar{L}\left(\boldsymbol{Z}_{d\times n,\text{opt}}, \boldsymbol{Y}_{d\times n}\right) = 0$ this simplifies to

$$\bar{L}_{\text{skill}}(\boldsymbol{Z}_{k\times n}, \boldsymbol{Y}_{d\times n}, \boldsymbol{Z}_{d\times n,\text{ref}}) := 1 - \bar{L}(\boldsymbol{Z}_{k\times n}, \boldsymbol{Y}_{d\times n})/\bar{L}(\boldsymbol{Z}_{d\times n,\text{ref}}, \boldsymbol{Y}_{d\times n}) \quad (3.17)$$

## 3.6 $M$-estimation

Having established a framework for prediction evaluation using strictly consistent loss functions, we now examine their role in parameter estimation. When a loss function takes an additive form, it generates an $M$-estimator (Huber 1964, 1967; Newey and McFadden 1994). A key result is that if a loss function is strictly consistent for a specific functional, then the corresponding $M$-estimator is consistent for the model's parameter (Dimitriadis et al. 2024a).

This connection bridges two major strands of statistical literature, i.e. prediction evaluation and model estimation. It allows for the transfer of established results between these domains. Furthermore, it implies a clear estimation directive. When tasked with predicting an elicitable functional under a correctly specified model, one must estimate the model using the $M$-estimator associated with a strictly consistent loss function for that functional.

In the following section, we present specific formulations of $M$-estimators for both one-dimensional and $k$-dimensional parameters, as well as for corresponding semiparametric models.

### *3.6.1 Estimation of one-dimensional parameters*

Consider estimating a scalar parameter $\theta$, based on observations $\boldsymbol{y}_n = (y_1, \dots, y_n)^{\mathrm{T}}$ and a loss function $L$. The corresponding $M$-estimator is defined as:

$$\hat{\theta}(\underline{\boldsymbol{y}}_n) := \arg\min_{\theta \in \Theta} (1/n) \sum_{j=1}^{n} L(\theta, \underline{y}_j) \tag{3.18}$$

or equivalently as

$$\hat{\theta}(\underline{\boldsymbol{y}}_n) := \arg\min_{\theta \in \Theta} \bar{L}(\theta \mathbf{1}_n, \underline{\boldsymbol{y}}_n) \tag{3.19}$$

where $\bar{L}$ is the realized loss from eq. (3.10).

### *3.6.2 Semiparametric models for one-dimensional functionals*

Assume a correctly specified semiparametric model $M$ predicts the one-dimensional functional $T(F_{\underline{y}|\underline{\boldsymbol{x}}_p})$, such that $T(F_{\underline{y}|\underline{\boldsymbol{x}}_p}) = M(\underline{\boldsymbol{x}}_p|\boldsymbol{\theta}_0)$ for a unique, unknown parameter $\boldsymbol{\theta}_0$. Given a response vector $\boldsymbol{y}_n = (y_1, \dots, y_n)^{\mathrm{T}}$ and a predictor matrix $\boldsymbol{X}_{p\times n}$ (whose columns are the predictor vectors $\boldsymbol{x}_p$), the corresponding $M$-estimator is given by (Dimitriadis et al. 2024a):

$$\hat{\boldsymbol{\theta}}(\underline{\boldsymbol{X}}_{p\times n}, \underline{\boldsymbol{y}}_n) := \arg\min_{\boldsymbol{\theta} \in \boldsymbol{\Theta}} (1/n) \sum_{j=1}^{n} L(M(\underline{\boldsymbol{X}}_{\cdot,j}|\boldsymbol{\theta}), \underline{y}_j) \tag{3.20}$$

### *3.6.3 Estimation of $k$-dimensional parameters based on $d$-dimensional realizations*

Now, consider estimating a $k$-dimensional parameter $\boldsymbol{\theta}_k = (\theta_1, \dots, \theta_k)^{\mathrm{T}}$. The observations are a $d \times n$ matrix $\boldsymbol{Y}_{d\times n} = [y_{ij}]$. The $M$-estimator for $\boldsymbol{\theta}_k$ is defined as

$$\widehat{\boldsymbol{\theta}}(\underline{\boldsymbol{Y}}_{d\times n}) := \underset{\boldsymbol{\theta}_k \in \boldsymbol{\Theta}}{\arg\min} (1/n) \sum_{j=1}^{n} L(\boldsymbol{\theta}_k, \underline{\boldsymbol{Y}}_{\cdot,j}) \tag{3.21}$$

or equivalently as

$$\widehat{\boldsymbol{\theta}}(\underline{\boldsymbol{Y}}_{d\times n}) := \underset{\boldsymbol{\theta}_k \in \boldsymbol{\Theta}}{\arg\min} \bar{L}(\boldsymbol{\theta}_k \mathbf{1}_n^{\mathrm{T}}, \underline{\boldsymbol{Y}}_{d\times n}) \tag{3.22}$$

where $\bar{L}$ is the realized loss from eq. (3.12).

*3.6.4 Semiparametric models for k-dimensional functionals*

Assume a correctly specified semiparametric model $M$ predicts the $k$-dimensional functional $\boldsymbol{T}_k(F_{\underline{\boldsymbol{y}}_d|\underline{\boldsymbol{x}}_p})$, such that $\boldsymbol{T}_k(F_{\underline{\boldsymbol{y}}_d|\underline{\boldsymbol{x}}_p}) = M(\underline{\boldsymbol{x}}_p|\boldsymbol{\theta}_0)$ for a unique, unknown parameter $\boldsymbol{\theta}_0$. Given a response matrix $\boldsymbol{Y}_{d\times n}$ (whose columns are the observation vectors $\boldsymbol{y}_d$) and a predictor matrix $\boldsymbol{X}_{p\times n}$ (whose columns are the predictor vectors $\boldsymbol{x}_p$), the corresponding $M$-estimator is given by (Dimitriadis et al. 2024a):

$$\widehat{\boldsymbol{\theta}}(\underline{\boldsymbol{X}}_{p\times n}, \underline{\boldsymbol{Y}}_{d\times n}) := \underset{\boldsymbol{\theta} \in \boldsymbol{\Theta}}{\arg\min} (1/n) \sum_{j=1}^{n} L(M(\underline{\boldsymbol{X}}_{\cdot,j}|\boldsymbol{\theta}), \underline{Y}_{\cdot,j}) \tag{3.23}$$

## 3.7 Examples of strict identification functions and strictly consistent loss functions

Gneiting (2011) reviewed the theory of strict identification functions and strictly consistent loss functions, and Tyralis and Papacharalampous (2024; 2025) compiled an extensive list of them. Our focus, here, is on functions that identify or elicit the mean or component-wise mean, as they are essential for constructing the NS loss function.

*3.7.1 Squared error loss function*

Loss functions of the Bregman form (Banerjee et al. 2005) are strictly $\mathcal{F}_{\mathrm{mean},1}$-consistent for the mean functional $\mathbb{E}_F[\underline{y}]$ (Savage 1971).

$$\mathbb{E}_F[\underline{y}] := \int_{-\infty}^{\infty} y \mathrm{d}F(y) \tag{3.24}$$

For the one-dimensional case ($\mathrm{BR}_1$), the loss is defined as:

$$L_{\mathrm{BR}_1}(z, y) := \varphi(y) - \varphi(z) - \frac{\partial \varphi(z)}{\partial z}(y - z) \tag{3.25}$$

where $\varphi$ is a strictly convex function, $\mathcal{F}_{\mathrm{mean},1} = \{F: \mathbb{E}_F[\underline{y}] \text{ exists and is finite}\}$ is the family of distributions with finite mean and 1 in the subscript of $\mathcal{F}_{\mathrm{mean},1}$ for one-dimensional case. A prominent special case is the squared error loss function $L_{\mathrm{SE}}$, previously defined in eq. (1.2) (Gneiting 2011), that arises from eq. (3.25) when $\varphi(t) = t^2$. A strict $\mathcal{F}_{\mathrm{mean},1}$- identification function for the mean is given by (Gneiting 2011)

$$V_{\text{mean}}(z, y) := z - y \tag{3.26}$$

The corresponding realized loss is the MSE, defined in eq. (1.1). Its associated empirical identification function is the Mean Error (ME), defined as (based on eq. (3.11)):

$$\text{ME}(\boldsymbol{z}_n, \boldsymbol{y}_n) := (1/n) \sum_{i=1}^{n} (z_i - y_i) \tag{3.27}$$

and the mean climatology is defined by

$$T_1(\boldsymbol{y}_n) = \mu(\boldsymbol{y}_n) \tag{3.28}$$

*3.7.2 Bregman loss functions for d-dimensional functionals*

Now, consider a $d$-dimensional functional $\boldsymbol{T}_d$ and a $d$-dimensional random variable $\underline{\boldsymbol{y}}_d$. The Bregman loss function $L_{\text{BR}}$ is strictly $\mathcal{F}_{\text{mean},d}$-consistent for the component-wise mean (Banerjee et al. 2005; Gneiting 2011; Osband and Reichelstein 1985; Savage 1971)

$$\boldsymbol{T}_d(F) := \mathbb{E}_F[\underline{\boldsymbol{y}}_d] = (\mathbb{E}_F[\underline{y}_1], \dots, \mathbb{E}_F[\underline{y}_d])^{\text{T}}, F \in \mathcal{F}_{\text{mean},d} \tag{3.29}$$

where $F$ is the joint distribution of $\underline{\boldsymbol{y}}_d$ and $\mathcal{F}_{\text{mean},d} = \{F : \boldsymbol{T}_d(F) \text{ exists and is finite}\}$ is the family of distributions with finite component-wise mean. It is defined as:

$$L_{\text{BR}}(\boldsymbol{z}_d, \boldsymbol{y}_d) := \varphi(\boldsymbol{y}_d) - \varphi(\boldsymbol{z}_d) - \langle \nabla\varphi(\boldsymbol{z}_d), \boldsymbol{y}_d - \boldsymbol{z}_d \rangle \tag{3.30}$$

where $\varphi: \mathbb{R}^d \to \mathbb{R}$ is a strictly convex function with gradient $\nabla\varphi: \mathbb{R}^d \to \mathbb{R}^d$, the gradient of a scalar-valued function of a vector is defined in eq. (A.24) and $\langle \boldsymbol{x}_n, \boldsymbol{y}_n \rangle$ is the Euclidean inner product of two vectors, defined in eq. (A.11).

A strict $\mathcal{F}_{\text{mean},d}$-identification function for the component-wise mean is given by:

$$V_{\text{mean},d}(\boldsymbol{z}_d, \boldsymbol{y}_d) := \boldsymbol{z}_d - \boldsymbol{y}_d \tag{3.31}$$

Substituting $\varphi(\boldsymbol{x}) = \|\boldsymbol{x}\|_2^2$, simplifies $L_{\text{BR}}$ to the Euclidean norm loss function $L_{\text{EN}}$ from eq. (1.8). This loss can also be expressed as:

$$L_{\text{EN}}(\boldsymbol{z}_d, \boldsymbol{y}_d) = \sum_{i=1}^{d} (z_i - y_i)^2 \tag{3.32}$$

The corresponding realized loss is given by (after applying eq. (3.12)):

$$\bar{L}_{\text{EN}}(\boldsymbol{Z}_{d\times n}, \boldsymbol{Y}_{d\times n}) = (1/n) \sum_{j=1}^{n} ||\boldsymbol{Z}_{\cdot,j} - \boldsymbol{Y}_{\cdot,j}||_2^2 \tag{3.33}$$

which is equivalent to:

$$\bar{L}_{\text{EN}}(\boldsymbol{Z}_{d\times n}, \boldsymbol{Y}_{d\times n}) = (1/n) \sum_{j=1}^{n} \sum_{i=1}^{d} (z_{ij} - y_{ij})^2 \tag{3.34}$$

The empirical identification function is

$$\bar{V}_{\text{mean},d}(\boldsymbol{Z}_{d\times n}, \boldsymbol{Y}_{d\times n}) = (1/n) \sum_{j=1}^{n} V_{\text{mean},d}(\boldsymbol{Z}_{\cdot,j}, \boldsymbol{Y}_{\cdot,j}) = (1/n) \sum_{j=1}^{n} (\boldsymbol{Z}_{\cdot,j} - \boldsymbol{Y}_{\cdot,j}) \tag{3.35}$$

The component-wise mean climatology (the empirical counterpart of the component-wise mean) is

$$\boldsymbol{T}_d(\boldsymbol{Y}_{d\times n}) = (\mu(\boldsymbol{Y}_{1,\cdot}^{\mathrm{T}}), \dots, \mu(\boldsymbol{Y}_{d,\cdot}^{\mathrm{T}}))^{\mathrm{T}} \tag{3.36}$$

To employ $L_{\mathrm{EN}}$ as an $M$-estimator for a $d$-dimensional parameter $\boldsymbol{\theta}_d = (\theta_1, \dots, \theta_d)^{\mathrm{T}}$, consider the case where the prediction is a constant vector $\boldsymbol{\theta}_d$ for all observations. The realized loss then simplifies to

$$\overline{L}_{\mathrm{EN}}(\boldsymbol{\theta}_d \mathbf{1}_n^{\mathrm{T}}, \boldsymbol{Y}_{d\times n}) = (1/n) \sum_{j=1}^{n} ||\boldsymbol{\theta}_d - \boldsymbol{Y}_{\cdot,j}||_2^2 \tag{3.37}$$

Given the strict consistency of $L_{\mathrm{EN}}$ for the component-wise mean, the form of $M$-estimators for $d$-dimensional functionals from eq. (3.22) and the equivalence between strict consistency and $M$-estimator consistency discussed in Section 3.6, it follows directly that the minimizer is the vector of sample means:

$$\widehat{\boldsymbol{\theta}}_d(\underline{\boldsymbol{Y}}_{d\times n}) = \underset{\boldsymbol{\theta}_d \in \mathbb{R}^d}{\arg\min}\, \overline{L}_{\mathrm{EN}}(\boldsymbol{\theta}_d \mathbf{1}_n^{\mathrm{T}}, \underline{\boldsymbol{Y}}_{d\times n}) = (\mu(\underline{\boldsymbol{Y}}_{1,\cdot}^{\mathrm{T}}), \dots, \mu(\underline{\boldsymbol{Y}}_{d,\cdot}^{\mathrm{T}}))^{\mathrm{T}} \tag{3.38}$$

which is a consistent estimator of the component-wise mean. For $d = 1$, eqs. (1.8) and (3.32) reduce to the one-dimensional squared error loss in eq. (1.2). Consequently, all relevant results for estimation and evaluation in the $d$-dimensional case naturally extend to the one-dimensional case.

# 4. The Nash-Sutcliffe loss function

In what follows, we present original findings on the Nash-Sutcliffe loss function, focusing on its strict consistency (Section 4.1). The Nash-Sutcliffe loss elicits a functional that we name the Nash-Sutcliffe functional. Its identifiability is examined in Section 4.2. A framework for parameter $M$-estimation with the Nash-Sutcliffe loss is introduced in Section 4.3. We then develop intuition about the Nash-Sutcliffe functional by contrasting it with the component-wise mean in Section 4.4. This section is essential for understanding not only the Nash-Sutcliffe functional but also the broader practice of making and assessing forecasts across multiple time series. Because of its form, the strict consistency and strict identification properties of the Nash-Sutcliffe loss apply to distribution families with support on $\mathbb{R}^d \backslash \{\boldsymbol{y}_d \in \mathbb{R}^d : ||\mu(\boldsymbol{y}_d)\mathbf{1}_d - \boldsymbol{y}_d||_2^2 = 0\}$. Section 4.5 illustrates how to build such distributions from Gaussian distributions with that support. Constructions of this kind are feasible for general distribution families, showing that the Nash-Sutcliffe loss can be applied widely. In Section 4.6, we show that an extended version of the Nash-Sutcliffe loss, which keeps the denominator from being 0 without restricting the support, elicits a transformation of the Nash-Sutcliffe functional. Section 4.7 introduces skill scores constructed from the Nash-Sutcliffe loss. Finally, Section 4.8 presents analytical solutions for the $M$-estimation of linear models using the Nash-Sutcliffe loss, an approach we call Nash-Sutcliffe regression. We demonstrate that Nash-Sutcliffe regression produces different results from ordinary least-squares linear regression in both the one-dimensional and the $d$-dimensional case. The corresponding proofs are included in Appendix C.

## 4.1 Strict consistency of the Nash-Sutcliffe loss function and Nash-Sutcliffe functional

We begin by explaining the motivation for introducing the Nash-Sutcliffe loss function to evaluate $d$-dimensional functionals. In practice, when working with multiple time series, modelers often evaluate predictions using the realized $\overline{\text{NSE}}$ from eq. (1.5) or, equivalently, the realized $\bar{L}_{\text{NS}}$ from eq. (1.10). The typical form of this loss function is:

$$\bar{L}_{\text{NS}}(\boldsymbol{Z}_{d\times n}, \boldsymbol{Y}_{d\times n}) = (1/n)\sum_{j=1}^{n}\left(\sum_{i=1}^{d}(z_{ij} - y_{ij})^2 \Big/ \sum_{i=1}^{d}(\mu(\boldsymbol{Y}_{\cdot,j}) - y_{ij})^2\right), d \geq 2 \quad (4.1)$$

Due to the empirical nature of this procedure, the common approach in the literature has been to estimate a model for each time series using the squared error loss and then compare the resulting predictions using $\bar{L}_{\text{NS}}$. Consequently, modelers have not adapted their estimation procedure to align with the given evaluation loss function. This step is necessary, as discussed in Section 3.6. This common estimation procedure generates predictions that target the one-dimensional mean of each time series (as explained in Section 3.7.1), which does not necessarily align with the, until now, unexplored strict consistency properties of the $L_{\text{NS}}$ loss. For completeness, we restate the definition of the $L_{\text{NS}}$ loss from Section 1.3 in its expanded form:

$$L_{\text{NS}}(\boldsymbol{z}_d, \boldsymbol{y}_d) = \|\boldsymbol{z}_d - \boldsymbol{y}_d\|_2^2 / \|\mu(\boldsymbol{y}_d)\mathbf{1}_d - \boldsymbol{y}_d\|_2^2, d \geq 2 \quad (4.2)$$

Now, consider the standard scenario where a modeler is given a directive to evaluate predictions across multiple time series using $\bar{L}_{\text{NS}}$. From eq. (1.10), one can work backward to arrive at the corresponding $L_{\text{NS}}$ loss in eq. (4.2). The form of eq. (4.2) matches that of a loss function evaluating a $d$-dimensional prediction against a $d$-dimensional realization, as in eq. (3.3). It therefore remains to determine whether Definition 1 of strict consistency applies in this case.

Using Theorem 1, we demonstrate in Proof C.1 (Appendix C) that the NS loss

function is strictly $\mathcal{F}_d^{(w)}$-consistent for the functional

$$\boldsymbol{T}_d^{(w)}(F) = \mathbb{E}_F[\underline{\boldsymbol{y}}_d w(\underline{\boldsymbol{y}}_d)]/\mathbb{E}_F[w(\underline{\boldsymbol{y}}_d)] = (\mathbb{E}_F[\underline{y}_1 w(\underline{\boldsymbol{y}}_d)]/\mathbb{E}_F[w(\underline{\boldsymbol{y}}_d)], \dots, \mathbb{E}_F[\underline{y}_d w(\underline{\boldsymbol{y}}_d)]/ \mathbb{E}_F[w(\underline{\boldsymbol{y}}_d)])^{\mathrm{T}}, d \geq 2 \quad (4.3)$$

Here $\mathcal{F}_d^{(w)} \subseteq \mathcal{F}_{\text{mean},d}$ is the subclass of distributions in $\mathcal{F}_{\text{mean},d}$ for which $w(\boldsymbol{y}_d)f(\boldsymbol{y}_d)$ has finite integral over $\mathbb{R}^d$, and for which the probability measure with density proportional to $w(\boldsymbol{y}_d)f(\boldsymbol{y}_d)$ also belongs to $\mathcal{F}_{\text{mean},d}$.

Appendix C presents an alternative proof that offers further insight into the minimization mechanism underlying the expectation of this loss function. We refer to the functional $\boldsymbol{T}_d^{(w)}(F)$ as the Nash-Sutcliffe functional, which we will elaborate on later in Section 4.4. This functional differs from the component-wise mean elicited by Bregman loss functions, as discussed in Section 3.7.2.

## 4.2 Identifiability of the Nash-Sutcliffe functional

Following Definition 3, the Nash-Sutcliffe functional is identifiable. A strict $\mathcal{F}_d^{(w)}$-identification function for this functional is

$$V_{\text{NS}}(\boldsymbol{z}_d, \boldsymbol{y}_d) = (\boldsymbol{z}_d - \boldsymbol{y}_d)w(\boldsymbol{y}_d), d \geq 2 \quad (4.4)$$

From eq. (3.13), the corresponding empirical identification function follows directly:

$$\bar{V}_{\text{NS}}(\boldsymbol{Z}_{d\times n}, \boldsymbol{Y}_{d\times n}) = (1/n)\sum_{j=1}^{n}(\boldsymbol{Z}_{\cdot,j} - \boldsymbol{Y}_{\cdot,j})w(\boldsymbol{Y}_{\cdot,j}), d \geq 2 \quad (4.5)$$

The Nash-Sutcliffe climatology is defined as the vector $\boldsymbol{T}_d^{(w)}(\boldsymbol{Y}_{d\times n})$ that satisfies $\bar{V}_{\text{NS}}(\boldsymbol{T}_d^{(w)}(\boldsymbol{Y}_{d\times n})\boldsymbol{1}_n^{\mathrm{T}}, \boldsymbol{Y}_{d\times n}) = \boldsymbol{0}_d$:

$$\boldsymbol{T}_d^{(w)}(\boldsymbol{Y}_{d\times n}) = (\sum_{j=1}^{n} w(\boldsymbol{Y}_{\cdot,j})\boldsymbol{Y}_{\cdot,j})/(\sum_{j=1}^{n} w(\boldsymbol{Y}_{\cdot,j})) = (T_1^{(w)}(\boldsymbol{Y}_{d\times n}), \dots, T_d^{(w)}(\boldsymbol{Y}_{d\times n}))^{\mathrm{T}} \quad (4.6)$$

where

$$T_k^{(w)}(\boldsymbol{Y}_{d\times n}) = (\sum_{j=1}^{n} w(\boldsymbol{Y}_{\cdot,j})y_{kj})/(\sum_{j=1}^{n} w(\boldsymbol{Y}_{\cdot,j})), k = 1, \dots, d, d \geq 2 \quad (4.7)$$

and $\boldsymbol{0}_d$ is the zero vector defined in eq. (A.2).

## 4.3 $M$-estimation with the Nash-Sutcliffe loss function

Following the framework established in Section 3.6.3, we now use the Nash-Sutcliffe loss function to construct an $M$-estimator for the Nash-Sutcliffe functional, which is a $d$-dimensional parameter $\boldsymbol{T}_d^{(w)}$ defined in eq. (4.3). This estimator is given by:

$$\widehat{\boldsymbol{\theta}}_d(\underline{\boldsymbol{Y}}_{d\times n}) := \underset{\boldsymbol{\theta}_d\in\mathbb{R}^d}{\arg\min}(1/n)\sum_{j=1}^{n} L_{\mathrm{NS}}(\boldsymbol{\theta}_d, \underline{\boldsymbol{Y}}_{\cdot,j})\,, d \geq 2 \tag{4.8}$$

Due to the strict consistency of $L_{\mathrm{NS}}$ for this functional and the established equivalence between strictly consistent loss functions and the consistency of their corresponding $M$-estimators (Dimitriadis et al. 2024a; see Section 3.6) this estimator is consistent for the Nash-Sutcliffe functional. An alternative proof based on convex analysis, which offers further insight into the properties of this $M$-estimator, appears in Proof C.2 of Appendix C.

## 4.4 Explaining the Nash-Sutcliffe functional

To develop a clearer understanding of the Nash-Sutcliffe loss function, we examine the functional it elicits (the Nash-Sutcliffe functional) and contrast it with the component-wise mean elicited by the Euclidean norm loss function, $L_{\mathrm{EN}}$. First, it is essential to clarify the distinction between $L_{\mathrm{SE}}$ and $L_{\mathrm{EN}}$.

Recall from eq. (1.7) that the Euclidean norm loss, $L_{\mathrm{EN}}$, constitutes the numerator of the Nash-Sutcliffe loss, $L_{\mathrm{NS}}$. The weight function $w(\boldsymbol{y}_d)$ transforms the elicited functional from the component-wise mean to the Nash-Sutcliffe functional. Now, consider the observation matrix $\boldsymbol{Y}_{d\times n}$ in expanded form:

$$\boldsymbol{Y}_{d\times n} = \begin{bmatrix} y_{11} & \cdots & y_{1n} \\ \vdots & \ddots & \vdots \\ y_{d1} & \cdots & y_{dn} \end{bmatrix} \tag{4.9}$$

Using $L_{\mathrm{EN}}$ and $L_{\mathrm{NS}}$ to evaluate predictions $\boldsymbol{Z}_{d\times n}$ implies the following:

(i) Each column of $\boldsymbol{Y}_{d\times n}$ is generated by the random vector $\underline{\boldsymbol{y}}_d$.

(ii) The $i^{\text{th}}$ element of $\underline{\boldsymbol{y}}_d$ has expectation $\mathbb{E}_F[\underline{y}_i]$. Consequently, each row of $\boldsymbol{Y}_{d\times n}$ can be used to estimate the corresponding marginal expectation $\mathbb{E}_F[\underline{y}_i]$.

To understand the relationship between the $d$-dimensional $L_{\mathrm{EN}}$ and the one-dimensional $L_{\mathrm{SE}}$, observe that the latter is a special case of the former for $d = 1$. For $d = 1$, we retain only the first row of the $\boldsymbol{Y}_{d\times n}$ matrix, leaving a single expectation $\mathbb{E}_F[\underline{y}_1]$ to predict. It is crucial not to confuse the realized loss for $L_{\mathrm{SE}}$, defined as $\mathrm{MSE}(\boldsymbol{z}_n, \boldsymbol{y}_n) := (1/n)\sum_{i=1}^{n} L_{\mathrm{SE}}(z_i, y_i)$ (see eq. (1.1)), with the definition of $L_{\mathrm{EN}}$, which is $L_{\mathrm{EN}}(\boldsymbol{z}_d, \boldsymbol{y}_d) = \sum_{i=1}^{d}(z_i - y_i)^2$ (see eq. (3.32)). The former averages over $n$ one-dimensional predictions (rows of $\boldsymbol{Y}_{d\times n}$), while the latter computes a sum over $d$ dimensions within a single prediction. The realized loss of $L_{\mathrm{EN}}$, $\overline{L}_{\mathrm{EN}}$, averages this sum across $n$ columns and is therefore appropriate for evaluating predictions of the $d$-dimensional component-wise expectations. In contrast, $L_{\mathrm{SE}}$ and its realized loss, MSE, are designed only for evaluating a single one-dimensional expectation.

Having clarified the differences between squared error-based losses, we now examine the key distinction between $L_{\mathrm{EN}}$ and $L_{\mathrm{NS}}$. The latter elicits a functional that is a data-weighted version of the component-wise mean. Specifically, in the Nash-Sutcliffe functional, each marginal mean is re-weighted by $w(\underline{\boldsymbol{y}}_d)$. The two functionals are equal if and only if $\mathbb{E}_F[\underline{\boldsymbol{y}}_d] = \mathbb{E}_F[\underline{\boldsymbol{y}}_d w(\underline{\boldsymbol{y}}_d)]/\mathbb{E}_F[w(\underline{\boldsymbol{y}}_d)]$, or equivalently, $\mathbb{E}_F[\underline{\boldsymbol{y}}_d w(\underline{\boldsymbol{y}}_d)] = \mathbb{E}_F[\underline{\boldsymbol{y}}_d]\mathbb{E}_F[w(\underline{\boldsymbol{y}}_d)]$. This condition holds if and only if $\mathbb{E}_F[\underline{y}_i w(\underline{\boldsymbol{y}}_d)] = \mathbb{E}_F[\underline{y}_i]\mathbb{E}_F[w(\underline{\boldsymbol{y}}_d)], i = 1, \dots, d$, meaning that the pairs $\{\underline{y}_i, w(\underline{\boldsymbol{y}}_d)\}$ are uncorrelated. This condition is rarely met in practice; therefore, the component-wise mean and the Nash-Sutcliffe functional are generally different. An illustrative example where the pairs are uncorrelated is when $\underline{\boldsymbol{y}}_d$ consists of independent and identically distributed (IID) normal variables, i.e. $\underline{\boldsymbol{y}}_d \sim N_d(\mu \mathbf{1}_d, \sigma^2 \boldsymbol{I}_d), d > 3$ (see Proof C.3), where $\boldsymbol{I}_d$ is the identity matrix, defined in eq. (A.9).

From a decision-theoretic perspective, the prerequisite for effectively using the Nash-Sutcliffe loss in empirical evaluation is that the $n$ time series in $\boldsymbol{Y}_{d\times n}$ are realizations of the same random vector $\underline{\boldsymbol{y}}_d$. This is a reasonable assumption for time series from the same domain (e.g. $n$ time series of daily temperature). However, the assumption does not hold when mixing series of fundamentally different natures (e.g., daily river flow and daily temperature or daily and monthly temperature). In environmental sciences, it has been empirically observed that NSE values differ substantially between dissimilar datasets (e.g., daily versus monthly river flows). Consequently, practitioners often avoid mixing results from such disparate data sources based on this observation.

We also comment on the influence of the weight $w(\underline{\boldsymbol{y}}_d)$, on the Nash-Sutcliffe loss. First, the value of the Nash-Sutcliffe loss is 1 when the prediction equals the climatology (the sample mean). A loss value below 1, indicates that the method outperforms the naïve climatology benchmark, but it does not, in itself, quantify the absolute quality of the method. For time series with low variability, $w(\underline{\boldsymbol{y}}_d)$ becomes large, causing the Nash-Sutcliffe loss to approach 0. Conversely, for time series with high variability, the loss may be closer to 1, even though the model still outperforms the climatology. In some domains, practitioners have developed intuition through experience to interpret absolute loss values. For example, for daily river flow time series, values near 0.7 may be considered adequate, whereas for monthly river flows, values around 0.3 might be deemed adequate.

As mentioned in Section 3.2, the primary purpose of loss functions lies in model ranking, while assessing absolute performance requires the use of identification functions.

## 4.5 Admissible distribution families for the Nash-Sutcliffe loss function

To establish the practical relevance of the Nash-Sutcliffe loss function, it is essential to characterize the families of probability distributions for which its strict consistency holds. Recall that the Nash-Sutcliffe loss is strictly $\mathcal{F}_d^{(w)}$-consistent for the Nash-Sutcliffe functional in eq. (4.3), $\boldsymbol{T}_d^{(w)}(F) = \mathbb{E}_F[\underline{\boldsymbol{y}}_d w(\underline{\boldsymbol{y}}_d)]/\mathbb{E}_F[w(\underline{\boldsymbol{y}}_d)]$, while the Euclidean norm loss $L_{\mathrm{EN}}$ is strictly $\mathcal{F}_{\mathrm{mean},d}$-consistent for the component-wise mean, in eq. (3.29), $\boldsymbol{T}_d(F) = \mathbb{E}_F[\underline{\boldsymbol{y}}_d]$.

Applying Theorem 1 to the Nash-Sutcliffe loss requires the following assumptions, adapted to our context:

(i) Class inclusion: $\mathcal{F}_d^{(w)} \subseteq \mathcal{F}_{\mathrm{mean},d}$ is the subclass of distributions in $\mathcal{F}_{\mathrm{mean},d}$ for which $w(\boldsymbol{y}_d)f(\boldsymbol{y}_d)$ has a finite integral over $\mathbb{R}^d$.

(ii) Weighted measure membership: The probability measure $F^{(w)}$, with density proportional to $w(\boldsymbol{y}_d)f(\boldsymbol{y}_d)$, belongs to $\mathcal{F}_{\mathrm{mean},d}$; i.e. its component-wise mean exists and is finite.

(iii) Measurability: The weight function $w\colon \mathbb{R}^d \to [0, \infty)$ is measurable, and $\forall F \in \mathcal{F}_d^{(w)}$, the expectation $\mathbb{E}_F[w(\underline{\boldsymbol{y}}_d)]$ exists and is finite.

A critical issue arises from the form of the weight function $w(\boldsymbol{y}_d) = 1/\|\mu(\boldsymbol{y}_d)\mathbf{1}_d - \boldsymbol{y}_d\|_2^2$, which becomes infinite on the set $A = \{\boldsymbol{y}_d \in \mathbb{R}^d \colon \|\mu(\boldsymbol{y}_d)\mathbf{1}_d - \boldsymbol{y}_d\|_2^2 = 0\}$. Since the density of $F^{(w)}$ is proportional to $w(\boldsymbol{y}_d)f(\boldsymbol{y}_d)$, the support of $F^{(w)}$ cannot include points in $A$. The following construction ensures that $F^{(w)}$ has support $\mathbb{R}^d \backslash A$.

A natural way to build such distributions is by truncating standard families. As an illustration, consider a multivariate normal distribution. Let $f_{N_d}(\boldsymbol{y}_d; \boldsymbol{\mu}_d, \boldsymbol{\Sigma}_{d\times d})$ be the density of a $d$-variate normal distribution, given by eq. (B.2). For a given $\delta > 0$ define the truncated set $B_\delta = \{\boldsymbol{y}_d \in \mathbb{R}^d \colon \|\mu(\boldsymbol{y}_d)\mathbf{1}_d - \boldsymbol{y}_d\|_2^2 \geq \delta\}$, which excludes a neighborhood of

the line where all components are equal. We then construct a truncated normal density $(1/C)\, f_{N_d}(\boldsymbol{y}_d; \boldsymbol{\mu}_d, \boldsymbol{\Sigma}_{d\times d})\, \mathbb{1}_{B_\delta}\{\boldsymbol{y}_d\}$, where $C = \int_{B_\delta} f_{N_d}(\boldsymbol{y}_d; \boldsymbol{\mu}_d, \boldsymbol{\Sigma}_{d\times d}) \mathrm{d}\boldsymbol{y}_d$ is the normalization constant and is the indicator function defined in eq. (A.29).

The assumptions are verified as in the following:

(i) Integrability of $w(\boldsymbol{y}_d) f(\boldsymbol{y}_d)$: On $B_\delta$, the weight is bounded, as $w(\boldsymbol{y}_d) \leq 1/\delta$. Since $f$ is proportional to a truncated normal density, the product $w(\boldsymbol{y}_d) f(\boldsymbol{y}_d)$ is integrable over $\mathbb{R}^d$. Thus, $\mathbb{E}_F[w(\underline{\boldsymbol{y}}_d)] < \infty$.

(ii) Membership of $F^{(w)}$ in $\mathcal{F}_{\text{mean},d}$: The density $f^{(w)}(\boldsymbol{y}_d) \propto w(\boldsymbol{y}_d) f(\boldsymbol{y}_d)$ is proportional to a bounded function times a truncated normal density. Its tails are dominated by Gaussian tails, ensuring that the component-wise mean $\mathbb{E}_{F^{(w)}}[\underline{\boldsymbol{y}}_d]$ exists and is finite. Hence, $F^{(w)} \in \mathcal{F}_{\text{mean},d}$.

(iii) Measurability of $w$: The function $w(\boldsymbol{y}_d) = 1/\|\mu(\boldsymbol{y}_d)\mathbf{1}_d - \boldsymbol{y}_d\|_2^2$ is continuous (and hence measurable) on $\mathbb{R}^d \backslash A$, which contains the support of $f$.

This construction demonstrates that for any multivariate normal distribution, a corresponding truncated distribution $F$ can be defined that satisfies all assumptions of Theorem 1 for the Nash-Sutcliffe loss. Since the family of multivariate normal distributions is rich (parameterized by mean vectors and covariance matrices) and such a truncated $F$ can be built from any member (for $d > 3$), the class $\mathcal{F}_d^{(w)}$ is also rich. This demonstrates that the theoretical results concerning the Nash-Sutcliffe loss function hold for a wide range of distributions. While we illustrate the construction using a multivariate normal distribution, the truncation approach applies to any distribution.

### 4.6 Extended Nash-Sutcliffe loss function

As explained in Section 4.5, the denominator of the Nash-Sutcliffe loss can become 0; requiring the class of distributions for which the Nash-Sutcliffe loss is strictly consistent to be restricted. An approach introduced in the form of a realized loss by Kratzert et al. (2019) adds a positive constant to the denominator, ensuring it remains positive. In that work, a small constant was added to mitigate numerical instability during deep learning model training when the denominator approached 0. We formalize this loss function, which we refer to as the extended Nash-Sutcliffe loss, as follows:

$$L_{\text{NSe}}(\boldsymbol{z}_d, \boldsymbol{y}_d; a) = \|\boldsymbol{z}_d - \boldsymbol{y}_d\|_2^2 / (\|\mu(\boldsymbol{y}_d)\mathbf{1}_d - \boldsymbol{y}_d\|_2^2 + a), a \geq 0, d \geq 2 \quad (4.10)$$

For $a = 0$, we have $L_{\text{NSe}}(\boldsymbol{z}_d, \boldsymbol{y}_d; 0) = L_{\text{NS}}(\boldsymbol{z}_d, \boldsymbol{y}_d)$. By following the same reasoning as in Proof C.1, (where only the weight function changes), it is straightforward to prove that the extended Nash-Sutcliffe loss is strictly consistent for a functional we call the extended Nash-Sutcliffe functional:

$$\boldsymbol{T}_{d,a}^{(w)}(F) = \mathbb{E}_F[\underline{\boldsymbol{y}}_d w_a(\underline{\boldsymbol{y}}_d)]/\mathbb{E}_F[w_a(\underline{\boldsymbol{y}}_d)], d \geq 2 \tag{4.11}$$

where

$$w_a(\underline{\boldsymbol{y}}_d) = 1/(\|\mu(\boldsymbol{y}_d)\mathbf{1}_d - \boldsymbol{y}_d\|_2^2 + a), a \geq 0 \tag{4.12}$$

The form of the corresponding realized loss follows directly. The extended Nash-Sutcliffe functional differs from the original Nash-Sutcliffe functional due to the inclusion of the constant $a$ in the weight function.

## 4.7 Skill scores for Nash-Sutcliffe-based forecast evaluation

In accordance with the general skill score formulation presented in eq. (3.17), we define a skill score based on the Nash-Sutcliffe loss function. This score measures the relative improvement of a set of predictions over a chosen reference method. The general form is given by:

$$\bar{L}_{\text{NS,skill}}(\boldsymbol{Z}_{k\times n}, \boldsymbol{Y}_{d\times n}, \boldsymbol{Z}_{d\times n,\text{ref}}) := 1 - \bar{L}_{\text{NS}}(\boldsymbol{Z}_{k\times n}, \boldsymbol{Y}_{d\times n})/\bar{L}_{\text{NS}}(\boldsymbol{Z}_{d\times n,\text{ref}}, \boldsymbol{Y}_{d\times n}), d \geq 2 \tag{4.13}$$

Two natural choices for the reference method lead to particularly interpretable special cases.

In the first case, the reference predictions are the naïve (one-dimensional) climatology, i.e., for each time series $j$, the prediction is the sample mean $\mu(\boldsymbol{Y}_{\cdot,j})$ repeated for all time steps $\boldsymbol{Z}_{\cdot,j} = \mu(\boldsymbol{Y}_{\cdot,j})\mathbf{1}_d$. Substituting this into eq. (4.1) gives $\bar{L}_{\text{NS}}(\boldsymbol{Z}_{d\times n,\text{ref}}, \boldsymbol{Y}_{d\times n}) = 1$. Consequently, the skill score simplifies to $\bar{L}_{\text{NS,skill}}(\boldsymbol{Z}_{k\times n}, \boldsymbol{Y}_{d\times n}, \boldsymbol{Z}_{d\times n,\text{ref}}) = 1 - \bar{L}_{\text{NS}}(\boldsymbol{Z}_{k\times n}, \boldsymbol{Y}_{d\times n})$. In this case, the skill score is simply the negatively oriented complement of the realized Nash-Sutcliffe loss. A value greater than 0 indicates that the model outperforms the naïve (one-dimensional) climatology benchmark.

A more natural reference is the Nash-Sutcliffe climatology, defined as the minimizer of the Nash-Sutcliffe loss (see eq. (4.7)). Here, the reference prediction for each component $i$ is the weighted average across series $\boldsymbol{Z}_{i,\cdot} = (\sum_{j=1}^{n} w(\boldsymbol{Y}_{\cdot,j}) y_{ij} /$ $\sum_{j=1}^{n} w(\boldsymbol{Y}_{\cdot,j}))\mathbf{1}_n^{\mathrm{T}}$ or equivalently $z_{ij} = \sum_{j=1}^{n} w(\boldsymbol{Y}_{\cdot,j}) y_{ij} \,/ \sum_{j=1}^{n} w(\boldsymbol{Y}_{\cdot,j})$. In this case, $\bar{L}_{\mathrm{NS}}(\boldsymbol{Z}_{d\times n,\mathrm{ref}}, \boldsymbol{Y}_{d\times n})$ is not a fixed constant (such as 1) but depends on the data and the weight function. This reference functions as a more tailored benchmark, as it represents the best climatology prediction in the sense of the Nash-Sutcliffe functional. The resulting skill score thus measures improvement relative to this theoretically optimal constant predictor under the Nash-Sutcliffe loss.

These formulations establish a principled framework for assessing relative forecast performance when evaluation is based on the Nash-Sutcliffe loss or its averaged realized version.

## 4.8 Nash-Sutcliffe linear regression

Linear regression is a fundamental model for predicting a continuous response variable from a set of predictors. In the classical setting, the goal is to predict the conditional mean of a one-dimensional response, called linear regression model (Gentle 2024, p.439). When the response is $d$-dimensional (a vector of length $d$), the model extends to multivariate linear regression (Gentle 2024, p.456). In both cases, the standard estimation method is ordinary least squares (OLS), which minimizes the sum of squared errors. In this section, we first review the known results for one-dimensional and $d$-dimensional linear regression. We then introduce a new regression framework, called Nash-Sutcliffe linear regression, which minimizes the Nash-Sutcliffe loss function. This approach gives different estimates than OLS in linear models and is appropriate when the evaluation metric is the realized Nash-Sutcliffe loss.

### *4.8.1 One-dimensional linear regression*

Consider a one-dimensional response variable $\underline{y}$ and a $p$-dimensional predictor vector $\underline{\boldsymbol{x}}_p$. The linear model predicts the conditional mean:

$$z = \boldsymbol{a}_p^{\mathrm{T}} \boldsymbol{x}_p + b \tag{4.14}$$

where $\boldsymbol{a}_p \in \mathbb{R}^p$ and $b \in \mathbb{R}$ are the regression coefficients and intercept, respectively. Let $\boldsymbol{\theta}_{1\times(p+1)} = \begin{bmatrix} \boldsymbol{a}_p^{\mathrm{T}} & b \end{bmatrix} \in \mathbb{R}^{1\times(p+1)}$ be the parameter vector.

Given a sample of $n$ observations $\boldsymbol{y}_n$ and a predictor matrix $\boldsymbol{X}_{p\times n}$, where each column is one observation of the predictor vector $\boldsymbol{x}_p$, the OLS estimate minimizes the realized $L_{\mathrm{SE}}$:

$$\widehat{\boldsymbol{\theta}}_{1\times(p+1)}(\boldsymbol{X}_{p\times n}, \boldsymbol{y}_n) := \arg\min_{\boldsymbol{\theta}\in\boldsymbol{\Theta}} (1/n)\textstyle\sum_{j=1}^{n} L_{\mathrm{SE}}(\boldsymbol{a}_p^{\mathrm{T}}\boldsymbol{X}_{\cdot,j} + b, y_j) \tag{4.15}$$

The well-known closed-form solution for the estimate is (Gentle 2024, p.441; see also Proof C.4 in Appendix C):

$$\widehat{\boldsymbol{\theta}}_{1\times(p+1)}(\boldsymbol{X}_{p\times n}, \boldsymbol{y}_n) = \left[\widehat{\boldsymbol{a}}_p^{\mathrm{T}}(\boldsymbol{X}_{p\times n}, \boldsymbol{y}_n) \quad \widehat{b}(\boldsymbol{X}_{p\times n}, \boldsymbol{y}_n)\right] = \boldsymbol{y}_n^{\mathrm{T}}\widetilde{\boldsymbol{X}}_{n\times(p+1)}\left(\widetilde{\boldsymbol{X}}_{n\times(p+1)}^{\mathrm{T}}\widetilde{\boldsymbol{X}}_{n\times(p+1)}\right)^{-1} \tag{4.16}$$

if the augmented matrix $\widetilde{\boldsymbol{X}}_{n\times(p+1)}$:

$$\widetilde{\boldsymbol{X}}_{n\times(p+1)} := \left[\boldsymbol{X}_{p\times n}^{\mathrm{T}} \quad \mathbf{1}_n\right] \tag{4.17}$$

is of full rank (see definition of full rank of a matrix in Gentle 2024, p.122).

This estimator is consistent for the conditional mean functional when the model is correctly specified. The consistency follows from the strict consistency of the squared error loss for the mean functional (discussed in Section 3.7.1) and the equivalence between the consistency of $M$-estimators and the strict consistency of loss functions, as established in Section 3.6.

The vector of predictions can be written as (see Proof C.4 in Appendix C)

$$\boldsymbol{z}_n^{\mathrm{T}} = \boldsymbol{\theta}_{1\times(p+1)}\widetilde{\boldsymbol{X}}_{n\times(p+1)}^{\mathrm{T}} \tag{4.18}$$

*4.8.2 Multi-dimensional linear regression*

Now consider a $d$-dimensional response vector $\underline{\boldsymbol{y}}_d$ and a $p$-dimensional predictor vector $\underline{\boldsymbol{x}}_p$. The linear model becomes:

$$\boldsymbol{z}_d = \boldsymbol{A}_{d\times p}\boldsymbol{x}_p + \boldsymbol{b}_d \tag{4.19}$$

where $\boldsymbol{A}_{d\times p} \in \mathbb{R}^{d\times p}$ and $\boldsymbol{b}_d \in \mathbb{R}^{d}$. Let $\boldsymbol{\theta}_{d\times(p+1)} = [\boldsymbol{A}_{d\times p} \quad \boldsymbol{b}_d] \in \mathbb{R}^{d\times(p+1)}$ be the parameter matrix.

Given a sample of $n$ observations, with response matrix variable $\boldsymbol{Y}_{d\times n}$, where each column is an observation of the random vector $\underline{\boldsymbol{y}}_d$ and a predictor matrix $\boldsymbol{X}_{p\times n}$, where each column is one observation of the predictor vector $\boldsymbol{x}_p$, the OLS estimate minimizes the realized Euclidean norm:

$$\widehat{\boldsymbol{\theta}}_{d\times(p+1)}(\boldsymbol{X}_{p\times n}, \boldsymbol{Y}_{d\times n}) := \arg\min_{\boldsymbol{\theta}\in\boldsymbol{\Theta}} (1/n)\textstyle\sum_{j=1}^{n} L_{\mathrm{EN}}(\boldsymbol{A}_{d\times p}\boldsymbol{X}_{\cdot,j} + \boldsymbol{b}_d, \boldsymbol{Y}_{\cdot,j}) \tag{4.20}$$

The well-known closed-form solution for the estimate is (Gentle 2024, p.457; see also Proof C.5 in Appendix C):

$$\widehat{\boldsymbol{\theta}}_{d\times(p+1)}(\boldsymbol{X}_{p\times n}, \boldsymbol{Y}_{d\times n}) = \begin{bmatrix}\widehat{\boldsymbol{A}}_{d\times p}(\boldsymbol{X}_{p\times n}, \boldsymbol{Y}_{d\times n}) & \widehat{\boldsymbol{b}}_d(\boldsymbol{X}_{p\times n}, \boldsymbol{Y}_{d\times n})\end{bmatrix} = \boldsymbol{Y}_{d\times n}\widetilde{\boldsymbol{X}}_{n\times(p+1)}(\widetilde{\boldsymbol{X}}^{\mathrm{T}}_{n\times(p+1)}\widetilde{\boldsymbol{X}}_{n\times(p+1)})^{-1} \quad (4.21)$$

where $\widetilde{\boldsymbol{X}}_{n\times(p+1)}$ is the augmented matrix and is of full rank:

$$\widetilde{\boldsymbol{X}}_{n\times(p+1)} := \begin{bmatrix}\boldsymbol{X}^{\mathrm{T}}_{p\times n} & \boldsymbol{1}_n\end{bmatrix} \quad (4.22)$$

This estimator is consistent for the conditional component-wise mean functional when the model is correctly specified. The consistency follows from the strict consistency of the Euclidean norm loss for the component-wise mean functional and the equivalence between the consistency of $M$-estimators and the strict consistency of loss functions, as established in Section 3.6.

The matrix of predictions can be written as (see Proof C.5 in Appendix C)

$$\boldsymbol{Z}_{d\times n} = \boldsymbol{\theta}_{d\times(p+1)}\widetilde{\boldsymbol{X}}^{\mathrm{T}}_{n\times(p+1)} \quad (4.23)$$

Eq. (4.21) shows that each row of $\widehat{\boldsymbol{\theta}}_{d\times(p+1)}$ is simply the estimate from a one-dimensional linear model (Section 4.8.1) fitted to the corresponding row of $\boldsymbol{Y}_{d\times n}$. Specifically, $\widehat{\boldsymbol{\theta}}_{i,\cdot}(\boldsymbol{X}_{p\times n}, \boldsymbol{Y}_{d\times n}) = \boldsymbol{Y}_{i,\cdot}\widetilde{\boldsymbol{X}}_{n\times(p+1)}(\widetilde{\boldsymbol{X}}^{\mathrm{T}}_{n\times(p+1)}\widetilde{\boldsymbol{X}}_{n\times(p+1)})^{-1}$. Similarly, predictions of the form (4.23) decompose into individual one-dimensional predictions of the form (4.18), because $\boldsymbol{Z}_{i,\cdot} = \boldsymbol{\theta}_{i,\cdot}\widetilde{\boldsymbol{X}}^{\mathrm{T}}_{n\times(p+1)}$. Hence, the multi-dimensional linear regression presented here is equivalent to fitting separate one-dimensional linear regressions, each following the model in Section 4.8.1.

#### *4.8.3 Nash-Sutcliffe linear regression*

We now introduce a new regression framework that minimizes the Nash-Sutcliffe loss function. Consider again the $d$-dimensional linear model:

$$\boldsymbol{z}_d = \boldsymbol{A}_{d\times p}\boldsymbol{x}_p + \boldsymbol{b}_d, d \geq 2 \quad (4.24)$$

The Nash-Sutcliffe regression estimate is defined as the minimizer of the realized NS loss over the sample:

$$\widehat{\boldsymbol{\theta}}_{d\times(p+1)}(\boldsymbol{X}_{p\times n}, \boldsymbol{Y}_{d\times n}) := \underset{\boldsymbol{\theta}\in\boldsymbol{\Theta}}{\arg\min}(1/n)\sum_{j=1}^{n} L_{\mathrm{NS}}(\boldsymbol{A}_{d\times p}\boldsymbol{X}_{\cdot,j} + \boldsymbol{b}_d, \boldsymbol{Y}_{\cdot,j}), d \geq 2 \quad (4.25)$$

The closed-form solution for the estimate is (see Proof C.6 in Appendix C):

$$\widehat{\boldsymbol{\theta}}_{d\times(p+1)}(\boldsymbol{X}_{p\times n}, \boldsymbol{Y}_{d\times n}) = \begin{bmatrix}\widehat{\boldsymbol{A}}_{d\times p}(\boldsymbol{X}_{p\times n}, \boldsymbol{Y}_{d\times n}) & \widehat{\boldsymbol{b}}_d(\boldsymbol{X}_{p\times n}, \boldsymbol{Y}_{d\times n})\end{bmatrix} = \boldsymbol{Y}_{d\times n}\boldsymbol{W}_{n\times n}(\boldsymbol{Y}_{d\times n})\widetilde{\boldsymbol{X}}_{n\times(p+1)}(\widetilde{\boldsymbol{X}}^{\mathrm{T}}_{n\times(p+1)}\boldsymbol{W}_{n\times n}(\boldsymbol{Y}_{d\times n})\widetilde{\boldsymbol{X}}_{n\times(p+1)})^{-1}, d \geq 2 \quad (4.26)$$

where $\widetilde{\boldsymbol{X}}_{n\times(p+1)}$ is the augmented matrix and is of full rank:

$$\widetilde{\boldsymbol{X}}_{n\times(p+1)} := \left[\boldsymbol{X}_{p\times n}^{\mathrm{T}} \quad \mathbf{1}_n\right] \tag{4.27}$$

and $\boldsymbol{W}_{n\times n}(\boldsymbol{Y}_{d\times n})$ is the $n \times n$ diagonal weight matrix (diagonal matrices are defined in eq. (A.10))

$$\boldsymbol{W}_{n\times n}(\boldsymbol{Y}_{d\times n}) := \mathrm{diag}(w(\boldsymbol{Y}_{\cdot,1}), \dots, w(\boldsymbol{Y}_{\cdot,n})) \tag{4.28}$$

This estimator is a weighted least squares (WLS) estimator (Gentle 2024, p.289), with weights determined by the variability of each response vector relative to its own mean. Unlike OLS, which treats all observations equally, Nash-Sutcliffe regression assigns higher weight to time series with lower internal variability (smaller denominator in the Nash-Sutcliffe loss). This aligns with the Nash-Sutcliffe loss property of eliciting the Nash-Sutcliffe functional (Section 4.1), which is a weighted version of the component-wise mean.

The matrix of predictions can be written as (see Proof C.6 in Appendix C)

$$\boldsymbol{Z}_{d\times n} = \boldsymbol{\theta}_{d\times(p+1)}\widetilde{\boldsymbol{X}}_{n\times(p+1)}^{\mathrm{T}} \tag{4.29}$$

*4.8.4 Remarks*

The one-dimensional OLS estimator is a special case of the $d$-dimensional OLS estimator, when $d = 1$. Nash-Sutcliffe regression produces different coefficient estimates than OLS. It is the appropriate estimation method when the goal is to generate predictions that perform well under the Nash-Sutcliffe loss. This is particularly relevant in hydrologic and environmental forecasting, where the $\overline{\mathrm{NSE}}$, is a common evaluation metric. Under the assumptions of Theorem 1, the Nash-Sutcliffe regression estimator is consistent for the Nash-Sutcliffe functional (eq. (4.3)) when the linear model is correctly specified for that functional.

In summary, Nash-Sutcliffe regression establishes a principled way to estimate linear models when the evaluation criterion is the Nash-Sutcliffe loss. It adapts the classical regression framework by incorporating a weight structure that reflects the relative importance of each observation vector in the loss function. This approach ensures that the estimation procedure aligns with the evaluation metric, a key principle in forecast evaluation (Dimitriadis et al. 2024a).

## 5. Nash-Sutcliffe loss for making and evaluating forecasts

In the preceding sections, we have examined the Nash-Sutcliffe loss function within a framework where each column of the observation matrix $\boldsymbol{Y}_{d\times n}$ represents a $d$- dimensional realization of a single underlying random vector $\underline{\boldsymbol{y}}_d$. This structure is natural when evaluating predictive performance across multiple fixed-length time series (e.g. $n$ time series each of length $d$).

However, for the purpose of training regression models that will later be used to forecasting settings, a different data arrangement is required. In this section, we therefore transpose the orientation of the data matrix. We now define an observation matrix $\boldsymbol{Y}_{n\times d}$ where each row $\boldsymbol{Y}_{i,\cdot}$ is an independent realization of the random vector $\underline{\boldsymbol{y}}_d$. Consequently, all definitions, including the weight function $w$, the Nash-Sutcliffe loss $L_{\mathrm{NS}}$ and the identification function $V_{\mathrm{NS}}$ are applied to those row vectors.

This row-wise perspective is the standard setup in forecasting. We have $n$ $d$-dimensional observations and the goal is to fit a model that can predict the $d$-dimensional outcome for a new input. The change in orientation necessitates a restatement of the theoretical results from Section 4, which we present below without proof due to their direct analogous nature.

## 5.1 Elicitable and identifiable functionals, loss and identification functions

Let $\underline{\boldsymbol{y}}_d$ be a $d$-dimensional random vector with joint CDF $F$. The definitions and properties (elicitability and identifiability) of the one-dimensional mean, component-wise and Nash-Sutcliffe functional, as well as the definitions and properties of their respective loss and identification functions (strict consistency and strict identification) remain unchanged with the change in orientation of the observation and prediction vectors. The corresponding references to definitions and equations, already introduced, are listed in Table D.1 and are not repeated here for conciseness.

The change in orientation, however, requires redefining the empirical counterparts of functionals (climatologies), loss functions (realized losses) and identification functions (empirical identification functions). Changes in the $M$-estimators and regression models are also necessary and are presented below. The empirical cases for the one-dimensional functional remain largely unchanged, with only minor adjustments required for the regression model.

## 5.2 Realized Euclidean norm and Nash-Sutcliffe losses

Given a matrix of observations $\boldsymbol{Y}_{n\times d}$ and a corresponding matrix of predictions $\boldsymbol{Z}_{n\times d}$, the realized Euclidean norm loss is

$$\bar{L}_{\mathrm{EN}}(\boldsymbol{Z}_{n\times d},\boldsymbol{Y}_{n\times d}) = (1/n)\sum_{i=1}^{n}||\boldsymbol{Z}_{i,\cdot}^{\mathrm{T}} - \boldsymbol{Y}_{i,\cdot}^{\mathrm{T}}||_2^2 = (1/n)\sum_{i=1}^{n}\sum_{j=1}^{d}(z_{ij} - y_{ij})^2 \quad (5.1)$$

and the realized Nash-Sutcliffe loss is

$$\bar{L}_{\mathrm{NS}}(\boldsymbol{Z}_{n\times d},\boldsymbol{Y}_{n\times d}) = (1/n)\sum_{i=1}^{n}(\sum_{j=1}^{d}(z_{ij} - y_{ij})^2 \,/\sum_{j=1}^{d}(\mu(\boldsymbol{Y}_{i,\cdot}^{\mathrm{T}}) - y_{ij})^2)\,, d \geq 2 \quad (5.2)$$

## 5.3 Empirical identification functions

The empirical identification function for the component-wise mean is

$$\bar{V}_{\mathrm{mean},d}(\boldsymbol{Z}_{n\times d},\boldsymbol{Y}_{n\times d}) = (1/n)\sum_{i=1}^{n}V_{\mathrm{mean},d}(\boldsymbol{Z}_{i,\cdot}^{\mathrm{T}},\boldsymbol{Y}_{i,\cdot}^{\mathrm{T}}) = (1/n)\sum_{i=1}^{n}(\boldsymbol{Z}_{i,\cdot}^{\mathrm{T}} - \boldsymbol{Y}_{i,\cdot}^{\mathrm{T}}) \quad (5.3)$$

The component-wise mean climatology (the empirical counterpart of the component-wise mean) is

$$\boldsymbol{T}_d(\boldsymbol{Y}_{n\times d}) = (\mu(\boldsymbol{Y}_{\cdot,1}),\dots,\mu(\boldsymbol{Y}_{\cdot,d}))^{\mathrm{T}} \quad (5.4)$$

For the row-wise data arrangement, the corresponding empirical identification function for the Nash-Sutcliffe functional is:

$$\bar{V}_{\mathrm{NS}}(\boldsymbol{Z}_{n\times d},\boldsymbol{Y}_{n\times d}) = (1/n)\sum_{i=1}^{n}\left(\boldsymbol{Z}_{i,\cdot}^{\mathrm{T}} - \boldsymbol{Y}_{i,\cdot}^{\mathrm{T}}\right)w\left(\boldsymbol{Y}_{i,\cdot}^{\mathrm{T}}\right), d \geq 2 \quad (5.5)$$

The Nash-Sutcliffe climatology is defined as the vector $\boldsymbol{T}_d^{(w)}(\boldsymbol{Y}_{n\times d})$ that satisfies $\bar{V}_{\mathrm{NS}}(\boldsymbol{1}_n(\boldsymbol{T}_d^{(w)}(\boldsymbol{Y}_{n\times d}))^{\mathrm{T}},\boldsymbol{Y}_{n\times d}) = \boldsymbol{0}_d$:

$$\boldsymbol{T}_d^{(w)}(\boldsymbol{Y}_{n\times d}) = (\sum_{i=1}^{n}w\left(\boldsymbol{Y}_{i,\cdot}^{\mathrm{T}}\right)\boldsymbol{Y}_{i,\cdot}^{\mathrm{T}})/(\sum_{i=1}^{n}w(\boldsymbol{Y}_{i,\cdot}^{\mathrm{T}})) =$$

$$(T_1^{(w)}(\boldsymbol{Y}_{n\times d}),\dots,T_d^{(w)}(\boldsymbol{Y}_{n\times d}))^{\mathrm{T}} \quad (5.6)$$

where

$$T_k^{(w)}(\boldsymbol{Y}_{n\times d}) = (\sum_{i=1}^{n}w(\boldsymbol{Y}_{i,\cdot}^{\mathrm{T}})y_{ik})/(\sum_{i=1}^{n}w(\boldsymbol{Y}_{i,\cdot}^{\mathrm{T}})), k = 1,\dots,d, d \geq 2 \quad (5.7)$$

## 5.4 *M*-estimation

Following the framework established in Section 3.6.3, the *M*-estimator for the component-wise mean based on a sample $\boldsymbol{Y}_{n\times d}$ is:

$$\widehat{\boldsymbol{\theta}}_d(\underline{\boldsymbol{Y}}_{n\times d}) := \underset{\boldsymbol{\theta}_d\in\mathbb{R}^d}{\arg\min}(1/n)\sum_{i=1}^{n}L_{\mathrm{EN}}(\boldsymbol{\theta}_d,\underline{\boldsymbol{Y}}_{i,\cdot}^{\mathrm{T}})\,, d \geq 2 \quad (5.8)$$

and the *M*-estimator for the Nash-Sutcliffe functional is:

$$\widehat{\boldsymbol{\theta}}_d(\underline{\boldsymbol{Y}}_{n\times d}) := \underset{\boldsymbol{\theta}_d\in\mathbb{R}^d}{\arg\min}(1/n)\sum_{i=1}^{n}L_{\mathrm{NS}}(\boldsymbol{\theta}_d,\underline{\boldsymbol{Y}}_{i,\cdot}^{\mathrm{T}})\,, d \geq 2 \quad (5.9)$$

Due to the strict consistency of the EN and the Nash-Sutcliffe loss functions for the component-wise mean and the Nash-Sutcliffe functional, respectively, these estimators are consistent for $\boldsymbol{T}_d(F)$ and $\boldsymbol{T}_d^{(w)}$.

## 5.5 Regression modeling

We now adapt the regression framework of Section 4.8 to the row-wise data arrangement. This adaptation is essential for forecasting tasks, where a model is trained on $n$ observed time series (rows) and then used to predict a new $d$-dimensional outcome for a previously unseen input.

### 5.5.1 *One-dimensional linear regression*

Consider a one-dimensional response variable $\underline{y}$ and a $p$-dimensional predictor vector $\underline{\boldsymbol{x}}_p$. The linear model predicts the conditional mean:

$$z = \boldsymbol{a}_p^{\mathrm{T}} \boldsymbol{x}_p + b \tag{5.10}$$

with parameters $\boldsymbol{a}_p \in \mathbb{R}^p$ and $b \in \mathbb{R}$. Let $\boldsymbol{\theta}_{1\times(p+1)} = \begin{bmatrix} \boldsymbol{a}_p^{\mathrm{T}} & b \end{bmatrix} \in \mathbb{R}^{1\times(p+1)}$.

Given a sample of $n$ observations $\boldsymbol{y}_n$ and a predictor matrix $\boldsymbol{X}_{n\times p}$, where each row is one observation of the predictor vector $\boldsymbol{x}_p$, the OLS estimate minimizes the realized $L_{\mathrm{SE}}$:

$$\widehat{\boldsymbol{\theta}}_{1\times(p+1)}(\boldsymbol{X}_{n\times p}, \boldsymbol{y}_n) := \underset{\boldsymbol{\theta}\in\boldsymbol{\Theta}}{\arg\min} (1/n) \textstyle\sum_{i=1}^{n} L_{\mathrm{SE}}(\boldsymbol{a}_p^{\mathrm{T}} \boldsymbol{X}_{i,\cdot}^{\mathrm{T}} + b, y_i) \tag{5.11}$$

The well-known closed-form solution for the estimate is:

$$\widehat{\boldsymbol{\theta}}_{1\times(p+1)}(\boldsymbol{X}_{n\times p}, \boldsymbol{y}_n) = \begin{bmatrix} \widehat{\boldsymbol{a}}_p^{\mathrm{T}}(\boldsymbol{X}_{n\times p}, \boldsymbol{y}_n) & \hat{b}(\boldsymbol{X}_{n\times p}, \boldsymbol{y}_n) \end{bmatrix} =$$
$$\boldsymbol{y}_n^{\mathrm{T}} \widetilde{\boldsymbol{X}}_{n\times(p+1)} (\widetilde{\boldsymbol{X}}_{n\times(p+1)}^{\mathrm{T}} \widetilde{\boldsymbol{X}}_{n\times(p+1)})^{-1} \tag{5.12}$$

if the augmented matrix $\widetilde{\boldsymbol{X}}_{n\times(p+1)}$:

$$\widetilde{\boldsymbol{X}}_{n\times(p+1)} := [\boldsymbol{X}_{n\times p} \quad \boldsymbol{1}_n] \tag{5.13}$$

is of full rank. The vector of predictions can be written as

$$\boldsymbol{z}_n = \widetilde{\boldsymbol{X}}_{n\times(p+1)} \boldsymbol{\theta}_{1\times(p+1)}^{\mathrm{T}} \tag{5.14}$$

### 5.5.2 *Multi-dimensional linear regression*

Now let the response be a $d$-dimensional response vector $\underline{\boldsymbol{y}}_d$ and the predictor be a $p$-dimensional vector $\underline{\boldsymbol{x}}_p$. The linear model becomes

$$\boldsymbol{z}_d = \boldsymbol{A}_{d\times p} \boldsymbol{x}_p + \boldsymbol{b}_d \tag{5.15}$$

with parameters $\boldsymbol{A}_{d\times p} \in \mathbb{R}^{d\times p}$ and $\boldsymbol{b}_d \in \mathbb{R}^d$. Define the parameter matrix $\boldsymbol{\theta}_{d\times(p+1)} = [\boldsymbol{A}_{d\times p} \quad \boldsymbol{b}_d] \in \mathbb{R}^{d\times(p+1)}$.

Given a response matrix variable $\boldsymbol{Y}_{n\times d}$, where each row is an observation of the random vector $\underline{\boldsymbol{y}}_d$ and a predictor matrix $\boldsymbol{X}_{n\times p}$, where each row is one observation of the predictor vector $\boldsymbol{x}_p$, the OLS estimate minimizes the realized Euclidean norm:

$$\widehat{\boldsymbol{\theta}}_{d\times(p+1)}(\boldsymbol{X}_{n\times p},\boldsymbol{Y}_{n\times d}) := \arg\min_{\boldsymbol{\theta}\in\boldsymbol{\Theta}}(1/n)\sum_{i=1}^{n} L_{\text{EN}}(\boldsymbol{A}_{d\times p}\boldsymbol{X}_{i,\cdot}^{\text{T}} + \boldsymbol{b}_d, \boldsymbol{Y}_{i,\cdot}^{\text{T}}) \tag{5.16}$$

The closed-form solution for the estimate is:

$$\widehat{\boldsymbol{\theta}}_{d\times(p+1)}(\boldsymbol{X}_{n\times p},\boldsymbol{Y}_{n\times d}) = \left[\widehat{\boldsymbol{A}}_{d\times p}(\boldsymbol{X}_{n\times p},\boldsymbol{Y}_{n\times d}) \quad \widehat{\boldsymbol{b}}_d(\boldsymbol{X}_{n\times p},\boldsymbol{Y}_{n\times d})\right] = \boldsymbol{Y}_{n\times d}^{\text{T}}\widetilde{\boldsymbol{X}}_{n\times(p+1)}(\widetilde{\boldsymbol{X}}_{n\times(p+1)}^{\text{T}}\widetilde{\boldsymbol{X}}_{n\times(p+1)})^{-1} \tag{5.17}$$

where $\widetilde{\boldsymbol{X}}_{n\times(p+1)}$ is the augmented matrix and is of full rank:

$$\widetilde{\boldsymbol{X}}_{n\times(p+1)} := [\boldsymbol{X}_{n\times p} \quad \boldsymbol{1}_n] \tag{5.18}$$

The matrix of predictions can be written as

$$\boldsymbol{Z}_{n\times d} = \widetilde{\boldsymbol{X}}_{n\times(p+1)}\boldsymbol{\theta}_{d\times(p+1)}^{\text{T}} \tag{5.19}$$

This multivariate regression formulation shares a key property with the model in Section 4.8.2. It can be solved by applying the one-dimensional linear regression method from Section 5.5.1, separately to each dimension of the response variable.

### *5.5.3 Nash-Sutcliffe linear regression*

Consider again the $d$-dimensional linear model:

$$\boldsymbol{z}_d = \boldsymbol{A}_{d\times p}\boldsymbol{x}_p + \boldsymbol{b}_d, d \geq 2 \tag{5.20}$$

The Nash-Sutcliffe regression estimate is defined as the minimizer of the realized Nash-Sutcliffe loss over the sample:

$$\widehat{\boldsymbol{\theta}}_{d\times(p+1)}(\boldsymbol{X}_{n\times p},\boldsymbol{Y}_{n\times d}) := \arg\min_{\boldsymbol{\theta}\in\boldsymbol{\Theta}}(1/n)\sum_{i=1}^{n} L_{\text{NS}}(\boldsymbol{A}_{d\times p}\boldsymbol{X}_{i,\cdot}^{\text{T}} + \boldsymbol{b}_d, \boldsymbol{Y}_{i,\cdot}^{\text{T}}), d \geq 2 \tag{5.21}$$

The closed-form solution for the estimate is:

$$\widehat{\boldsymbol{\theta}}_{d\times(p+1)}(\boldsymbol{X}_{n\times p},\boldsymbol{Y}_{n\times d}) = \left[\widehat{\boldsymbol{A}}_{d\times p}(\boldsymbol{X}_{n\times p},\boldsymbol{Y}_{n\times d}) \quad \widehat{\boldsymbol{b}}_d(\boldsymbol{X}_{n\times p},\boldsymbol{Y}_{n\times d})\right] = \boldsymbol{Y}_{n\times d}^{\text{T}}\boldsymbol{W}_{n\times n}(\boldsymbol{Y}_{n\times d})\widetilde{\boldsymbol{X}}_{n\times(p+1)}(\widetilde{\boldsymbol{X}}_{n\times(p+1)}^{\text{T}}\boldsymbol{W}_{n\times n}(\boldsymbol{Y}_{n\times d})\widetilde{\boldsymbol{X}}_{n\times(p+1)})^{-1}, d \geq 2 \tag{5.22}$$

where $\widetilde{\boldsymbol{X}}_{n\times(p+1)}$ is the augmented matrix and is of full rank:

$$\widetilde{\boldsymbol{X}}_{n\times(p+1)} := [\boldsymbol{X}_{n\times p} \quad \boldsymbol{1}_n] \tag{5.23}$$

and $\boldsymbol{W}_{n\times n}(\boldsymbol{Y}_{n\times d})$ is the $n \times n$ diagonal weight matrix

$$\boldsymbol{W}_{n\times n}(\boldsymbol{Y}_{n\times d}) := \text{diag}(w(\boldsymbol{Y}_{1,\cdot}^{\text{T}}), \dots, w(\boldsymbol{Y}_{n,\cdot}^{\text{T}})) \tag{5.24}$$

The matrix of predictions can be written as

$$\boldsymbol{Z}_{n\times d} = \widetilde{\boldsymbol{X}}_{n\times(p+1)}\boldsymbol{\theta}_{d\times(p+1)}^{\text{T}} \tag{5.25}$$

## 5.6 Estimating the Section 5.5.3 forecasting linear model with the realized Nash-Sutcliffe loss from Section 4

We now estimate the model in eq. (5.25), which aggregates the separate predictions from model (5.20), using the realized Nash-Sutcliffe loss defined in eq. (1.10). Under the current data orientation, the realized loss takes the form

$$\bar{L}_{\mathrm{NS}}(\boldsymbol{Z}_{n\times d}, \boldsymbol{Y}_{n\times d}) = (1/d)\sum_{j=1}^{d} L_{\mathrm{NS}}\left(\boldsymbol{Z}_{\cdot,j}, \boldsymbol{Y}_{\cdot,j}\right), d \geq 2 \quad (5.26)$$

Here, the roles of $d$ and $n$ are reversed compared to eq. (1.10). The index $j$ runs over the $d$ columns of the data matrix, while the model is estimated across the $n$ rows. Using the predictive model (5.20), the $j^{\mathrm{th}}$ column of the prediction matrix can be expressed as

$$\boldsymbol{Z}_{\cdot,j} = \widetilde{\boldsymbol{X}}_{n\times(p+1)}\boldsymbol{\theta}_{j,\cdot}^{\mathrm{T}}, j = 1, \dots, d \quad (5.27)$$

where $\boldsymbol{\theta}_{j,\cdot}$ is the $j^{\mathrm{th}}$ row of the parameter matrix $\boldsymbol{\theta}_{d\times(p+1)}$. Substituting (5.27) into (5.26) gives

$$\bar{L}_{\mathrm{NS}}(\boldsymbol{Z}_{n\times d}, \boldsymbol{Y}_{n\times d}) = (1/d)\sum_{j=1}^{d} w(\boldsymbol{Y}_{\cdot,j})||\widetilde{\boldsymbol{X}}_{n\times(p+1)}\boldsymbol{\theta}_{j,\cdot}^{\mathrm{T}} - \boldsymbol{Y}_{\cdot,j}||_2^2, d \geq 2 \quad (5.28)$$

Because the weights $w(\boldsymbol{Y}_{\cdot,j})$ are positive and each squared-norm term depends exclusively on a single row $\boldsymbol{\theta}_{j,\cdot}$, minimizing (5.28) separates into $d$ independent weighted least-squares problems. Moreover, since the weights do not affect the location of the minimum for each individual term, the overall minimization is equivalent to separately minimizing each of the unweighted squared-norm terms. Consequently, the solution reduces to performing $d$ separate one-dimensional linear regressions as described in Section 5.5.1.

Assuming the augmented design matrix $\widetilde{\boldsymbol{X}}_{n\times(p+1)}$ has full column rank, the estimate for the $j^{\mathrm{th}}$ row of the parameter matrix is

$$\widehat{\boldsymbol{\theta}}_{j,\cdot}^{\mathrm{T}} = \boldsymbol{Y}_{\cdot,j}^{\mathrm{T}}\widetilde{\boldsymbol{X}}_{n\times(p+1)}(\widetilde{\boldsymbol{X}}_{n\times(p+1)}^{\mathrm{T}}\widetilde{\boldsymbol{X}}_{n\times(p+1)})^{-1} \quad (5.29)$$

Collecting these column vectors into a matrix and transposing gives the full parameter estimate

$$\widehat{\boldsymbol{\theta}}_{d\times(p+1)}(\boldsymbol{X}_{n\times p}, \boldsymbol{Y}_{n\times d}) = \boldsymbol{Y}_{n\times d}^{\mathrm{T}}\widetilde{\boldsymbol{X}}_{n\times(p+1)}(\widetilde{\boldsymbol{X}}_{n\times(p+1)}^{\mathrm{T}}\widetilde{\boldsymbol{X}}_{n\times(p+1)})^{-1} \quad (5.30)$$

which is identical to the estimate for the multivariate linear regression model in Section 5.5.2. Thus, estimating the model under the realized Nash-Sutcliffe loss (5.26) recovers the same estimate as the standard multivariate least-squares approach.

Because each row $\boldsymbol{\theta}_{j,\cdot}$ of the parameter matrix is estimated independently using only column $j$ of the data, and because the weight $w(\boldsymbol{Y}_{\cdot,j})$ is constant with respect to $\boldsymbol{\theta}_{j,\cdot}$, minimizing the Nash-Sutcliffe loss for each column gives the same parameter estimates as minimizing the squared error loss.

## 6. Applications

In this section, we present simulation experiments and real-world applications that illustrate the theoretical properties of the Nash-Sutcliffe loss function and the associated Nash-Sutcliffe linear regression. We first compare the Nash-Sutcliffe functional with the component-wise mean under various distributional assumptions (Section 6.1). We then evaluate the performance of the Nash-Sutcliffe linear regression against multi-dimensional linear regression using simulated data, in both the $d \times n$ (Section 6.2) and $n \times d$ (Section 6.3). Finally, we apply these methods to real-world hydrometeorological forecasting tasks within the $n \times d$ setting (Section 6.4). All computations were performed in `R` programming language (R Core Team 2025). The specific `R` packages employed are documented in Appendix E. Full code and detailed results are available in the supplementary material.

### 6.1 Simulation experiment #1: Component-wise mean and Nash-Sutcliffe functional

The first experiment empirically investigates the relationship between the component-wise mean (elicited by the Euclidean norm loss) and the Nash-Sutcliffe functional (elicited by the Nash-Sutcliffe loss) under various distributional assumptions. The objective is to empirically verify the theoretical condition outlined in Section 4.4 (see Proof C.3). To this end, we examine the sample population counterparts of these functionals, namely the component-wise mean climatology and the Nash-Sutcliffe climatology. The two climatologies are equal when the data follow a multivariate Gaussian distribution with a common mean and independent components, or more generally, when the random vector $\underline{\boldsymbol{y}}_d$ and the weight $w(\underline{\boldsymbol{y}}_d)$ are uncorrelated. For non-Gaussian data, the climatologies are expected to differ.

We generated $n = 1\,000$ independent time series, each of length $d = 100$, under five distributional scenarios. All data were organized in $d \times n$ matrices $\boldsymbol{Y}_{d\times n}$, where columns correspond to individual time series and rows to time steps. This arrangement follows the framework of Section 4, where each column is a realization of a $d$-dimensional random vector $\underline{\boldsymbol{y}}_d$.

(i) Simulation Experiment #1a (IID Gaussian variables): We simulated $n$ independent samples from a multivariate Gaussian distribution $N_d(\boldsymbol{\mu}_d, \boldsymbol{\Sigma}_{d\times d})$ with constant mean vector $\boldsymbol{\mu}_d = \mathbf{1}_d$ and diagonal covariance matrix $\boldsymbol{\Sigma}_{d\times d} = 4\boldsymbol{I}_d$ (i.e., independent components, variance 4). After generation, the entire matrix was shifted to a global sample mean of exactly 1, removing any centering artifacts. This scenario satisfies the conditions of Proof C.3, so the component-wise mean climatology and the Nash-Sutcliffe climatology are expected to be nearly identical.

(ii) Simulation Experiment #1b (log-normal variables): To introduce non-Gaussianity, the data from Simulation Experiment #1a were exponentiated element-wise, thereby generating log-normal variables (with probability density functions (PDF) defined by eq. (B.8) in Appendix B). The matrix was then scaled so that its overall mean equaled 1. Log-normality introduces asymmetry, possibly breaking the independence between $\underline{\boldsymbol{y}}_d$ and $w(\underline{\boldsymbol{y}}_d)$.

(iii) Simulation Experiment #1c (independent Gaussian variables with varying component-wise means): We generated independent Gaussian vectors with a non-constant mean vector. Each component mean $\mu_i$ of $\boldsymbol{\mu}_d$ was generated from a $N(1,1)$ distribution. The covariance $\boldsymbol{\Sigma}_{d\times d}$ remained diagonal with variance 4. The $d$-dimensional random vector $\underline{\boldsymbol{y}}_d$ were then simulated from $N_d(\boldsymbol{\mu}_d, \boldsymbol{\Sigma}_{d\times d})$. This setup retains Gaussianity but introduces heterogeneity across time steps; nevertheless, independence between components is retained.

(iv) Simulation Experiment #1d (dependent Gaussian variables with varying component-wise means): Here we introduced dependence, simulating from $N_d(\boldsymbol{\mu}_d, \boldsymbol{\Sigma}_{d\times d})$. The mean vector $\boldsymbol{\mu}_d$ was generated as in Simulation Experiment #1c, but the covariance matrix $\boldsymbol{\Sigma}_{d\times d}$ was constructed using a correlation matrix scaled to variance 4. The data remain Gaussian, but become dependent.

(v) Simulation Experiment #1e (dependent log-normal variables with varying component-wise means): Finally, we exponentiated the Gaussian data from Simulation Experiment #1d, thus simulating log-normal variables with dependence and heterogeneous means. The resulting matrix was again scaled to have overall mean 1. This scenario combines non-Gaussianity, dependence, and varying marginal means and is expected to maximize the divergence between the two functionals.

For each generated dataset, we computed three climatologies:

(i) Component-wise mean climatology: The vector of sample means $\boldsymbol{T}_d(\boldsymbol{Y}_{d\times n}) = (\mu(\boldsymbol{Y}_{1,\cdot}^{\mathrm{T}}), \dots, \mu(\boldsymbol{Y}_{d,\cdot}^{\mathrm{T}}))^{\mathrm{T}}$ from eq. (3.36). This is the $M$-estimator associated with the Euclidean norm loss $L_{\mathrm{EN}}$ and is a consistent estimator of $\mathbb{E}_F[\underline{\boldsymbol{y}}_d]$ (Section 3.7.2).

(ii) Nash-Sutcliffe climatology: The vector $\boldsymbol{T}_d^{(w)}(\boldsymbol{Y}_{d\times n}) = (\sum_{j=1}^{n} w(\boldsymbol{Y}_{\cdot,j})\, \boldsymbol{Y}_{\cdot,j}) / (\sum_{j=1}^{n} w(\boldsymbol{Y}_{\cdot,j}))$ from eq. (4.6). This is a consistent estimator of the Nash-Sutcliffe functional $\boldsymbol{T}_d^{(w)}(F)$ (Section 4.3).

(iii) Time series means (climatology per time series): For each time series $j = 1, \dots, n$, we used its sample mean $\mu(\boldsymbol{Y}_{\cdot,j})$. By construction, this naïve benchmark gives a realized Nash-Sutcliffe loss of exactly 1 and serves as a reference for skill scores.

Table 1 summarizes the realized Euclidean norm (eq. (3.33)) and Nash-Sutcliffe (eq. (4.1)) losses for the five experiments, when the three climatologies are used as predictions. The patterns align closely with the theoretical expectations:

(i) In Simulation Experiment #1a (IID Gaussian), the component-wise mean climatology and the Nash-Sutcliffe climatology produce nearly identical realized Nash-Sutcliffe losses (1.008938 vs. 1.008917), confirming the equivalence from Section 4.4 (Proof C.3). Their realized Euclidean norm losses are also very close with the time series means, but performing slightly worse under $L_{\mathrm{EN}}$ possibly due to sampling uncertainty error.

(ii) Simulation Experiment #1b (log-normal) shows a clear divergence. The Nash-Sutcliffe climatology achieves a substantially lower realized Nash-Sutcliffe loss (1.0142) than the component-wise mean (1.3699), while the opposite holds for the Euclidean norm loss. This demonstrates that when the distribution is non-Gaussian, the two functionals target different population quantities.

(iii) Simulation Experiments #1c and #1d (Gaussian with varying means, independent or dependent) still exhibit near-equivalence of the two climatologies under the Nash-Sutcliffe loss, despite the heterogeneity and correlation. This reinforces that Gaussianity, rather than independence or homogeneity, is the key condition for the equivalence. A theoretical proof for this remains an open problem. Due to introducing heterogeneity, time series means now are outperformed from both the remaining climatologies.

(iv) Simulation Experiment #1e (dependent log-normal) produces the most pronounced

divergence. The Nash-Sutcliffe climatology gives a Nash-Sutcliffe loss of 0.959821, well below the component-wise mean's climatology 1.164348 and also outperforms the time series means benchmark (1.00). The component-wise mean climatology, in turn, has a slight advantage in Euclidean norm loss, as expected.

An overall conclusion from the experiments is that, while realized Nash-Sutcliffe loss values can differ substantially, these differences are not always fully reflected in the corresponding absolute values of the realized Euclidean norm loss. This important finding helps interpret the results presented in the subsequent sections.

Figure 1 visualizes the climatologies for two representative experiments. Panels (a)-(b) correspond to Simulation Experiment #1a. The component-wise mean climatology (blue) and the Nash-Sutcliffe climatology (red) are practically indistinguishable. Panels (c)-(d) show Simulation Experiment #1e, where the Nash-Sutcliffe climatology consistently lies below the component-wise mean climatology.

These results underscore a central message of this paper. When multiple time series forecasts are evaluated by the $\overline{\mathrm{NSE}}$ (or equivalently by $\bar{L}_{\mathrm{NS}}$), the target functional is not the component-wise mean or the time series mean but the Nash-Sutcliffe functional. Consequently, estimation procedures that ignore this distinction by fitting separate models with squared error loss will be suboptimal with respect to the target evaluation metric. The next section demonstrates this point in a regression context.

Table 1. Comparison of realized Euclidean norm and Nash-Sutcliffe losses for three estimation methods (component-wise mean climatology, Nash-Sutcliffe climatology and time series means) across five simulation experiments ($d \times n$ setting). The best performance for each metric is indicated in bold.

| Experiment | Estimation method | | |
|---|---|---|---|
| | Component-wise mean climatology | Nash-Sutcliffe climatology | Time series means |
| Simulation Experiment #1a: IID Gaussian variables | | | |
| Realized Euclidean norm loss | 399.9416 | 399.9496 | **396.4401** |
| Realized Nash-Sutcliffe loss | 1.008938 | 1.008917 | **1.000000** |
| Simulation Experiment #1b: Log-normal variables | | | |
| Realized Euclidean norm loss | 81922.23 | 82017.56 | **81177.59** |
| Realized Nash-Sutcliffe loss | 1.36988 | 1.014202 | **1.000000** |
| Simulation Experiment #1c: Independent Gaussian variables with different component-wise means | | | |
| Realized Euclidean norm loss | **399.8833** | 399.9183 | 517.2399 |
| Realized Nash-Sutcliffe loss | 0.775798 | **0.775729** | 1.000000 |
| Simulation Experiment #1d: Dependent Gaussian variables with different component-wise means | | | |
| Realized Euclidean norm loss | **398.1114** | 398.158 | 516.794 |
| Realized Nash-Sutcliffe loss | 0.7724183 | **0.772326** | 1.000000 |
| Simulation Experiment #1e: Dependent log-normal variables with different component-wise means | | | |
| Realized Euclidean norm loss | **6846.977** | 6917.679 | 6902.888 |
| Realized Nash-Sutcliffe loss | 1.164348 | **0.959821** | 1.000000 |

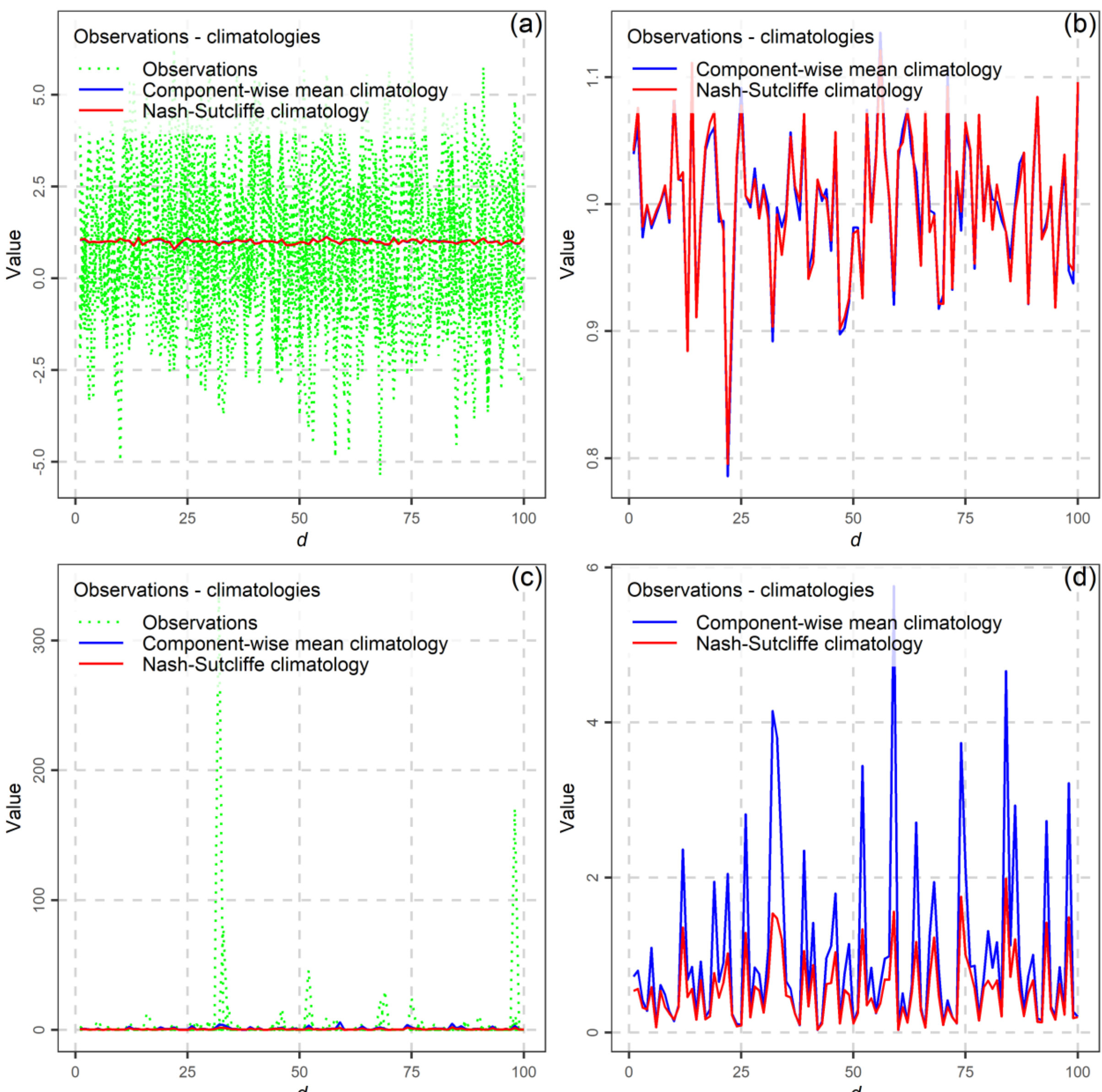

Figure 1. Comparison of the component-wise mean climatology and the Nash-Sutcliffe climatology for simulated data. (a) Ten randomly selected time series (green dotted lines) from Simulation Experiment #1a (IID Gaussian variables), plotted with the component-wise mean climatology (blue solid line) and the Nash-Sutcliffe climatology (red solid line). (b) The same estimates as in (a), shown without the observations to highlight the near-perfect overlap between the two climatologies for Gaussian data. (c) Ten randomly selected time series from Simulation Experiment #1e (dependent log-normal variables), plotted with the component-wise mean climatology (blue) and Nash-Sutcliffe climatology (red). (d) The same estimates as in (c), shown without the observations. The Nash-Sutcliffe climatology consistently lies below the component-wise mean climatology in (d).

## 6.2 Simulation Experiment #2: Linear regression ($d \times n$ setting)

This experiment compares two linear regression estimation methods in the $d \times n$ data setting (Section 4.8), where each column of the data matrix represents a $d$-dimensional realization of the response vector and the goal is to predict the entire vector from a set of predictors. This setting corresponds to the case that observations of the predictor and response (time series) variables are given at $n$ points in space and we predict time series at new points in space where only predictor variables are available. The two methods are:

(i) Multi-dimensional linear regression: The model to be estimated is given by eq. (4.19). The classical estimator, defined in eq. (4.21), minimizes the realized Euclidean norm loss (eq. (4.20)) and is strictly consistent for the conditional component-wise mean functional (eq. (3.29)).

(ii) Nash-Sutcliffe linear regression: The model to be estimated is given by eq. (4.24). The estimator introduced in Section 4.8.3 , defined in eq. (4.26), minimizes the realized Nash-Sutcliffe loss (eq. (4.25)) and is strictly consistent for the conditional Nash-Sutcliffe functional (eq. (4.3)).

The objective is to verify that, under a correctly specified linear model, but with a log-normal error distribution that makes the two target functionals differ, each estimation method performs best according to its own evaluation metric.

We generated synthetic data following the linear model

$$\underline{\boldsymbol{Y}}_{d\times n} = \boldsymbol{A}_{d\times p}\underline{\boldsymbol{X}}_{p\times n} + \boldsymbol{b}_d\boldsymbol{1}_n^{\mathrm{T}} + \underline{\boldsymbol{E}}_{d\times n} \tag{6.1}$$

with dimension $d = 100$ (length of each time series), number of series $n = 1\,000$ and $p = 6$ predictors. The predictor matrix $\boldsymbol{X}_{p\times n}$ was filled with independent standard normal variates. The true coefficient matrix $\boldsymbol{A}_{d\times p}$ and intercept vector $\boldsymbol{b}_d$ were generated independently from Gaussian distributions: $a_{i,j} \sim N(0{,}0.5)$, $b_i \sim N(1{,}4)$.

To create a scenario where the component-wise mean and the Nash-Sutcliffe functional diverge, we used a correlated log-normal error structure. First, a random mean vector $\underline{\boldsymbol{\mu}_{\varepsilon}} \sim N_d(\boldsymbol{0}_d, 6^2\boldsymbol{I}_d)$ was generated and a covariance matrix $\boldsymbol{\Sigma}_{\underline{\varepsilon}}$ was constructed with variance 4. Then $n$ independent error vectors were generated from a multivariate log-normal distribution $\underline{\boldsymbol{E}}_{\cdot,j} = \exp(\underline{\boldsymbol{\varepsilon}}_j)$ where $\underline{\boldsymbol{\varepsilon}}_j \sim N_d(\boldsymbol{\mu}_{\underline{\varepsilon}}, \boldsymbol{\Sigma}_{\underline{\varepsilon}})$. The final response matrix was constructed by adding the linear predictor $\boldsymbol{A}_{d\times p}\boldsymbol{X}_{p\times n} + \boldsymbol{b}_d\boldsymbol{1}_n^{\mathrm{T}}$ to the error matrix $\boldsymbol{E}_{d\times n}$.

The $n = 1\,000$ series were split into a training set (first $n_1 = 500$ columns) and a test set (remaining $n_2 = 500$ columns), where predictors are created by training and test matrices $\boldsymbol{X}_{p\times n_1}$ and $\boldsymbol{X}_{p\times n_2}$ respectively, while observed responses are formed as training and test matrices of $\boldsymbol{Y}_{d\times n_1}$ and $\boldsymbol{Y}_{d\times n_2}$ respectively. The two regression methods were estimated on the training set. Predictions were then generated for both training and test sets using eqs. (4.23) and (4.29). Performance was measured by:

(i) The realized Euclidean norm loss $\bar{L}_{\mathrm{EN}}$ (eqs. (3.33) and (3.34)).

(ii) The realized Nash-Sutcliffe loss $\bar{L}_{\mathrm{NS}}$ (eq. (1.10) or (4.1)), which is the natural metric for the Nash-Sutcliffe functional.

Table 2 reports the realized losses for both methods on the training and test sets. The results are in good agreement with the theoretical consistency properties of the estimators:

(i) Euclidean norm loss: Multi-dimensional linear regression attains a lower value than Nash-Sutcliffe linear regression in both training and test sets. This is expected because multi-dimensional linear regression directly targets the conditional component-wise mean, which minimizes the expected Euclidean norm loss.

(ii) Nash-Sutcliffe loss: Nash-Sutcliffe linear regression dramatically outperforms multi-dimensional linear regression, achieving a realized Nash-Sutcliffe loss of 0.8358 (training) and 0.8913 (test) compared to multi-dimensional linear regression values of 25.6133 and 26.3717. This confirms that Nash-Sutcliffe linear regression is the appropriate estimator when the evaluation criterion is the realized NSE ($\overline{\mathrm{NSE}}$).

Figure 2 visualizes the predictions for the first time series in the test set (panel a) and the overall scatter of observed versus predicted values across all test set series (panel b). The blue points (OLS) and red points (Nash-Sutcliffe) form distinct clusters. OLS predictions scatter more widely around the 1: 1 line in terms of Nash-Sutcliffe loss, while Nash-Sutcliffe predictions are systematically shifted, reflecting the different functional they target. The plot illustrates that the two estimators are not interchangeable; their optimality depends on the evaluation metric employed.

In addition, neither model appears to be a good absolute predictor of the response variable, as the distance between observations and predictions is substantial. This is by design, because the log-normal error structure was intentionally chosen to deviate substantially from Gaussianity, thereby inducing asymmetry that is inherently difficult to predict. The limitation, therefore, lies with the model rather than with the loss function itself. In contrast, the linear model predictions in the real-world applications presented in subsequent sections align more closely with the observations. It is also important to recall that the models are designed to predict functionals of the predictive distributions, not the observed values themselves. Given the error structure and its spread, the point forecasts do not lend themselves to a favorable visual impression. Finally, the realized Nash-Sutcliffe losses exhibit substantial variation, while the Euclidean norm losses appear more similar, although the differences there remain notable. This pattern was also observed in the experiment of Section 6.1.

The results of this simulation reinforce the central message of the paper in regression settings. When forecasts are to be evaluated by the $\overline{\text{NSE}}$ (or equivalently by the realized Nash-Sutcliffe loss), the estimation procedure must be aligned with that evaluation metric. Using multi-dimensional linear regression (which minimizes the Euclidean norm loss) leads to predictions that are suboptimal under the Nash-Sutcliffe loss, even though the model is correctly specified. Conversely, Nash-Sutcliffe linear regression, which is specifically designed to minimize the Nash-Sutcliffe loss, issues predictions that perform best under that metric, albeit at a slight cost in Euclidean norm performance. The slight disadvantage in Euclidean norm does not indicate any deficiency of the method but rather reflects the fact that the two functionals are distinct under non-Gaussian, dependent errors.

These findings have direct practical implications for time series modeling and other fields where the NSE is routinely used to compare models across multiple time series. Modelers aiming to to optimize directly for the $\overline{\text{NSE}}$ should adopt the Nash-Sutcliffe regression framework introduced here.

Table 2. Comparison of realized Euclidean norm and Nash-Sutcliffe losses for multi-dimensional linear regression versus Nash-Sutcliffe linear regression in a $d \times n$ data setting (Simulation Experiment #2). The best performance for each metric is indicated in bold.

| | Regression method | |
|---|---|---|
| Performance metric | Multi-dimensional linear regression | Nash-Sutcliffe linear regression |
| Training set performance | | |
| Realized Euclidean norm loss | **1.038644 $10^{16}$** | 1.085268 $10^{16}$ |
| Realized Nash-Sutcliffe loss | 25.6133 | **0.8358** |
| Test set performance | | |
| Realized Euclidean norm loss | **5.670116 $10^{15}$** | 5.804128 $10^{15}$ |
| Realized Nash-Sutcliffe loss | 26.3717 | **0.8913** |

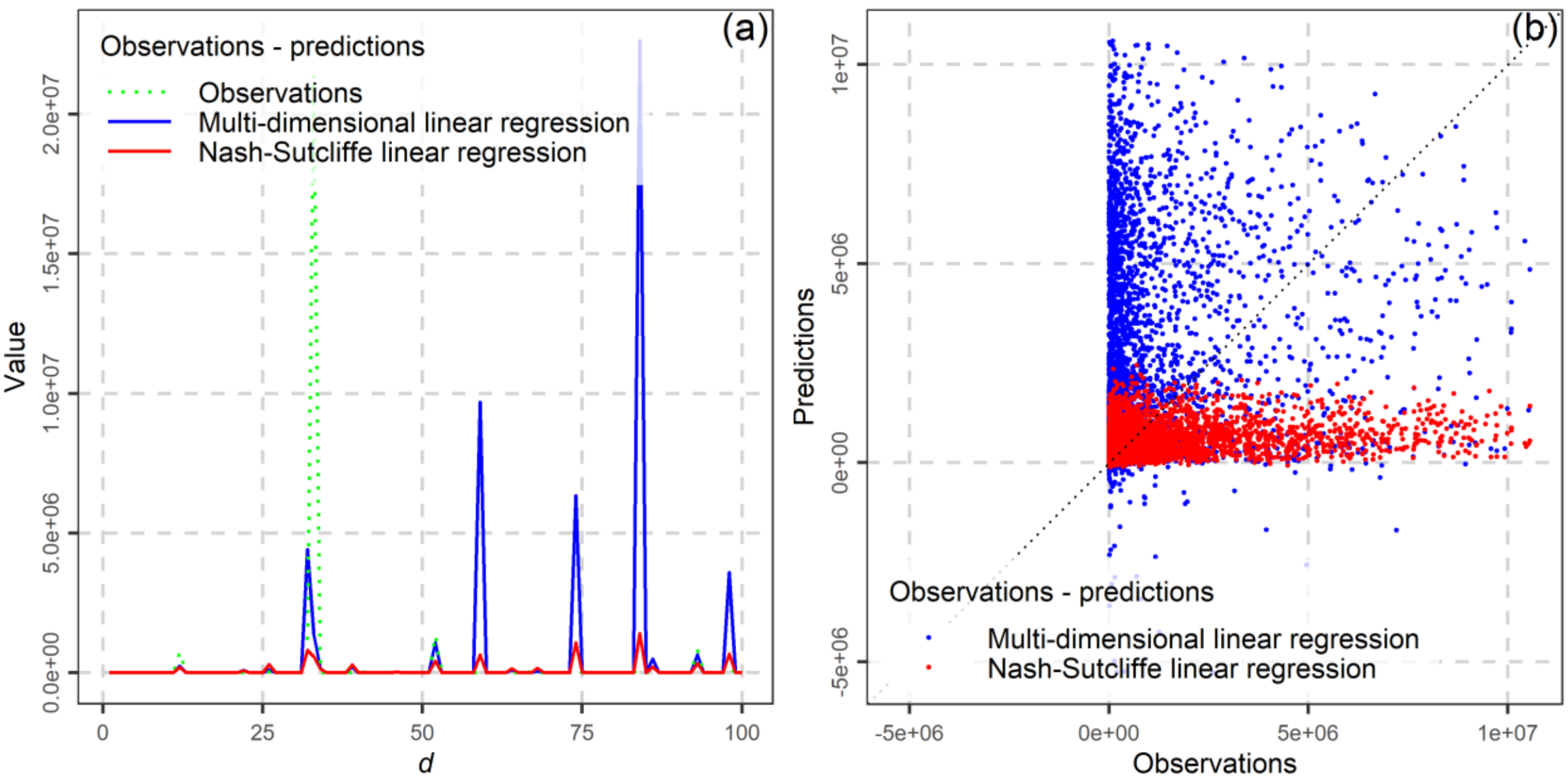

Figure 2. Comparison of multi-dimensional linear regression and Nash-Sutcliffe linear regression predictions for Simulation Experiment #2 ($d \times n$ setting). (a) Time series from the test set (green dotted line) plotted with predictions from multi-dimensional linear regression (blue solid line) and Nash-Sutcliffe linear regression (red solid line). (b) Scatterplot of all observed values versus predicted values for the test set; the range of values is restricted for visual clarity. Blue points correspond to multi-dimensional linear regression predictions and red points correspond to Nash-Sutcliffe linear regression predictions. The dotted line represents the 1: 1 diagonal.

### 6.3 Simulation Experiment #3: Linear regression ($n \times d$ setting)

This experiment replicates the comparison of Section 6.2 but in the $n \times d$ data orientation, which corresponds to the forecasting setup described in Section 5. Here each row of the data matrix represents a realization of a $d$-dimensional response vector, i.e., a multivariate observation and the goal is to predict the entire new vector from a set of predictors. The two estimation methods are again:

(i) Multi-dimensional linear regression: The model to be estimated is given by eq. (5.15). The classical estimator, defined in eq. (5.17), minimizes the realized Euclidean norm loss (eq. (5.1)) and is strictly consistent for the conditional component-wise mean functional (eq. (3.29)).

(ii) Nash-Sutcliffe linear regression: The model to be estimated is given by eq. (5.20). The estimator introduced in Section 5.5.3 , defined in eq. (5.22), minimizes the realized Nash-Sutcliffe loss (eq. (5.2)) and is strictly consistent for the conditional Nash-Sutcliffe functional (eq. (4.3)).

The data-generating process is identical to that of Section 6.2, except that the matrices are transposed to conform to the $n \times d$ layout. Following the linear model

$$\underline{\boldsymbol{Y}}_{n\times d} = \underline{\boldsymbol{X}}_{n\times p}\boldsymbol{A}_{d\times p}^{\mathrm{T}} + \mathbf{1}_n\boldsymbol{b}_d^{\mathrm{T}} + \underline{\boldsymbol{E}}_{n\times d} \quad (6.2)$$

we set the dimensions to $n = 1\,000$ (number of observations), $d = 100$ (dimension of the response) and $p = 6$ (number of predictors). The predictor matrix $\boldsymbol{X}_{n\times p}$ was filled with independent standard normal variates. The true coefficient matrix $\boldsymbol{A}_{d\times p}$ and intercept vector $\boldsymbol{b}_d$ were generated from the same distributions as in Section 6.2, i.e. $a_{i,j}{\sim}N(0{,}0.5), b_i{\sim}N(1{,}4)$.

To create a discrepancy between the two target functionals, we again used correlated log-normal errors. First, a random mean vector $\underline{\boldsymbol{\mu}_{\varepsilon}}{\sim}N_d(\mathbf{0}_d, 6^2\boldsymbol{I}_d)$ was generated and a covariance matrix $\boldsymbol{\Sigma}_{\underline{\varepsilon}}$ with variance 4 and correlations was constructed. Then $n$ independent error vectors were generated from a multivariate log-normal distribution $\underline{\boldsymbol{E}}_{i,\cdot} = \exp(\underline{\boldsymbol{\varepsilon}}_i)$ where $\underline{\boldsymbol{\varepsilon}}_i{\sim}N_d(\boldsymbol{\mu}_{\underline{\varepsilon}}, \boldsymbol{\Sigma}_{\underline{\varepsilon}})$. The response matrix was constructed by adding the linear predictor $\boldsymbol{X}_{n\times p}\boldsymbol{A}_{d\times p}^{\mathrm{T}} + \mathbf{1}_n\boldsymbol{b}_d^{\mathrm{T}}$ to the error matrix $\boldsymbol{E}_{n\times d}$.

The $n = 1\,000$ series were split into a training set (first $n_1 = 500$ columns) and a test set (remaining $n_2 = 500$ columns), where predictors are created by training and test matrices $\boldsymbol{X}_{n_1\times p}$ and $\boldsymbol{X}_{n_2\times p}$ respectively, while observed responses are formed as training and test matrices of $\boldsymbol{Y}_{n_1\times d}$ and $\boldsymbol{Y}_{n_2\times d}$ respectively. The two regression methods were estimated on the training set. Predictions were then generated for both training and test sets using eqs. (4.23) and (4.29). Performance was measured by:

(i) The realized Euclidean norm loss $\bar{L}_{\mathrm{EN}}$ (eq. (5.1)).

(ii) The realized Nash-Sutcliffe loss $\bar{L}_{\mathrm{NS}}$ (eq. (5.2)), which is the natural metric for the Nash-Sutcliffe functional.

Table 3 reports the realized losses for both methods on the training and test sets. The pattern is identical to that observed in the $d \times n$ orientation:

(i) Realized Euclidean norm loss: Multi-dimensional linear regression achieves a lower value than Nash-Sutcliffe linear regression in both training and test sets, confirming that it is optimal for the component-wise mean.

(ii) Realized Nash-Sutcliffe loss: Nash-Sutcliffe linear regression dramatically outperforms multi-dimensional linear regression, achieving a realized Nash-Sutcliffe loss of 0.8404 (training) and 0.8944 (test) compared to multi-dimensional linear regression values of 32.4277 and 27.1941. This demonstrates that Nash-Sutcliffe regression is the correct estimator when the evaluation metric is the Nash-Sutcliffe loss, regardless of whether the data are organized by columns ($d \times n$) or by rows ($n \times d$).

Table 3. Comparison of realized Euclidean norm and Nash-Sutcliffe losses for multi-dimensional linear regression versus Nash-Sutcliffe linear regression in a $n \times d$ data setting (Simulation Experiment #3). The best performance for each metric is indicated in bold.

| Performance metric | Regression method | |
|---|---|---|
| | Multi-dimensional linear regression | Nash-Sutcliffe linear regression |
| Training set performance | | |
| Realized Euclidean norm loss | **1.027420 $10^{16}$** | 1.085776 $10^{16}$ |
| Realized Nash-Sutcliffe loss | 32.4277 | **0.8404** |
| Test set performance | | |
| Realized Euclidean norm loss | **5.713625 $10^{15}$** | 5.811100 $10^{15}$ |
| Realized Nash-Sutcliffe loss | 27.1941 | **0.8944** |

## 6.4 Real data applications

To bridge the theoretical developments with practical cases, we apply the three regression frameworks of Section 5.5 to two real-world datasets. The data are sourced from the `airGRdatasets R` package (Delaigue et al. 2023), which includes hydrometeorological time series for a set of French river basins. For each river basin, the package includes time series of mean temperature (measured in °C) and streamflow (measured in mm/day). From this dataset, we selected the ten basins with complete records.

For each variable (streamflow and temperature), we form a response matrix $\boldsymbol{Y}_{n \times d}$ with $n = 7\,305$ days (time steps) and $d = 10$ basins (time series). To construct predictors that represent temporal dynamics, we formed lagged versions of the variable of interest. Specifically, for every time series we include its value at the previous day (lag 1) and two days earlier (lag 2). Because the dataset contains 10 time series, this gives a predictor matrix $\boldsymbol{X}_{n \times p}$ with $p = 20$ columns (2 lags × 10 series). The first two rows of the original series are lost due to lagging, leaving $n = 7\,303$ complete observations. We then split the data chronologically. The first 4 000 days serve as the training set and the remaining 3 303 days are held out for testing. This train/test split mimics a forecasting situation where a model is estimated on historical data and evaluated on unseen observations.

The three estimation methods compared are:

(i) One-dimensional linear regression: For each time series $j$, a univariate linear model is fitted using only its own two lagged values $y_{t-1,j}$ and $y_{t-2,j}$ as predictors. This is the conventional approach when each series is modeled independently; it ignores any cross-series information. The model to be estimated is given by eq. (5.10). The classical estimator, defined in eq. (5.12), minimizes the MSE (eq. (1.1)) and is strictly consistent for the conditional mean functional (eq. (3.24)).

(ii) Multi-dimensional linear regression: The model to be estimated is given by eq. (5.15). The classical estimator, defined in eq. (5.17), minimizes the realized Euclidean norm loss (eq. (5.1)) and is strictly consistent for the conditional component-wise mean functional (eq. (3.29)).

(iii) Nash-Sutcliffe linear regression: The model to be estimated is given by eq. (5.20). The estimator introduced in Section 5.5.3, defined in eq. (5.22), minimizes the realized Nash-Sutcliffe loss (eq. (5.2)) and is strictly consistent for the conditional Nash-Sutcliffe functional (eq. (4.3)).

All three methods are estimated on the training set. For one-dimensional linear regression, 10 independent models are fitted. For multi-dimensional linear regression and Nash-Sutcliffe linear regression, a single model is fitted across all time series. Predictions are then generated for both training and test sets using the estimated parameters. Performance is measured by two realized losses:

(i) The realized Euclidean norm loss $\overline{L}_{\mathrm{EN}}$ (eq. (5.1)).

(ii) The realized Nash-Sutcliffe loss $\overline{L}_{\mathrm{NS}}$ (eq. (5.2)), which is the natural metric for the Nash-Sutcliffe functional.

The results are presented and interpreted separately for streamflow (Section 6.4.1) and temperature (Section 6.4.2).

### *6.4.1 Streamflow forecasting*

Table 4 reports the realized losses for the three methods:

(i) Euclidean norm loss: The multi-dimensional linear regression achieves the lowest realized Euclidean norm loss on both training and test sets. This is expected because multi-dimensional linear regression is the $M$-estimator associated with the Euclidean norm loss and is strictly consistent for the conditional component-wise mean. By pooling information across time series, it reduces the average squared error compared to one-dimensional linear regression, which ignore cross-time series dependencies. Nash-Sutcliffe linear regression, although it uses the same set of predictors, achieves a slightly higher realized Euclidean norm loss (about 3% higher on the test set) because it targets the Nash-Sutcliffe functional.

(ii) Realized Nash-Sutcliffe loss: Nash-Sutcliffe linear regression strongly outperforms both other methods. On the test set, its realized Nash-Sutcliffe loss is 0.1222, compared to 0.3791 for one-dimensional linear regression and 0.2244 for multi-dimensional linear regression. In relative terms, Nash-Sutcliffe linear regression reduces the Nash-Sutcliffe loss by approximately 68% relative to one-dimensional linear regression and by 46% relative to multi-dimensional linear regression. These dramatic improvements confirm that when the goal is to maximize the $\overline{\text{NSE}}$ (or equivalently minimize the realized Nash-Sutcliffe loss), the estimation procedure must be tailored to that metric. The multi-dimensional linear regression, despite its use of cross-time series information, remains suboptimal under the $\overline{\text{NSE}}$ because it minimizes a mismatched loss function during training.

Figure 3 (panel a) shows the observed and predicted streamflow for a single series of the test set, while Panel b presents a scatterplot of all observed versus predicted values across all 10 time series in the test set. Predictions of all methods match observation better compared to illustrations of the corresponding experiment of Section 6.2. These results demonstrate empirical evidence that Nash-Sutcliffe linear regression is the appropriate estimation method when the evaluation metric is the NSE.

Table 4. Performance comparison of three regression methods (one-dimensional linear regression, multi-dimensional linear regression and Nash-Sutcliffe linear regression) for forecasting daily streamflow. The best performance for each metric is indicated in bold.

| Performance metric | Regression method | | |
|---|---|---|---|
| | One-dimensional linear regression | Multi-dimensional linear regression | Nash-Sutcliffe linear regression |
| Training set performance | | | |
| Realized Euclidean norm loss | 3.6082 | **3.2535** | 3.5098 |
| Realized Nash-Sutcliffe loss | 0.3180 | 0.2057 | **0.1288** |
| Test set performance | | | |
| Realized Euclidean norm loss | 2.6781 | **2.5359** | 2.6214 |
| Realized Nash-Sutcliffe loss | 0.3791 | 0.2244 | **0.1222** |

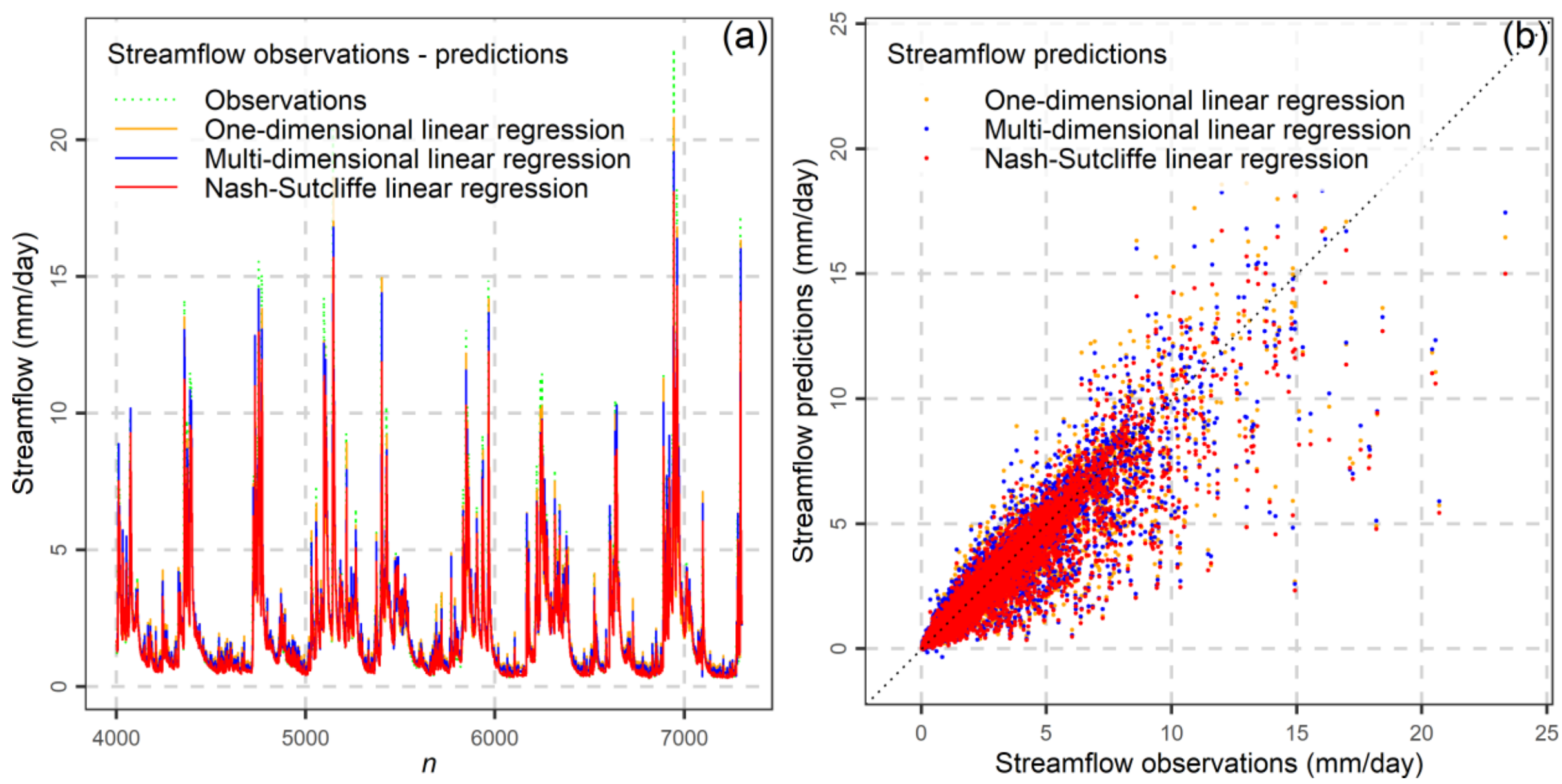

Figure 3. Streamflow forecasting results. (a) Observed daily streamflow (green dotted line) for the first catchment in the test set, plotted with predictions from one-dimensional linear regression (orange solid line), multi-dimensional linear regression (blue solid line) and Nash-Sutcliffe linear regression (red solid line). (b) Scatterplot of observed versus predicted daily streamflow for all 10 time series in the test set (3 303 time points each). Points are colored by method: One-dimensional linear regression (orange), multi-dimensional linear regression (blue) and Nash-Sutcliffe linear regression (red). The dotted line represents the 1: 1 diagonal.

*6.4.2 Temperature forecasting*

The same ten river basins are used and the temporal coverage is identical. Temperature series exhibit strong seasonality. Table 5 summarizes the performances.

(i) Realized Euclidean norm loss: As before, the multi-dimensional linear regression achieves the lowest Euclidean norm loss, underscoring its optimality for the component-wise mean. One-dimensional linear regression models perform considerably worse (about 34% higher Euclidean norm loss on the test set) because they cannot exploit the cross-time series information. Nash-Sutcliffe linear regression again incurs a small penalty in Euclidean norm loss ($\approx$ 6% higher than multi-dimensional linear regression on the test set), consistent with its different optimization target.

(ii) Realized Nash-Sutcliffe loss: Nash-Sutcliffe linear regression attains the lowest realized Nash-Sutcliffe loss, with a test-set value of 2.2500 compared to 2.4006 for multi-OLS and 3.5512 for one-dimensional linear regression models. The improvement over one-dimensional linear regression is about 37% and over multi-dimensional linear regression about 6%. The gains are less dramatic than for streamflow, possibly because temperature series are more Gaussian, bringing the Nash-Sutcliffe functional closer to the component-wise mean, as shown in experiments of Section 6.1. Nevertheless, Nash-Sutcliffe linear regression still offers a tangible benefit under the Nash-Sutcliffe evaluation metric, confirming that the theoretical distinction persists even in highly Gaussian settings.

Figure 4 (panel a) displays the observed and predicted temperature for the first time series. The scatterplot (panel b) shows that all three methods produce predictions that are highly correlated with observations. Predictions of all methods match observation better compared to illustrations of the corresponding experiment of Section 6.2.

Both applications consistently demonstrate that the choice of estimation method must align with the evaluation metric. When forecasts are to be evaluated by $\overline{\text{NSE}}$. the Nash-Sutcliffe linear regression introduced here is the optimal model. Using ordinary least squares, even in a multivariate, cross-series formulation, leads to inferior Nash-Sutcliffe performance.

Table 5. Performance comparison of three regression methods (one-dimensional linear regression, multi-dimensional linear regression and Nash-Sutcliffe linear regression) for forecasting daily temperature. The best performance for each metric is indicated in bold.

| | Regression method | | |
|---|---|---|---|
| Performance metric | One-dimensional linear regression | Multi-dimensional linear regression | Nash-Sutcliffe linear regression |
| Training set performance | | | |
| Realized Euclidean norm loss | 41.6731 | **31.3421** | 33.8090 |
| Realized Nash-Sutcliffe loss | 3.1818 | 2.2190 | **2.0990** |
| Test set performance | | | |
| Realized Euclidean norm loss | 43.8880 | **32.6507** | 34.7666 |
| Realized Nash-Sutcliffe loss | 3.5512 | 2.4006 | **2.2500** |

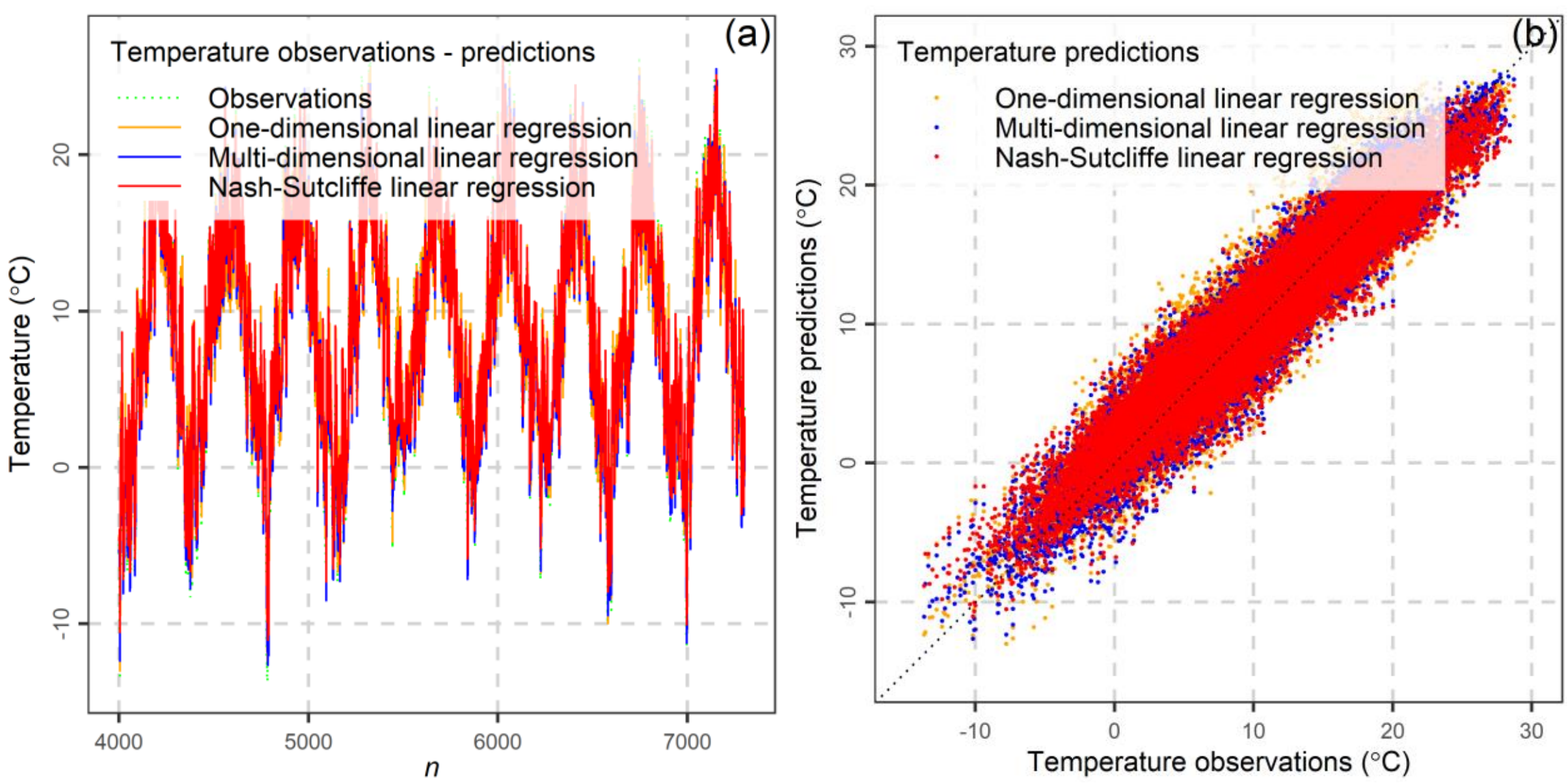

Figure 4. Temperature forecasting results. (a) Observed daily temperature (green dotted line) for the first time series in the test set, plotted with predictions from one-dimensional linear regression (orange solid line), multi-dimensional linear regression (blue solid line) and Nash-Sutcliffe linear regression (red solid line). (b) Scatterplot of observed versus predicted daily temperature for all 10 time series in the test set (3 303 time points each). Points are colored by method: One-dimensional linear regression (orange), multi-dimensional linear regression (blue) and Nash-Sutcliffe linear regression (red). The dotted line represents the 1: 1 diagonal.

## 7. Discussion

We examine the connections of the framework with the extant literature in Section 7.1, the implications of the two distinct data settings in Section 7.2, the methodological parallels with other fields in Section 7.3 and a practical guide for applying the NSE in Section 7.4.

### 7.1 Discussions related to NSE empirical investigations

The theoretical framework developed in this paper directly addresses the empirical observations and critiques that have emerged in the geosciences, most notably Williams (2025), who evaluates streamflow predictions from river basins across a large spatial domain (the contiguous United States). Williams (2025) observed that NSE values vary widely across sites even when models have similar MSE. We have shown that this is a direct consequence of the NSE eliciting a different functional than the MSE. Specifically, the weight $w(\boldsymbol{y}_d)$ in the Nash-Sutcliffe loss makes the metric sensitive to the variability of each individual time series.

Williams (2025) further argued that cross-site comparisons of NSE values are statistically invalid. Our framework makes this point explicit. Specifically, evaluating forecasts with the NSE across multiple sites is statistically valid only when the time series are assumed to be realizations of a single underlying stochastic process (Section 4.1). Consequently, the observed spatial patterns in NSE across a large sample reflect not only differential model skill but also the differing stochastic properties of the time series themselves.

Finally, Williams (2025) observed that training a model with MSE, while evaluating it with NSE is inconsistent in the practical settings he considers. This inconsistency represents the central practical implication of our work. The strict consistency framework (Section 4.1) demonstrates that minimizing the MSE during training, targets the conditional mean, whereas evaluating with the NSE targets the Nash-Sutcliffe functional. These are generally different quantities. Nash-Sutcliffe regression (Sections 4.8.3, 5.5.3) constitutes the correct estimation method when the evaluation metric is the NSE, thereby aligning the functional elicited during training with that used for evaluation. The framework also clarifies why comparing NSE values across variables of different nature (e.g., streamflow and temperature, Sections 6.4.1, 6.4.2) is problematic, as such variables arise from stochastic processes with distinct properties.

The theoretical developments presented here can be readily extended to NSE variants that employ bijective transformations of both predictions and realizations, as frequently examined in the literature (Thirel et al. 2024). These variants target different functionals than the Nash-Sutcliffe functional. The statistical machinery required to identify these functionals is developed in Tyralis and Papacharalampous (2026).

### 7.2 Discussion of the different settings

A key contribution of this work is the analysis of the two distinct data settings in which the Nash-Sutcliffe loss can be applied. The $d \times n$ in Section 4 and $n \times d$ in Section 5 orientations, while structurally dual, represent fundamentally different assumptions and modeling tasks. For intuition, the following discussion assumes that time series are observed at different spatial locations; however, the implications remain equivalent regardless of whether observations correspond to spatial locations or simply to different time series.

The $d \times n$ setting is the one commonly applied in the environmental sciences literature, particularly in large-sample hydrology (Perrin et al. 2001). Here, the $n$ columns of the realization matrix $\boldsymbol{Y}_{d\times n}$ represent $n$ different spatial locations, each containing a $d$-dimensional time series. The core assumption is that each of the $n$ time series constitutes a realization of the same $d$-dimensional stochastic process. Consequently, while marginal distributions might differ across time steps, they should remain stationary across space, meaning that the $d$-dimensional time series comprises $d$ variables with potentially different marginal distributions. This is a strong assumption, implying that the underlying dynamics and marginal distributions at each time step are identical across all locations. In this setting, the Nash-Sutcliffe linear regression introduced in Section 4.8.3 issues predictions of a $d$-dimensional functional (the entire time series) for a new spatial location. As proved in Section 5.6, specifying a multi-dimensional linear model to forecast new time steps, while optimizing the standard realized Nash-Sutcliffe loss (as in the models of Section 5.5), gives results equivalent to those from a multi-dimensional linear regression, trained with the Euclidean norm loss. This equivalence arises because the model's parameters are not shared across locations, thereby restricting its ability to target the Nash-Sutcliffe functional directly. Deep learning models, in contrast, can share parameters across locations and may therefore prove more useful in forecasting applications that employ the standard realized Nash-Sutcliffe loss.

The $n \times d$ setting of Section 5 represents the natural forecasting setup. Here, $n$ represents the time series length, while $d$ represents the number of distinct variables or series observed, possibly at different locations, within the realizations matrix $\boldsymbol{Y}_{n\times d}$. The underlying assumption is that each row of the observation matrix $\boldsymbol{Y}_{n\times d}$ is a realization of a $d$-dimensional random vector. In this configuration, the model is trained to predict a $d$-dimensional outcome at time $n$ from a set of predictors available today at time $n-1$. This setting allows the properties of stochastic processes to differ across locations, i.e. marginal distributions may vary spatially, while remaining stationary over time at each location. This framework aligns with standard MSE-based evaluations for a single location, which assume that the stochastic process distribution remains stationary through time.

The two settings give different realized Nash-Sutcliffe losses, as expressed in eqs. (4.1) and (5.2). This divergence arises from the asymmetry induced by the weight function. In contrast, the realized Euclidean loss remains invariant to data orientation due to its symmetry, as shown in eqs. (3.33) and (5.1). Consequently, evaluating with the realized Euclidean norm loss, implicitly assumes that either of the two earlier assumptions could serve as the basis for evaluation.

## 7.3 Connections to weighted least squares regression

The Nash-Sutcliffe regression formulation takes the form of multivariate weighted least squares (Gentle 2024, p.452), but with a crucial distinction. The weights are not fixed; rather they are data-dependent. This fundamental insight places Nash-Sutcliffe linear regression at the intersection of several statistical methodologies. In standard weighted least squares (WLS), the weights are known constants and the objective is to align estimation with the Euclidean norm loss. By contrast, Nash-Sutcliffe linear regression constitutes a weighted least-squares estimator where the weighting formulation emerges from the data.

## 7.4 Practical guide on how to use the NSE

The theoretical developments in this paper enable the formulation of a practical guide for the proper use of the NSE and the $\overline{\text{NSE}}$:

(i) When evaluating predictions for a single time series, treat the NSE as a skill score, not an accuracy metric. In this case, the NSE quantifies the improvement of a model's predictions over a simple benchmark that always predicts the mean climatology of the observations. A positive NSE indicates that the model outperforms the mean benchmark; a value of 0 means the model is as accurate as using the mean climatology, while a negative value suggests the model's predictions are less accurate than this naïve benchmark.

(ii) When working with the average NSE ($\overline{\text{NSE}}$), only compare or average NSE across locations if all time series are assumed to be realizations of the same underlying stochastic process. In practice, this requires grouping time series according to their stochastic properties, thereby excluding the combination of datasets such as daily and monthly records or data originating from distinct stochastic processes. Interpret spatial patterns in individual NSE values as purely a map of model skill, only under the assumption that the series share similar stochastic properties.

(iii) Choose between the realized Nash-Sutcliffe loss as defined in eq. (1.10) or in eq. (5.2), depending on the assumptions of the stochastic process generating the data. The first orientation assumes that the observed time series are realizations of a single stochastic process, while the second assumes that the vector of realizations at each time step is generated from a single stochastic process, with the marginal properties of each time series remaining stationary over time. The second orientation is the implicit assumption when evaluating forecasts for a single time series using the MSE.

(iv) If you evaluate with $\overline{\text{NSE}}$, you must train your model using the Nash-Sutcliffe loss. Employing Nash-Sutcliffe linear regression (or an equivalent machine learning model optimized with a Nash-Sutcliffe loss function) is essential to align the estimation procedure with the evaluation metric. Training separate models with the Euclidean norm loss targets the component-wise mean functional and leads to suboptimal predictions under the $\overline{\text{NSE}}$, unless specific conditions are met (e.g. intependent Gaussian variables).

(v) Use the extended Nash-Sutcliffe loss to handle near-zero variance. In practice, if the generated vector realizations have zero or near-zero variance, the denominator of the Nash-Sutcliffe loss approaches 0, causing numerical instability. Adding a small constant $a$ (as in eq. (4.10)) mitigates this issue while retaining the core properties of the loss function. In this case, however, the target becomes the extended Nash-Sutcliffe functional.

## 8. Conclusions

We establish a formal decision-theoretic foundation for the Nash-Sutcliffe efficiency (NSE), a commonly used metric in multi-series forecast evaluation. By framing multi-series forecast evaluation within the theory of strictly consistent loss functions, we demonstrate that comparing models via averaging the NSE of forecasts for multiple time series is equivalent to evaluating predictions of a multi-dimensional functional, which we call Nash-Sutcliffe functional, using the Nash-Sutcliffe loss (the negatively oriented counterpart of NSE).

A primary contribution of this work is the rigorous characterization of the Nash-Sutcliffe functional, which takes the form of a data-weighted component-wise mean and differs fundamentally from the component-wise mean, the latter elicited by Bregman loss functions. We prove that the Nash-Sutcliffe loss is strictly consistent for the Nash-Sutcliffe functional, and we further demonstrate that this functional is identifiable. This theoretical distinction explains empirical findings in the literature; for example it allows a mathematical explanation for why model rankings diverge when using the NSE versus the mean squared error (MSE), despite the latter being a fundamental component of the former.

Furthermore, we bridge the gap between model estimation and evaluation by introducing the Nash-Sutcliffe linear regression framework. A key finding is the necessity of alignment. If a model is to be evaluated using the NSE, it must be trained using the Nash-Sutcliffe loss. Training with the MSE or more generally with Bregman loss functions, targets a different statistical functional, namely the component-wise mean, leading to suboptimal predictive performance under the intended evaluation metric.

The practical implications are demonstrated through diverse applications, involving both simulated and real-world data, which confirm that the proposed regression approach improves performance in multi-series forecasting tasks. Additionally, we outline essential guidance for practitioners, including the use of an extended loss version to handle numerical instabilities and the requirement that averaged series share a common underlying stochastic process for statistical validity. These findings offer a principled methodology for theoretically sound model estimation and evaluation in disciplines that employ relative measures.

## Appendix A Vector and matrix notation

This appendix defines the vector and matrix operations, norms, and statistical functions used throughout the manuscript.

### Basic vector definitions

Vector notation:

$$\boldsymbol{x}_n = (x_1, \dots, x_n)^{\mathrm{T}} \tag{A.1}$$

where the superscript $^{\mathrm{T}}$ indicates the transpose.

Zero vector (all elements zero):

$$\boldsymbol{0}_n = (0, \dots, 0)^{\mathrm{T}} \tag{A.2}$$

All-ones vector (all elements unity):

$$\boldsymbol{1}_n = (1, \dots, 1)^{\mathrm{T}} \tag{A.3}$$

### Element-wise operations on vectors

Element-wise comparison (holds for all $i \in \{1, \dots, n\}$):

$$\boldsymbol{x}_n > \boldsymbol{y}_n \Leftrightarrow x_i > y_i \text{ (similarly for } \geq, =, \leq, <) \tag{A.4}$$

### Basic matrix definitions

Matrix notation:

$$\boldsymbol{X}_{n \times m} = [x_{ij}], i = 1, \dots, n, j = 1, \dots, m \tag{A.5}$$

Matrix row:

$$\boldsymbol{X}_{i,\cdot} = (x_{i1}, \ldots, x_{im}), i = 1, \ldots, n \quad (A.6)$$

Matrix column:

$$\boldsymbol{X}_{\cdot,j} = (x_{1j}, \ldots, x_{nj})^{\mathrm{T}}, j = 1, \ldots, m \quad (A.7)$$

Zero matrix (all elements zero):

$$\boldsymbol{0}_{n\times m} = [0] \quad (A.8)$$

Identity matrix ($\delta_{ij}$ is the Kronecker delta):

$$\boldsymbol{I}_n = [\delta_{ij}], \delta_{ij} = \begin{cases} 1, 1 \leq i = j \leq n \\ 0, \text{otherwise} \end{cases} \quad (A.9)$$

Diagonal matrix:

$$\operatorname{diag}(\boldsymbol{x}_n) = \begin{bmatrix} x_1 & \cdots & 0 \\ \vdots & \ddots & \vdots \\ 0 & \cdots & x_n \end{bmatrix} \quad (A.10)$$

**Inner products and norms**

Euclidean inner product:

$$\langle \boldsymbol{x}_n, \boldsymbol{y}_n \rangle := \boldsymbol{x}_n^{\mathrm{T}} \boldsymbol{y}_n = \sum_{i=1}^{n} x_i y_i = \langle \boldsymbol{y}_n, \boldsymbol{x}_n \rangle \quad (A.11)$$

Euclidean norm:

$$\|\boldsymbol{x}_n\|_2 = \left(\sum_{i=1}^{n} x_i^2\right)^{1/2} \text{ (Euclidean norm)} \quad (A.12)$$

Squared Euclidean norm:

$$\|\boldsymbol{x}_n\|_2^2 = \langle \boldsymbol{x}_n, \boldsymbol{x}_n \rangle = \boldsymbol{x}_n^{\mathrm{T}} \boldsymbol{x}_n = \sum_{i=1}^{n} x_i^2 \quad (A.13)$$

Frobenius norm:

$$\|\boldsymbol{X}_{n\times m}\|_{\mathrm{F}} = \left(\sum_{i=1}^{n} \sum_{j=1}^{m} x_{ij}^2\right)^{1/2} \quad (A.14)$$

**Kronecker product**

Kronecker product is the $nk \times ml$ matrix (Gentle 2024, p.116):

$$\boldsymbol{X}_{n\times m} \otimes \boldsymbol{Y}_{k\times l} = \begin{bmatrix} x_{11}\boldsymbol{Y}_{kl} & \cdots & x_{1m}\boldsymbol{Y}_{kl} \\ \vdots & \ddots & \vdots \\ x_{n1}\boldsymbol{Y}_{kl} & \cdots & x_{nm}\boldsymbol{Y}_{kl} \end{bmatrix} \quad (A.15)$$

Transpose of a Kronecker product is the $ml \times nk$ matrix (Gentle 2024, p.117):

$$(\boldsymbol{X}_{n\times m} \otimes \boldsymbol{Y}_{k\times l})^{\mathrm{T}} = \boldsymbol{X}_{n\times m}^{\mathrm{T}} \otimes \boldsymbol{Y}_{k\times l}^{\mathrm{T}} \quad (A.16)$$

Product of Kronecker products is the $nk \times nk$ matrix (Gentle 2024, p.117):

$$(\boldsymbol{X}_{n\times m} \otimes \boldsymbol{Y}_{k\times l})(\boldsymbol{Z}_{m\times n} \otimes \boldsymbol{W}_{l\times k}) = (\boldsymbol{X}_{n\times m}\boldsymbol{Z}_{m\times n}) \otimes (\boldsymbol{Y}_{k\times l}\boldsymbol{W}_{l\times k}) \quad (A.17)$$

Inverse of a Kronecker product is the $nm \times nm$ matrix (Gentle 2024, p.140):

$$(\boldsymbol{X}_{n\times n} \otimes \boldsymbol{Y}_{m\times m})^{-1} = \boldsymbol{X}_{n\times n}^{-1} \otimes \boldsymbol{Y}_{m\times m}^{-1} \tag{A.18}$$

**Matrix vectorization**

The vectorization of an $n \times m$ matrix $\boldsymbol{X}_{n\times m}$ is the $nm$-dimensional column vector by stacking its columns (Gentle 2024, p.80):

$$\text{vec}(\boldsymbol{X}_{n\times m}) := (x_{11}, \dots, x_{n1}, x_{12}, \dots, x_{n2}, x_{1m}, \dots, x_{nm})^{\text{T}} \tag{A.19}$$

For a product of three matrices, vectorization gives an $np$-dimensional column vector (Gentle 2024, p.118):

$$\text{vec}(\boldsymbol{X}_{n\times m}\boldsymbol{Y}_{m\times l}\boldsymbol{Z}_{l\times p}) = (\boldsymbol{Z}_{l\times p}^{\text{T}} \otimes \boldsymbol{X}_{n\times m})\text{vec}(\boldsymbol{Y}_{m\times l}) \tag{A.20}$$

**Vector and matrix derivatives:**

This section presents the conventions for vector and matrix derivatives used throughout the manuscript, following numerator-layout notation. All derivatives are defined with respect to real-valued scalars, vectors and matrices:

Derivative of a scalar with respect to a vector: For a scalar $y$ that is a function of the vector $\boldsymbol{x}_n$, the derivative is the row vector:

$$\frac{\partial y}{\partial \boldsymbol{x}_n} = (\frac{\partial y}{\partial x_1}, \dots, \frac{\partial y}{\partial x_n}) \tag{A.21}$$

Derivative of a vector with respect to a scalar: For a vector $\boldsymbol{y}_n$ that is a function of a scalar $x$, the derivative is the column vector:

$$\frac{\partial \boldsymbol{y}_n}{\partial x} = (\frac{\partial y_1}{\partial x}, \dots, \frac{\partial y_n}{\partial x})^{\text{T}} \tag{A.22}$$

Derivative of a vector with respect to a vector: For a vector $\boldsymbol{y}_m$ that is a function of the vector $\boldsymbol{x}_n$, the derivative is the $m \times n$ matrix:

$$\frac{\partial \boldsymbol{y}_m}{\partial \boldsymbol{x}_n} = \begin{bmatrix} \frac{\partial y_1}{\partial x_1} & \cdots & \frac{\partial y_1}{\partial x_n} \\ \vdots & \ddots & \vdots \\ \frac{\partial y_m}{\partial x_1} & \cdots & \frac{\partial y_m}{\partial x_n} \end{bmatrix} \tag{A.23}$$

Let $f$ be a scalar-valued function of the vector $\boldsymbol{x}_n = (x_1, \dots, x_n)^{\text{T}}$.

Gradient of $f$ (Gentle 2024, p.330): The gradient is the column vector

$$\nabla_{\boldsymbol{x}_n} f(\boldsymbol{x}_n) = (\frac{\partial f(\boldsymbol{x}_n)}{\partial \boldsymbol{x}_n})^{\text{T}} = (\frac{\partial f(\boldsymbol{x}_n)}{\partial x_1}, \dots, \frac{\partial f(\boldsymbol{x}_n)}{\partial x_n})^{\text{T}} \tag{A.24}$$

Hessian of $f$ (Gentle 2024, p.334): The Hessian matrix is the $n \times n$ symmetric matrix of second-order partial derivatives, which can be expressed as

$$\boldsymbol{H}_{x_n} f(\boldsymbol{x}_n) = \frac{\partial \nabla_{x_n} f(x_n)}{\partial x_n} = \begin{bmatrix} \frac{\partial^2 f(x_n)}{\partial x_1^2} & \cdots & \frac{\partial^2 f(x_n)}{\partial x_1 \, \partial x_n} \\ \vdots & \ddots & \vdots \\ \frac{\partial^2 f(x_n)}{\partial x_n \, \partial x_1} & \cdots & \frac{\partial^2 f(x_n)}{\partial x_n^2} \end{bmatrix} \quad \text{(A.25)}$$

Derivative of a Euclidean inner product: For vectors $\boldsymbol{x}_n$ and $\boldsymbol{y}_n$, where $\boldsymbol{y}_n$ is not function of $\boldsymbol{x}_n$,

$$\frac{\partial x_n^{\mathrm{T}} y_n}{\partial x_n} = \frac{\partial y_n^{\mathrm{T}} x_n}{\partial x_n} = \boldsymbol{y}_n^{\mathrm{T}} \quad \text{(A.26)}$$

Derivative of a quadratic form: For a symmetric matrix $\boldsymbol{Y}_{n\times n}$ that is not a function of $\boldsymbol{x}_n$,

$$\frac{\partial x_n^{\mathrm{T}} Y_{nn} x_n}{\partial x_n} = 2\boldsymbol{x}_n^{\mathrm{T}} \boldsymbol{Y}_{n\times n} \quad \text{(A.27)}$$

Derivative of a matrix-vector product: For a matrix $\boldsymbol{Y}_{m\times n}$ that is not a function of $\boldsymbol{x}_n$,

$$\frac{\partial Y_{mn} x_n}{\partial x_n} = \boldsymbol{Y}_{m\times n} \quad \text{(A.28)}$$

**<u>Indicator functions</u>**

Indicator function:

$$\mathbb{1}_A\{x\} := \begin{cases} 1, x \in A \\ 0, \text{otherwise} \end{cases} \quad \text{(A.29)}$$

**<u>Statistical functions for vectors</u>**

Sample mean:

$$\mu(\boldsymbol{x}_n) := (1/n)\mathbf{1}_n^{\mathrm{T}} \boldsymbol{x}_n = (1/n) \sum_{i=1}^{n} x_i = (1/n)\langle \boldsymbol{x}_n, \mathbf{1}_n \rangle \quad \text{(A.30)}$$

**<u>Centered vectors and key identities:</u>**

Centered vector (mean-zero):

$$\boldsymbol{x}_{nc} := \boldsymbol{x}_n - \mathbf{1}_n \mu(\boldsymbol{x}_n) \quad \text{(A.31)}$$

Orthogonality to the ones vector:

$$\langle \boldsymbol{x}_{nc}, \mathbf{1}_n \rangle = 0 \quad \text{(A.32)}$$

## Appendix B Probability distributions

This appendix defines the probability density functions (PDFs) for the probability distributions referenced in the main text.

(i) Univariate Gaussian (normal) distribution: The PDF of a Gaussian random variable $\underline{y} \sim N(\mu, \sigma)$ is defined as:

$$f_N(y; \mu, \sigma) := \frac{1}{\sigma\sqrt{2\pi}} \exp\left(-\frac{(y-\mu)^2}{2\sigma^2}\right), y \in \mathbb{R}, \mu \in \mathbb{R}, \sigma > 0 \quad \text{(B.1)}$$

(ii) Multivariate Gaussian (normal) distribution: The PDF of a multivariate Gaussian random variable $\underline{\boldsymbol{y}}_d \sim N_d(\boldsymbol{\mu}_d, \boldsymbol{\Sigma}_{d\times d})$ is defined as (Eaton 1983, Proposition 3.2):

$$f_{N_d}(\boldsymbol{y}_d; \boldsymbol{\mu}_d, \boldsymbol{\Sigma}_{d\times d}) := (2\pi)^{-d/2}(\det(\boldsymbol{\Sigma}_{d\times d}))^{-1/2}\exp(-(1/2)(\boldsymbol{y}_d - \boldsymbol{\mu}_d)^{\mathrm{T}}\boldsymbol{\Sigma}_{d\times d}^{-1}(\boldsymbol{y}_d - \boldsymbol{\mu}_d)), \boldsymbol{y}_d \in \mathbb{R}^d, \boldsymbol{\mu}_d \in \mathbb{R}^d, \boldsymbol{\Sigma}_{d\times d} \text{ positive definite} \quad \text{(B.2)}$$

(iii) Chi-squared distribution: The PDF of a chi-squared random variable $\underline{y} \sim \chi^2(k)$ with $k$ degrees of freedom is defined as:

$$f_{\chi^2}(y; k) := \frac{1}{2^{k/2}\Gamma(k/2)} y^{(k/2)-1}\exp(-y/2), y > 0, k \in \mathbb{N}\backslash\{0\} \quad \text{(B.3)}$$

For independent standard Gaussian random variables $\underline{y}_i \sim N(0,1)$, the sum of their squares follows a chi-squared distribution: $\sum_{i=1}^{n} \underline{y}_i^2 \sim \chi^2(n)$.

(iv) Inverse chi-squared distribution: The PDF of an inverse chi-squared random variable $\underline{y} \sim \mathrm{Inv} - \chi^2(k)$ with $k$ degrees of freedom is defined as:

$$f_{\mathrm{Inv}-\chi^2}(y; k) := \frac{2^{-k/2}}{\Gamma(k/2)} y^{-(k/2+1)}\exp(-1/(2y)), y > 0, k \in \mathbb{N}\backslash\{0\} \quad \text{(B.4)}$$

If $\underline{y} \sim \chi^2(k)$, then $1/\underline{y} \sim \mathrm{Inv} - \chi^2(k)$.

(v) Gamma distribution: The PDF of a Gamma distributed random variable $\underline{y} \sim \mathrm{Gamma}(\theta, a)$ is defined as:

$$f_{\mathrm{Gamma}}(y; \theta, a) := \frac{1}{\theta^a\Gamma(a)} y^{a-1}\exp(-y/\theta), y > 0, \theta > 0, a > 0 \quad \text{(B.5)}$$

The chi-squared distribution is a special case of the Gamma distribution for $\theta = 2$, $a = k/2$.

(vi) Inverse Gamma distribution: The PDF of an inverse Gamma distributed random variable $\underline{y} \sim \mathrm{Inv} - \mathrm{Gamma}(\theta, a)$ is defined as (Bernardo and Smith 1994, Appendix A.1):

$$f_{\mathrm{Inv-Gamma}}(y; \theta, a) := \frac{1}{\theta^a\Gamma(a)} y^{-(a+1)}\exp(-1/(\theta y)), y > 0, \theta > 0, a > 0 \quad \text{(B.6)}$$

The inverse chi-squared distribution is a special case of the inverse Gamma distribution for $\theta = 2, a = k/2$.

(vii) Square-root inverse Gamma distribution: The PDF of a square-root inverted Gamma distributed random variable $\underline{y} \sim \mathrm{Inv} - \mathrm{Gamma}^{-1/2}(\theta, a)$ is defined as (Bernardo and Smith 1994, Appendix A.1):

$$f_{\mathrm{Inv-Gamma}^{-1/2}}(y; \theta, a) := \frac{2}{\theta^a\Gamma(a)} y^{-(2a+1)}\exp(-1/(\theta y^2)), y > 0, \theta > 0, a > 0 \quad \text{(B.7)}$$

If $\underline{y} \sim \text{Gamma}(\theta, a)$, then $1/\underline{y} \sim \text{Inv} - \text{Gamma}(\theta, a)$ and $1/\underline{y}^{1/2} \sim \text{Inv} - \text{Gamma}^{-1/2}(\theta, a)$. Furthermore, if $\underline{y} \sim \chi^2(k) = \text{Gamma}(2, k/2)$, then $1/\underline{y} \sim \text{Inv} - \text{Gamma}(2, k/2)$ and $1/\underline{y}^{1/2} \sim \text{Inv} - \text{Gamma}^{-1/2}(2, k/2)$.

(viii) Log-normal distribution: The PDF of a log-normally distributed random variable $\underline{y} \sim \text{Lognormal}(\mu, \sigma)$ is defined as:

$$f_{\text{Lognormal}}(y; \mu, \sigma) := \frac{1}{y\sigma\sqrt{2\pi}} \exp\left(-\frac{(\log(y) - \mu)^2}{2\sigma^2}\right), y > 0, \mu \in \mathbb{R}, \sigma > 0 \tag{B.8}$$

## Appendix C Proofs

**Proof C.1** (Strict consistency of Nash-Sutcliffe loss function):

We present two distinct proofs. The first is a direct application of Theorem 1, while the second employs convex analysis to explain the mechanism underlying the loss function's consistency. Both proofs operate under the following assumption:

Let $\mathcal{F}_d^{(w)} \subseteq \mathcal{F}_{\text{mean},d}$ be the subclass of distributions in $\mathcal{F}_{\text{mean},d}$ for which $w(\boldsymbol{y}_d) f(\boldsymbol{y}_d)$ has finite integral over $\mathbb{R}^d, d \geq 0$, $w(\boldsymbol{y}_d) = 1/\|\mu(\boldsymbol{y}_d)\mathbf{1}_d - \boldsymbol{y}_d\|_2^2$ from eq. (1.9), and for which the probability measure with density proportional to $w(\boldsymbol{y}_d) f(\boldsymbol{y}_d)$ also belongs to $\mathcal{F}_{\text{mean},d}$.

**Strict consistency of the Nash-Sutcliffe loss via Theorem 1:**

As established in Section 3.7.2, the function $L_{\text{EN}}$ is strictly $\mathcal{F}_{\text{mean},d}$-consistent for the functional $\boldsymbol{T}_d(F) = \mathbb{E}_F[\underline{\boldsymbol{y}}_d]$.

The density $f^{(w)}(\boldsymbol{y}_d)$ is proportional to $w(\boldsymbol{y}_d) f(\boldsymbol{y}_d)$, and is explicitly given by

$$f^{(w)}(\boldsymbol{y}_d) = w(\boldsymbol{y}_d) f(\boldsymbol{y}_d) / \mathbb{E}_F[w(\boldsymbol{y}_d)] \tag{C.1}$$

Therefore, by Theorem 1, the loss function $L_{\text{NS}}$ is strictly $\mathcal{F}_d^{(w)}$-consistent for the functional $\boldsymbol{T}_d^{(w)}(F) = \boldsymbol{T}_d\left(F^{(w)}\right) = \left(\mathbb{E}_{F^{(w)}}[\underline{y}_1], \dots, \mathbb{E}_{F^{(w)}}[\underline{y}_d]\right)^{\text{T}}$.

We now write each component of $\boldsymbol{T}_d^{(w)}(F)$ using the original distribution $F$. For $k = 1, \dots, d$:

$$\mathbb{E}_{F^{(w)}}[\underline{y}_k] = \int_{\mathbb{R}^d} y_k f^{(w)}(\boldsymbol{y}_d) \mathrm{d}\boldsymbol{y}_d \tag{C.2}$$

Substituting the expression for $f^{(w)}$ from eq. (C.1) gives:

$$\mathbb{E}_{F^{(w)}}[\underline{y}_k] = (1/\mathbb{E}_F[w(\boldsymbol{y}_d)]\,) \int_{\mathbb{R}^d} y_k w(\boldsymbol{y}_d) f(\boldsymbol{y}_d)(\boldsymbol{y}_d) \mathrm{d}\boldsymbol{y}_d \tag{C.3}$$

This is equivalent to:

$$\mathbb{E}_{F^{(w)}}[\underline{y}_k] = \mathbb{E}_F[\underline{y}_k w(\underline{\boldsymbol{y}}_d)]/\mathbb{E}_F[w(\underline{\boldsymbol{y}}_d)], k = 1, \dots, d \tag{C.4}$$

Consequently, the functional $\boldsymbol{T}_d^{(w)}(F)$ is given by:,

$$\boldsymbol{T}_d^{(w)}(F) = (\mathbb{E}_F[\underline{y}_1 w(\underline{\boldsymbol{y}}_d)]/\mathbb{E}_F[w(\underline{\boldsymbol{y}}_d)], \dots, \mathbb{E}_F[\underline{y}_d w(\underline{\boldsymbol{y}}_d)]/\mathbb{E}_F[w(\underline{\boldsymbol{y}}_d)])^{\mathrm{T}} \tag{C.5}$$

**Strict consistency of the Nash-Sutcliffe loss via direct convexity analysis**:

To establish the strict consistency of the loss function $L_{\mathrm{NS}}$, we analyze the expectation of the loss with respect to a distribution $F \in \mathcal{F}_d^{(w)}$

$$\mathbb{E}_F[L_{\mathrm{NS}}(\boldsymbol{z}_d, \underline{\boldsymbol{y}}_d)] = \int_{\mathbb{R}^d} w(\boldsymbol{y}_d) ||\boldsymbol{z}_d - \boldsymbol{y}_d||_2^2 f(\boldsymbol{y}_d) \mathrm{d}\boldsymbol{y}_d \tag{C.6}$$

We proceed by examining the first and second derivatives of this expected loss:

The first partial derivative with respect to a component $z_k$ is:

$$\frac{\partial \mathbb{E}_F[L_{\mathrm{NS}}(\boldsymbol{z}_d, \underline{\boldsymbol{y}}_d)]}{\partial z_k} = 2 \int_{\mathbb{R}^d} w(\boldsymbol{y}_d)(z_k - y_k) f(\boldsymbol{y}_d) \mathrm{d}\boldsymbol{y}_d \, , k = 1, \dots, d \tag{C.7}$$

The second-order partial derivatives are:

$$\frac{\partial^2 \mathbb{E}_F[L_{\mathrm{NS}}(\boldsymbol{z}_d, \underline{\boldsymbol{y}}_d)]}{\partial z_k^2} = 2 \int_{\mathbb{R}^d} w(\boldsymbol{y}_d) f(\boldsymbol{y}_d) \mathrm{d}\boldsymbol{y}_d = 2\mathbb{E}_F[w(\underline{\boldsymbol{y}}_d)] > 0, k = 1, \dots, d \tag{C.8}$$

The cross-derivatives are:

$$\frac{\partial^2 \mathbb{E}_F[L_{\mathrm{NS}}(\boldsymbol{z}_d, \underline{\boldsymbol{y}}_d)]}{\partial z_k z_l} = 0, k, l = 1, \dots, d, k \neq l \tag{C.9}$$

The function $\mathbb{E}_F[L_{\mathrm{NS}}(\boldsymbol{z}_d, \underline{\boldsymbol{y}}_d)]$ is twice continuously differentiable for all $\boldsymbol{z}_d \in \mathbb{R}^d$. Its Hessian matrix is diagonal with strictly positive entries $2\mathbb{E}_F[w(\underline{\boldsymbol{y}}_d)]$ on the diagonal and is therefore positive definite for all $\boldsymbol{z}_d \in \mathbb{R}^d$.

This implies that $\mathbb{E}_F[L_{\mathrm{NS}}(\boldsymbol{z}_d, \underline{\boldsymbol{y}}_d)]$ is strictly convex over $\mathbb{R}^d$ (Rockafellar 1970, Theorem 4.5, Boyd and Vandenberghe 2004, p. 71). Strictly convex functions possess at most one critical point, which, if it exists, is the unique global minimizer (Boyd and Vandenberghe 2004, p. 69).

To find this minimizer, we set the gradient in eq. (C.7) to 0. For each $k$ this gives:

$$2 \int_{\mathbb{R}^d} w(\boldsymbol{y}_d)(z_k - y_k) f(\boldsymbol{y}_d) \mathrm{d}\boldsymbol{y}_d = 0 \tag{C.10}$$

Solving for $z_k$, gives the unique critical point:

$$T_k = \mathbb{E}_F[\underline{y}_k w(\underline{\boldsymbol{y}}_d)]/\mathbb{E}_F[w(\underline{\boldsymbol{y}}_d)], k = 1, \dots, d \tag{C.11}$$

Therefore, $\mathbb{E}_F[L_{\text{NS}}(z_d, \underline{y}_d)]$ has a unique global minimum at $T_d = (T_1, \dots, T_d)^{\text{T}}$. By Definition 1, this proves that $L_{\text{NS}}$ is strictly $\mathcal{F}_d^{(w)}$-consistent for the functional $\boldsymbol{T}_d$.■

**Proof C.2** (Consistency of Nash-Sutcliffe $M$-estimator):

To prove the consistency of the $M$-estimator $\widehat{\boldsymbol{\theta}}_d(\underline{\boldsymbol{Y}}_{d\times n})$ defined in eq. (4.8), we start from its form in eq. (3.22):

$$\widehat{\boldsymbol{\theta}}_d(\boldsymbol{Y}_{d\times n}) := \underset{\boldsymbol{\theta}_d \in \mathbb{R}^d}{\arg\min}\, \bar{L}_{\text{NS}}(\boldsymbol{\theta}_d \mathbf{1}_n^{\text{T}}, \boldsymbol{Y}_{d\times n}) \tag{C.12}$$

From eqs. (1.7) and (1.10), the realized loss $\bar{L}_{\text{NS}}$ takes the form:

$$\bar{L}_{\text{NS}}(\boldsymbol{\theta}_d \mathbf{1}_n^{\text{T}}, \boldsymbol{Y}_{d\times n}) = (1/n) \sum_{j=1}^{n} w(\boldsymbol{Y}_{\cdot,j}) ||\boldsymbol{\theta}_d - \boldsymbol{Y}_{\cdot,j}||_2^2 \tag{C.13}$$

To find the minimizer, we analyze the critical points of this function. The first partial derivative with respect to a component $\theta_k$ is:

$$\frac{\partial \bar{L}_{\text{NS}}(\boldsymbol{\theta}_d \mathbf{1}_n^{\text{T}}, \boldsymbol{Y}_{d\times n})}{\partial \theta_k} = (2/n) \sum_{j=1}^{n} w(\boldsymbol{Y}_{\cdot,j})(\theta_k - y_{kj}), k = 1, \dots, d \tag{C.14}$$

The second-order partial derivatives are:

$$\frac{\partial^2 \bar{L}_{\text{NS}}(\boldsymbol{\theta}_d \mathbf{1}_n^{\text{T}}, \boldsymbol{Y}_{d\times n})}{\partial \theta_k^2} = (2/n) \sum_{j=1}^{n} w(\boldsymbol{Y}_{\cdot,j}) > 0, k = 1, \dots, d \tag{C.15}$$

The cross-derivatives are:

$$\frac{\partial^2 \bar{L}_{\text{NS}}(\boldsymbol{\theta}_d \mathbf{1}_n^{\text{T}}, \boldsymbol{Y}_{d\times n})}{\partial \theta_k \theta_l} = 0, k, l = 1, \dots, d, k \neq l \tag{C.16}$$

The function $\bar{L}_{\text{NS}}(\boldsymbol{\theta}_d \mathbf{1}_n^{\text{T}}, \boldsymbol{Y}_{d\times n})$ is twice continuously differentiable for all $\boldsymbol{\theta}_d \in \mathbb{R}^d$. Its Hessian matrix is diagonal with strictly positive entries $(2/n) \sum_{j=1}^{n} w(\boldsymbol{Y}_{\cdot,j})$ on the diagonal and is therefore positive definite for all $\boldsymbol{\theta}_d \in \mathbb{R}^d$.

This implies that $\bar{L}_{\text{NS}}(\boldsymbol{\theta}_d \mathbf{1}_n^{\text{T}}, \boldsymbol{Y}_{d\times n})$ is strictly convex over $\mathbb{R}^d$ (Rockafellar 1970, Theorem 4.5, Boyd and Vandenberghe 2004, p. 71). A strictly convex function has at most one critical point, which, if it exists, is the unique global minimizer (Boyd and Vandenberghe 2004, p. 69).

Setting the gradient in eq. (C.14) to 0 gives for each $k$:

$$(2/n) \sum_{j=1}^{n} w(\boldsymbol{Y}_{\cdot,j})(\theta_k - y_{kj}) = 0 \tag{C.17}$$

Solving for $\theta_k$, leads to the unique critical point:

$$\hat{\theta}_k(\boldsymbol{Y}_{d\times n}) = \sum_{j=1}^{n} w(\boldsymbol{Y}_{\cdot,j}) y_{kj} \,/\, \sum_{j=1}^{n} w(\boldsymbol{Y}_{\cdot,j}), k = 1, \dots, d \tag{C.18}$$

This can be written equivalently as:

$$\hat{\theta}_k(\boldsymbol{Y}_{d\times n}) = (\boldsymbol{Y}_{k,\cdot}\boldsymbol{w}_n)/(n\mu(\boldsymbol{w}_n)), k = 1, \dots, d \qquad \text{(C.19)}$$

where $\boldsymbol{w}_n = (w(\boldsymbol{Y}_{\cdot,1}), \dots, w(\boldsymbol{Y}_{\cdot,n}))^{\mathrm{T}}$.

From eq. (C.18), it follows directly that $\hat{\boldsymbol{\theta}}_d(\underline{\boldsymbol{Y}}_{d\times n})$ is a consistent estimator of the population parameter $(\mathbb{E}_F[\underline{y}_1 w(\underline{\boldsymbol{y}}_d)]/\mathbb{E}_F[w(\underline{\boldsymbol{y}}_d)], \dots, \mathbb{E}_F[\underline{y}_d w(\underline{\boldsymbol{y}}_d)]/\mathbb{E}_F[w(\underline{\boldsymbol{y}}_d)])^{\mathrm{T}}$.■

**Proof C.3** (Verification of uncorrelated $\underline{\boldsymbol{y}}_d$ and $w(\underline{\boldsymbol{y}}_d)$ for IID normal vectors $\underline{\boldsymbol{y}}_d$, as referenced in Section 4.4):

Let $\underline{\boldsymbol{y}}_d \sim N_d(\mu\mathbf{1}_d, \sigma^2\boldsymbol{I}_d), d > 3$ with CDF $F_{\underline{\boldsymbol{y}}_d}$. We show that $\mathbb{E}_{F_{\underline{\boldsymbol{y}}_d}}[\underline{\boldsymbol{y}}_d w(\underline{\boldsymbol{y}}_d)] = \mathbb{E}_{F_{\underline{\boldsymbol{y}}_d}}[\underline{\boldsymbol{y}}_d]\mathbb{E}_{F_{\underline{\boldsymbol{y}}_d}}[w(\underline{\boldsymbol{y}}_d)]$, where $w(\underline{\boldsymbol{y}}_d) = 1/\|\mu(\boldsymbol{y}_d)\mathbf{1}_d - \boldsymbol{y}_d\|_2^2$ from eq. (1.9).

Define the centered vector (see eq. (A.31) for definition of centered vectors) $\underline{\boldsymbol{y}}_{dc} = \underline{\boldsymbol{y}}_d - \mathbf{1}_d\mu(\underline{\boldsymbol{y}}_d) = (\boldsymbol{I}_d - (1/d)\mathbf{1}_d\mathbf{1}_d^{\mathrm{T}})\underline{\boldsymbol{y}}_d$ with CDF $F_{\underline{\boldsymbol{y}}_{dc}}$. Then $\underline{\boldsymbol{y}}_{dc} \sim N_d(\mathbf{0}_d, \sigma^2(\boldsymbol{I}_d - (1/d)\mathbf{1}_d\mathbf{1}_d^{\mathrm{T}}))$. Its covariance matrix has rank $d-1$ because $\mathbf{1}_d^{\mathrm{T}}\underline{\boldsymbol{y}}_{dc} = 0$ (from eq. (A.32)). The random variables $\mathbf{1}_d\mu(\underline{\boldsymbol{y}}_d) = (1/d)\mathbf{1}_d\mathbf{1}_d^{\mathrm{T}}\underline{\boldsymbol{y}}_d$ and $\underline{\boldsymbol{y}}_{dc}$ are independent because they are jointly Gaussian and their cross-covariance is 0. Specifically, $\mathrm{Cov}[\underline{\boldsymbol{y}}_{dc}, \mathbf{1}_d\mu(\underline{\boldsymbol{y}}_d)] = (\boldsymbol{I}_d - (1/d)\mathbf{1}_d\mathbf{1}_d^{\mathrm{T}})\mathrm{Cov}[\underline{\boldsymbol{y}}_d, \underline{\boldsymbol{y}}_d]((1/d)\mathbf{1}_d\mathbf{1}_d^{\mathrm{T}})^{\mathrm{T}} = (1/d)(\boldsymbol{I}_d - (1/d)\mathbf{1}_d\mathbf{1}_d^{\mathrm{T}})(\sigma^2\boldsymbol{I}_d)\mathbf{1}_d\mathbf{1}_d^{\mathrm{T}} = \mathbf{0}_{d\times d}$, where the zero matrix $\mathbf{0}_{d\times d}$ is defined in eq. (A.8). Independence follows from Proposition 3.3 in Eaton (1983), which applies to jointly Gaussian vectors with zero cross-covariance even when the individual covariance matrices are positive semidefinite (see Proposition 3.2 in Eaton 1983). This covers the case of $\underline{\boldsymbol{y}}_{dc}$, whose covariance matrix is positive semidefinite.

Furthermore, the weight function satisfies $w(\boldsymbol{y}_d) = 1/||\boldsymbol{y}_{dc}||_2^2$, so $w(\boldsymbol{y}_d) = w(\boldsymbol{y}_{dc})$ depends only on $\boldsymbol{y}_{dc}$. Consequently, $\mathbf{1}_d\mu(\underline{\boldsymbol{y}}_d)$ and $w(\underline{\boldsymbol{y}}_d)$ are independent.

Now examine the expectation $\mathbb{E}_{F_{\underline{\boldsymbol{y}}_d}}[\underline{\boldsymbol{y}}_d w(\underline{\boldsymbol{y}}_d)] = \mathbb{E}_{F_{\underline{\boldsymbol{y}}_d}}[\mathbf{1}_d\mu(\underline{\boldsymbol{y}}_d)w(\underline{\boldsymbol{y}}_d)] + \mathbb{E}_{F_{\underline{\boldsymbol{y}}_d}}[\underline{\boldsymbol{y}}_{dc}w(\underline{\boldsymbol{y}}_d)]$.

The first term, by independence of $\mathbf{1}_d\mu(\underline{\boldsymbol{y}}_d)$ and $w(\underline{\boldsymbol{y}}_d)$ simplifies to $\mathbb{E}_{F_{\underline{\boldsymbol{y}}_d}}[\mathbf{1}_d\mu(\underline{\boldsymbol{y}}_d)w(\underline{\boldsymbol{y}}_d)] = \mathbb{E}_{F_{\underline{\boldsymbol{y}}_d}}[\mathbf{1}_d\mu(\underline{\boldsymbol{y}}_d)]\mathbb{E}_{F_{\underline{\boldsymbol{y}}_d}}[w(\underline{\boldsymbol{y}}_d)] = \mathbb{E}_{F_{\underline{\boldsymbol{y}}_d}}[\underline{\boldsymbol{y}}_d]\mathbb{E}_{F_{\underline{\boldsymbol{y}}_d}}[w(\underline{\boldsymbol{y}}_d)]$.

Regarding the second term, the residual vector $\underline{\boldsymbol{y}}_{dc}$ has symmetric distribution about $\mathbf{0}_d$. Recall that $w(\boldsymbol{y}_d) = w(\boldsymbol{y}_{dc})$. Therefore, $\mathbb{E}_{F_{\underline{\boldsymbol{y}}_d}}[\underline{\boldsymbol{y}}_{dc}w(\underline{\boldsymbol{y}}_d)] = \mathbb{E}_{F_{\underline{\boldsymbol{y}}_d}}[(\underline{\boldsymbol{y}}_d - \mathbf{1}_d\mu(\underline{\boldsymbol{y}}_d))w(\underline{\boldsymbol{y}}_d)] = \mathbb{E}_{F_{\underline{\boldsymbol{y}}_{dc}}}[\underline{\boldsymbol{y}}_{dc}w(\underline{\boldsymbol{y}}_{dc})]$ by the law of the unconscious statistician. The function $\underline{\boldsymbol{y}}_{dc}w(\underline{\boldsymbol{y}}_{dc})$ is an odd function of $\underline{\boldsymbol{y}}_{dc}$, while $\underline{\boldsymbol{y}}_{dc}$ has a symmetric distribution about $\mathbf{0}_d$. For any odd integrable function vector valued function $g$, with components $g_i, i = 1, \dots, d$ and symmetric distribution $F$, $\mathbb{E}_F[g_i(\underline{\boldsymbol{y}})] = \int_{-\infty}^{\infty} g_i(\boldsymbol{y}) f(\boldsymbol{y}) \mathrm{d}\boldsymbol{y} = \int_{\infty}^{-\infty} g_i(-\boldsymbol{u}) f(-\boldsymbol{u}) \mathrm{d}(-\boldsymbol{u}) = \int_{-\infty}^{\infty} g_i(-\boldsymbol{u}) f(-\boldsymbol{u}) \mathrm{d}\boldsymbol{u} = -\int_{-\infty}^{\infty} g_i(\boldsymbol{u}) f(\boldsymbol{u}) \mathrm{d}\boldsymbol{u} = -\mathbb{E}_F[g_i(\underline{\boldsymbol{y}})]$, which implies that $\mathbb{E}_F[g_i(\underline{\boldsymbol{y}})] = 0$. Therefore, $\mathbb{E}_{F_{\underline{\boldsymbol{y}}_{dc}}}[\underline{\boldsymbol{y}}_{dc}w(\underline{\boldsymbol{y}}_{dc})] = \mathbf{0}_d$.

Combining the two terms gives $\mathbb{E}_{F_{\underline{\boldsymbol{y}}_d}}[\underline{\boldsymbol{y}}_d w(\underline{\boldsymbol{y}}_d)] = \mathbb{E}_{F_{\underline{\boldsymbol{y}}_d}}[\underline{\boldsymbol{y}}_d]\mathbb{E}_F[w(\underline{\boldsymbol{y}}_d)]$, which is the target equality.

We verify that $\mathbb{E}_{F_{\underline{\boldsymbol{y}}_d}}[w(\underline{\boldsymbol{y}}_d)]$ exists and is finite. The covariance matrix of $\underline{\boldsymbol{y}}_{dc}$ has rank $(d-1)$, which implies $||\underline{\boldsymbol{y}}_{dc}||_2^2/\sigma^2 \sim \chi^2(d-1)$ (Cochran 1934). Here $\underline{U} = ||\underline{\boldsymbol{y}}_{dc}||_2^2$ is the sum of squared Gaussian random variables, and the chi-squared PDF appears in eq. (B.3). From eq. (B.4), $\sigma^2/\underline{U}$ follows an inverse chi-squared distribution with $d-1$ degrees of freedom. Therefore, $\mathbb{E}_{F_{\underline{\boldsymbol{y}}_d}}[w(\underline{\boldsymbol{y}}_d)] = \mathbb{E}[1/\underline{U}] = 1/(\sigma^2(d-3))$.

We now turn to $\mathbb{E}_{F_{\underline{\boldsymbol{y}}_{dc}}}[\underline{\boldsymbol{y}}_{dc}w(\underline{\boldsymbol{y}}_{dc})]$, beginning with component $\mathbb{E}_{F_{\underline{\boldsymbol{y}}_{dc}}}[\underline{\boldsymbol{y}}_{dc,i}w(\underline{\boldsymbol{y}}_{dc})]$. Observe that $w(\underline{\boldsymbol{y}}_{dc}) = 1/\underline{U}$ with $\underline{U}/\sigma^2 \sim \chi^2(d-1)$. Each component satisfies $|\underline{\boldsymbol{y}}_{dc,i}| \leq \underline{U}^{1/2}$ (as the absolute value of a coordinate cannot exceed the vector norm), giving $|\underline{\boldsymbol{y}}_{dc,i}/\underline{U}| \leq \underline{U}^{-1/2}$. Because $\underline{U}/\sigma^2 \sim \chi^2(d-1)$, Appendix B shows that $\sigma/\underline{U}^{1/2} \sim \mathrm{Inv}-\mathrm{Gamma}^{-1/2}(2, (d-1)/2)$; see eq. (B.7) for the density. Hence for $d > 3$ the expectation $\mathbb{E}[\underline{U}^{-1/2}]$ exists and is finite. Consequently, $\mathbb{E}[|\underline{\boldsymbol{y}}_{dc,i}/\underline{U}|] < \infty$, which ensures that $\mathbb{E}_{F_{\underline{\boldsymbol{y}}_{dc}}}[\underline{\boldsymbol{y}}_{dc,i}w(\underline{\boldsymbol{y}}_{dc})]$ exists and is finite.■

**Proof C.4** (Proof for estimators of one-dimensional linear regression in Section 4.8, Gentle 2024, p.441):

We prove that the parameter estimates of the one-dimensional linear regression are given by eq. (4.16).

First, observe that

$$\boldsymbol{a}_p^{\mathrm{T}}\boldsymbol{X}_{\cdot,j} + b = \begin{bmatrix} \boldsymbol{a}_p^{\mathrm{T}} & b \end{bmatrix} \begin{bmatrix} \boldsymbol{X}_{\cdot,j} \\ 1 \end{bmatrix} = \boldsymbol{\theta}_{1\times(p+1)}\widetilde{\boldsymbol{X}}_{j,\cdot}^{\mathrm{T}} \tag{C.20}$$

where $\boldsymbol{\theta}_{1\times(p+1)} = \begin{bmatrix} \boldsymbol{a}_p^{\mathrm{T}} & b \end{bmatrix}$ are the parameters of the model in eq. (4.14) and $\widetilde{\boldsymbol{X}}_{n\times(p+1)}$ is the augmented matrix constructed from the realized values predictor matrix $\boldsymbol{X}_{p\times n}$ in eq. (4.15) together with a column of ones

$$\widetilde{\boldsymbol{X}}_{n\times(p+1)} = \begin{bmatrix}\boldsymbol{X}^{\mathrm{T}}_{p\times n} & \mathbf{1}_n\end{bmatrix} = \begin{bmatrix}\boldsymbol{X}^{\mathrm{T}}_{\cdot,1} & 1 \\ \vdots & \vdots \\ \boldsymbol{X}^{\mathrm{T}}_{\cdot,n} & 1\end{bmatrix} \tag{C.21}$$

The product $\widetilde{\boldsymbol{X}}^{\mathrm{T}}_{n\times(p+1)}\widetilde{\boldsymbol{X}}_{n\times(p+1)}$ is a symmetric square matrix (Gentle 2024, p.137). Under the assumption that $\widetilde{\boldsymbol{X}}_{n\times(p+1)}$ has full column rank, $\widetilde{\boldsymbol{X}}^{\mathrm{T}}_{n\times(p+1)}\widetilde{\boldsymbol{X}}_{n\times(p+1)}$ is of full rank and positive definite (Gentle 2024, p.138). Consequently, it is nonsingular (Gentle 2024, p.122) and therefore invertible (Gentle 2024, p.129).

The vector of predictions can be written as

$$\boldsymbol{z}^{\mathrm{T}}_n = \boldsymbol{\theta}_{1\times(p+1)}\widetilde{\boldsymbol{X}}^{\mathrm{T}}_{n\times(p+1)} \tag{C.22}$$

Hence, the estimate in eq. (4.15) becomes

$$\widehat{\boldsymbol{\theta}}_{1\times(p+1)}(\boldsymbol{X}_{p\times n}, \boldsymbol{y}_n) = \underset{\boldsymbol{\theta}_{1(p+1)}\in\boldsymbol{\Theta}}{\arg\min}(1/n)||(\boldsymbol{\theta}_{1(p+1)}\widetilde{\boldsymbol{X}}^{\mathrm{T}}_{n\times(p+1)} - \boldsymbol{y}^{\mathrm{T}}_n)^{\mathrm{T}}||^2_2 \tag{C.23}$$

where $\boldsymbol{y}_n$ represents the realizations of the random variable $\underline{y}$ in eq. (4.15).

Define the objective function:

$$f(\boldsymbol{\theta}_{1\times(p+1)}) = ||(\boldsymbol{\theta}_{1\times(p+1)}\widetilde{\boldsymbol{X}}^{\mathrm{T}}_{n\times(p+1)} - \boldsymbol{y}^{\mathrm{T}}_n)^{\mathrm{T}}||^2_2 \tag{C.24}$$

Expanding this gives

$$f(\boldsymbol{\theta}_{1\times(p+1)}) = \boldsymbol{\theta}_{1\times(p+1)}\widetilde{\boldsymbol{X}}^{\mathrm{T}}_{n\times(p+1)}\widetilde{\boldsymbol{X}}_{n\times(p+1)}\boldsymbol{\theta}^{\mathrm{T}}_{1\times(p+1)} - 2\boldsymbol{y}^{\mathrm{T}}_n\widetilde{\boldsymbol{X}}_{n\times(p+1)}\boldsymbol{\theta}^{\mathrm{T}}_{1\times(p+1)} + ||\boldsymbol{y}_n||^2_2 \tag{C.25}$$

Taking the gradient with respect to $\boldsymbol{\theta}^{\mathrm{T}}_{1\times(p+1)}$ (using eqs. (A.26) and (A.27)) gives:

$$\frac{\partial f(\boldsymbol{\theta}_{1\times(p+1)})}{\partial\boldsymbol{\theta}^{\mathrm{T}}_{1\times(p+1)}} = 2(\boldsymbol{\theta}_{1\times(p+1)}\widetilde{\boldsymbol{X}}^{\mathrm{T}}_{n\times(p+1)}\widetilde{\boldsymbol{X}}_{n\times(p+1)} - \boldsymbol{y}^{\mathrm{T}}_n\widetilde{\boldsymbol{X}}_{n\times(p+1)}) \tag{C.26}$$

The gradient (using eq. (A.24)) is

$$\nabla_{\boldsymbol{\theta}^{\mathrm{T}}_{1\times(p+1)}} f(\boldsymbol{\theta}_{1\times(p+1)}) = \left(\frac{\partial f(\boldsymbol{\theta}_{1\times(p+1)})}{\partial\boldsymbol{\theta}^{\mathrm{T}}_{1\times(p+1)}}\right)^{\mathrm{T}} = 2(\widetilde{\boldsymbol{X}}^{\mathrm{T}}_{n\times(p+1)}\widetilde{\boldsymbol{X}}_{n\times(p+1)}\boldsymbol{\theta}^{\mathrm{T}}_{1\times(p+1)} - \widetilde{\boldsymbol{X}}^{\mathrm{T}}_{n\times(p+1)}\boldsymbol{y}_n) \tag{C.27}$$

Setting the gradient to $\mathbf{0}_{p+1}$ gives the critical point

$$\widehat{\boldsymbol{\theta}}_{1\times(p+1)}(\boldsymbol{X}_{p\times n}, \boldsymbol{y}_n) = \boldsymbol{y}^{\mathrm{T}}_n\widetilde{\boldsymbol{X}}_{n\times(p+1)}(\widetilde{\boldsymbol{X}}^{\mathrm{T}}_{n\times(p+1)}\widetilde{\boldsymbol{X}}_{n\times(p+1)})^{-1} \tag{C.28}$$

The function $f(\boldsymbol{\theta}_{1\times(p+1)})$ is strictly convex, because the Hessian matrix (using eqs. (A.25) and (A.28))

$$\boldsymbol{H}_{\boldsymbol{\theta}^{\mathrm{T}}_{1\times(p+1)}} f(\boldsymbol{\theta}_{1\times(p+1)}) = \frac{\partial\nabla_{\boldsymbol{\theta}^{\mathrm{T}}_{1\times(p+1)}} f(\boldsymbol{\theta}_{1\times(p+1)})}{\partial\boldsymbol{\theta}^{\mathrm{T}}_{1\times(p+1)}} = 2\widetilde{\boldsymbol{X}}^{\mathrm{T}}_{n\times(p+1)}\widetilde{\boldsymbol{X}}_{n\times(p+1)} \tag{C.29}$$

is positive definite (Boyd and Vandenberghe 2004, p.71). Therefore, the critical point is the unique global minimizer (Boyd and Vandenberghe 2004, p.69). This confirms that the estimate for the one-dimensional linear regression model is given by eq. (4.16).■

**Proof C.5** (Proof for estimators of multi-dimensional linear regression in Section 4.8, Gentle 2024, p.457):

We prove that the parameter estimates of the multi-dimensional linear regression are given by eq. (4.21).

The model in eq. (4.19) for a single observation $j$ can be written as

$$\boldsymbol{z}_j = \boldsymbol{A}_{d\times p}\boldsymbol{X}_{\cdot,j} + \boldsymbol{b}_d = [\boldsymbol{A}_{d\times p} \quad \boldsymbol{b}_d]\begin{bmatrix}\boldsymbol{X}_{\cdot,j}\\ 1\end{bmatrix} = \boldsymbol{\theta}_{d\times(p+1)}\widetilde{\boldsymbol{X}}_{j,\cdot}^{\mathrm{T}} \tag{C.30}$$

where $\boldsymbol{\theta}_{d\times(p+1)} = [\boldsymbol{A}_{d\times p} \quad \boldsymbol{b}_d]$ are the parameters of the model in eq. (4.19) and $\widetilde{\boldsymbol{X}}_{n\times(p+1)}$ is the augmented matrix constructed from the realized values predictor matrix $\boldsymbol{X}_{p\times n}$ in eq. (4.20) together with a column of ones,

$$\widetilde{\boldsymbol{X}}_{n\times(p+1)} = [\boldsymbol{X}_{p\times n}^{\mathrm{T}} \quad \boldsymbol{1}_n] = \begin{bmatrix}\boldsymbol{X}_{\cdot,1}^{\mathrm{T}} & 1\\ \vdots & \vdots\\ \boldsymbol{X}_{\cdot,n}^{\mathrm{T}} & 1\end{bmatrix} \tag{C.31}$$

The product $\widetilde{\boldsymbol{X}}_{n\times(p+1)}^{\mathrm{T}}\widetilde{\boldsymbol{X}}_{n\times(p+1)}$ is a symmetric square matrix (Gentle 2024, p.137). Under the assumption that $\widetilde{\boldsymbol{X}}_{n\times(p+1)}$ has full column rank, $\widetilde{\boldsymbol{X}}_{n\times(p+1)}^{\mathrm{T}}\widetilde{\boldsymbol{X}}_{n\times(p+1)}$ is of full rank and positive definite (Gentle 2024, p.138). Consequently, it is nonsingular (Gentle 2024, p.122) and therefore invertible (Gentle 2024, p.129).

Let $\boldsymbol{Z}_{d\times n}$ be the matrix of predictions in eq, (4.20), with columns $\boldsymbol{Z}_{\cdot,j} = \boldsymbol{z}_j$. Then

$$\boldsymbol{Z}_{d\times n} = \boldsymbol{\theta}_{d\times(p+1)}\widetilde{\boldsymbol{X}}_{n\times(p+1)}^{\mathrm{T}} \tag{C.32}$$

The estimate in eq, (4.20) becomes:

$$\widehat{\boldsymbol{\theta}}_{d\times(p+1)}(\boldsymbol{X}_{p\times n}, \boldsymbol{Y}_{d\times n}) := \underset{\boldsymbol{\theta}_{d\times(p+1)}\in\boldsymbol{\Theta}}{\arg\min}\ (1/n)\sum_{j=1}^{n} L_{\mathrm{EN}}(\boldsymbol{\theta}_{d\times(p+1)}\widetilde{\boldsymbol{X}}_{j,\cdot}^{\mathrm{T}}, \boldsymbol{Y}_{\cdot,j}) \tag{C.33}$$

From eq. (3.32), this can be rewritten as:

$$\widehat{\boldsymbol{\theta}}_{d\times(p+1)}(\boldsymbol{X}_{p\times n}, \boldsymbol{Y}_{d\times n}) := \underset{\boldsymbol{\theta}_{d\times(p+1)}\in\boldsymbol{\Theta}}{\arg\min}\ (1/n)\sum_{j=1}^{n} ||\boldsymbol{\theta}_{d\times(p+1)}\widetilde{\boldsymbol{X}}_{j,\cdot}^{\mathrm{T}} - \boldsymbol{Y}_{\cdot,j}||_2^2 \tag{C.34}$$

Define the objective function

$$f(\boldsymbol{\theta}_{d\times(p+1)}) = \sum_{j=1}^{n} ||\boldsymbol{\theta}_{d\times(p+1)}\widetilde{\boldsymbol{X}}_{j,\cdot}^{\mathrm{T}} - \boldsymbol{Y}_{\cdot,j}||_2^2 = ||\boldsymbol{\theta}_{d\times(p+1)}\widetilde{\boldsymbol{X}}_{n\times(p+1)}^{\mathrm{T}} - \boldsymbol{Y}_{d\times n}||_F^2 = ||\boldsymbol{Z}_{d\times n} - \boldsymbol{Y}_{d\times n}||_F^2 \tag{C.35}$$

where $||\cdot||_{\mathrm{F}}^2$ is the squared Frobenius norm (see eq. (A.14)).

To facilitate the analysis, we vectorize the parameter matrix $\boldsymbol{\theta}_{d\times(p+1)}$, the realization matrix $\boldsymbol{Y}_{d\times n}$ and the prediction matrix $\boldsymbol{Z}_{d\times n}$. Define respectively the $d(p+1)$-, $dn$- and $dn$–dimensional vectors (see eqs. (A.19) and (A.20))

$$\boldsymbol{\beta}_{d(p+1)} = \mathrm{vec}(\boldsymbol{\theta}_{d\times(p+1)}) \tag{C.36}$$

$$\boldsymbol{y}_{dn} = \mathrm{vec}(\boldsymbol{Y}_{d\times n}) \tag{C.37}$$

$$\mathrm{vec}(\boldsymbol{Z}_{d\times n}) = \mathrm{vec}(\boldsymbol{I}_d\boldsymbol{\theta}_{d\times(p+1)}\widetilde{\boldsymbol{X}}^{\mathrm{T}}_{n\times(p+1)}) = (\widetilde{\boldsymbol{X}}_{n\times(p+1)} \otimes \boldsymbol{I}_d)\boldsymbol{\beta}_{d(p+1)} \tag{C.38}$$

The objective function then becomes

$$f(\boldsymbol{\beta}_{d(p+1)}) = ||(\widetilde{\boldsymbol{X}}_{n\times(p+1)} \otimes \boldsymbol{I}_d)\boldsymbol{\beta}_{d(p+1)} - \boldsymbol{y}_{dn}||_2^2 \tag{C.39}$$

Expanding $f$ gives

$$f(\boldsymbol{\beta}_{d(p+1)}) = \boldsymbol{\beta}^{\mathrm{T}}_{d(p+1)}(\widetilde{\boldsymbol{X}}_{n\times(p+1)} \otimes \boldsymbol{I}_d)^{\mathrm{T}}(\widetilde{\boldsymbol{X}}_{n\times(p+1)} \otimes \boldsymbol{I}_d)\boldsymbol{\beta}_{d(p+1)} - \boldsymbol{y}^{\mathrm{T}}_{dn}(\widetilde{\boldsymbol{X}}_{n\times(p+1)} \otimes \boldsymbol{I}_d)\boldsymbol{\beta}_{d(p+1)} - \boldsymbol{\beta}^{\mathrm{T}}_{d(p+1)}(\widetilde{\boldsymbol{X}}_{n\times(p+1)} \otimes \boldsymbol{I}_d)^{\mathrm{T}}\boldsymbol{y}_{dn} + \boldsymbol{y}^{\mathrm{T}}_{dn}\boldsymbol{y}_{dn} \tag{C.40}$$

By eqs. (A.16) and (A.17),

$$(\widetilde{\boldsymbol{X}}_{n\times(p+1)} \otimes \boldsymbol{I}_d)^{\mathrm{T}}(\widetilde{\boldsymbol{X}}_{n\times(p+1)} \otimes \boldsymbol{I}_d) = (\widetilde{\boldsymbol{X}}^{\mathrm{T}}_{n\times(p+1)}\widetilde{\boldsymbol{X}}_{n\times(p+1)}) \otimes \boldsymbol{I}_d \tag{C.41}$$

$\boldsymbol{\beta}^{\mathrm{T}}_{d(p+1)}(\widetilde{\boldsymbol{X}}_{n\times(p+1)} \otimes \boldsymbol{I}_d)^{\mathrm{T}}\boldsymbol{y}_{dn}$ is a scalar and therefore equals its transpose:

$$\boldsymbol{\beta}^{\mathrm{T}}_{d(p+1)}(\widetilde{\boldsymbol{X}}_{n\times(p+1)} \otimes \boldsymbol{I}_d)^{\mathrm{T}}\boldsymbol{y}_{dn} = (\boldsymbol{\beta}^{\mathrm{T}}_{d(p+1)}(\widetilde{\boldsymbol{X}}_{n\times(p+1)} \otimes \boldsymbol{I}_d)^{\mathrm{T}}\boldsymbol{y}_{dn})^{\mathrm{T}} = \boldsymbol{y}^{\mathrm{T}}_{dn}(\widetilde{\boldsymbol{X}}_{n\times(p+1)} \otimes \boldsymbol{I}_d)\boldsymbol{\beta}_{d(p+1)} \tag{C.42}$$

Thus:

$$f(\boldsymbol{\beta}_{d(p+1)}) = \boldsymbol{\beta}^{\mathrm{T}}_{d(p+1)}((\widetilde{\boldsymbol{X}}^{\mathrm{T}}_{n\times(p+1)}\widetilde{\boldsymbol{X}}_{n\times(p+1)}) \otimes \boldsymbol{I}_d)\boldsymbol{\beta}_{d(p+1)} - 2\boldsymbol{y}^{\mathrm{T}}_{dn}(\widetilde{\boldsymbol{X}}_{n\times(p+1)} \otimes \boldsymbol{I}_d)\boldsymbol{\beta}_{d(p+1)} + \boldsymbol{y}^{\mathrm{T}}_{dn}\boldsymbol{y}_{dn} \tag{C.43}$$

The gradient of $f$ with respect to $\boldsymbol{\beta}_{d(p+1)}$ is (applying eqs. (A.26) and (A.27))

$$\frac{\partial f(\boldsymbol{\beta}_{d(p+1)})}{\partial \boldsymbol{\beta}_{d(p+1)}} = 2(\boldsymbol{\beta}^{\mathrm{T}}_{d(p+1)}((\widetilde{\boldsymbol{X}}^{\mathrm{T}}_{n\times(p+1)}\widetilde{\boldsymbol{X}}_{n\times(p+1)}) \otimes \boldsymbol{I}_d) - \boldsymbol{y}^{\mathrm{T}}_{dn}(\widetilde{\boldsymbol{X}}_{n\times(p+1)} \otimes \boldsymbol{I}_d)) \tag{C.44}$$

Applying eqs. (A.16) and (A.24), the gradient vector becomes

$$\nabla_{\boldsymbol{\beta}_{d(p+1)}} f(\boldsymbol{\beta}_{d(p+1)}) = \left(\frac{\partial f(\boldsymbol{\beta}_{d(p+1)})}{\partial \boldsymbol{\beta}_{d(p+1)}}\right)^{\mathrm{T}} = 2(((\widetilde{\boldsymbol{X}}^{\mathrm{T}}_{n\times(p+1)}\widetilde{\boldsymbol{X}}_{n\times(p+1)}) \otimes \boldsymbol{I}_d)\boldsymbol{\beta}_{d(p+1)} - (\widetilde{\boldsymbol{X}}^{\mathrm{T}}_{n\times(p+1)} \otimes \boldsymbol{I}_d)\boldsymbol{y}_{dn}) \tag{C.45}$$

Setting the gradient to $\boldsymbol{0}_{d(p+1)}$:

$$((\widetilde{\boldsymbol{X}}^{\mathrm{T}}_{n\times(p+1)}\widetilde{\boldsymbol{X}}_{n\times(p+1)}) \otimes \boldsymbol{I}_d)\boldsymbol{\beta}_{d(p+1)} = (\widetilde{\boldsymbol{X}}^{\mathrm{T}}_{n\times(p+1)} \otimes \boldsymbol{I}_d)\boldsymbol{y}_{dn} \tag{C.46}$$

$\widetilde{\boldsymbol{X}}^{\mathrm{T}}_{n\times(p+1)}\widetilde{\boldsymbol{X}}_{n\times(p+1)}$ is invertible. Applying eq. (A.18) to invert the Kronecker product $((\widetilde{\boldsymbol{X}}^{\mathrm{T}}_{n\times(p+1)}\widetilde{\boldsymbol{X}}_{n\times(p+1)})\otimes \boldsymbol{I}_d)$:

$$\boldsymbol{\beta}_{d(p+1)} = ((\widetilde{\boldsymbol{X}}^{\mathrm{T}}_{n\times(p+1)}\widetilde{\boldsymbol{X}}_{n\times(p+1)})^{-1}\otimes \boldsymbol{I}_d)(\widetilde{\boldsymbol{X}}^{\mathrm{T}}_{n\times(p+1)}\otimes \boldsymbol{I}_d)\boldsymbol{y}_{dn} \tag{C.47}$$

Simplifying the Kronecker products with eq. (A.17) gives

$$\boldsymbol{\beta}_{d(p+1)} = (((\widetilde{\boldsymbol{X}}^{\mathrm{T}}_{n\times(p+1)}\widetilde{\boldsymbol{X}}_{n\times(p+1)})^{-1}\widetilde{\boldsymbol{X}}^{\mathrm{T}}_{n\times(p+1)})\otimes \boldsymbol{I}_d)\boldsymbol{y}_{dn} = \\ ((\widetilde{\boldsymbol{X}}_{n\times(p+1)}(\widetilde{\boldsymbol{X}}^{\mathrm{T}}_{n\times(p+1)}\widetilde{\boldsymbol{X}}_{n\times(p+1)})^{-1})^{\mathrm{T}}\otimes \boldsymbol{I}_d)\mathrm{vec}(\boldsymbol{Y}_{d\times n}) \tag{C.48}$$

By eq. (A.20):

$$\boldsymbol{\beta}_{d(p+1)} = \mathrm{vec}(\boldsymbol{I}_d\boldsymbol{Y}_{d\times n}\widetilde{\boldsymbol{X}}_{n\times(p+1)}(\widetilde{\boldsymbol{X}}^{\mathrm{T}}_{n\times(p+1)}\widetilde{\boldsymbol{X}}_{n\times(p+1)})^{-1}) \tag{C.49}$$

Applying eq. (C.36) and because the vectorization operator is bijective, it follows that

$$\widehat{\boldsymbol{\theta}}_{d\times(p+1)}(\boldsymbol{X}_{p\times n},\boldsymbol{Y}_{d\times n}) = \boldsymbol{Y}_{d\times n}\widetilde{\boldsymbol{X}}_{n\times(p+1)}(\widetilde{\boldsymbol{X}}^{\mathrm{T}}_{n\times(p+1)}\widetilde{\boldsymbol{X}}_{n\times(p+1)})^{-1} \tag{C.50}$$

which is exactly eq. (4.21).

To verify that this critical point is a minimum, we examine the second derivative. The Hessian matrix of $f(\boldsymbol{\beta}_{d(p+1)})$ (using eq. (A.28)) is

$$\boldsymbol{H}_{\boldsymbol{\beta}_{d(p+1)}}f(\boldsymbol{\beta}_{d(p+1)}) = \frac{\partial \nabla_{\boldsymbol{\beta}_{d(p+1)}} f(\boldsymbol{\beta}_{d(p+1)})}{\partial \boldsymbol{\beta}_{d(p+1)}} = 2((\widetilde{\boldsymbol{X}}^{\mathrm{T}}_{n\times(p+1)}\widetilde{\boldsymbol{X}}_{n\times(p+1)})\otimes \boldsymbol{I}_d) \tag{C.51}$$

The matrices $\boldsymbol{I}_d$ and $\widetilde{\boldsymbol{X}}^{\mathrm{T}}_{n\times(p+1)}\widetilde{\boldsymbol{X}}_{n\times(p+1)}$ are symmetric and positive definite. Therefore, their eigenvalues are positive (Gentle 2024, p.186). Their Kronecker product is a symmetric matrix (Gentle 2024, p.118). The eigenvalues of the Kronecker product are all products of eigenvalues of the individual matrices, hence positive (Horn and Johnson 1991, p.245). Consequently, $\boldsymbol{H}$ is symmetric with all eigenvalues positive, therefore, it is positive definite for all $\boldsymbol{\beta}$ (Gentle 2024, p.186), confirming that $f(\boldsymbol{\beta}_{d(p+1)})$ is strictly convex (Boyd and Vandenberghe 2004, p.71). Therefore, the critical point is the unique global minimizer (Boyd and Vandenberghe 2004, p.69). This confirms that the estimate for the multi-dimensional linear regression model is given by eq. (4.21).■

**Proof C.6** (Proof for estimators of Nash-Sutcliffe linear regression in Section 4.8):

We prove that the parameter estimates of the Nash-Sutcliffe linear regression are given by eq. (4.26).

The model in eq. (4.24) for a single observation $j$ can be written as

$$\boldsymbol{A}_{d\times p}\boldsymbol{X}_{\cdot,j} + \boldsymbol{b}_d = [\boldsymbol{A}_{d\times p} \quad \boldsymbol{b}_d]\begin{bmatrix}\boldsymbol{X}_{\cdot,j}\\ 1\end{bmatrix} = \boldsymbol{\theta}_{d\times(p+1)}\widetilde{\boldsymbol{X}}^{\mathrm{T}}_{j,\cdot} \tag{C.52}$$

where $\boldsymbol{\theta}_{d\times(p+1)} = [\boldsymbol{A}_{d\times p} \quad \boldsymbol{b}_d]$ are the parameters of the model in eq. (4.24) and $\widetilde{\boldsymbol{X}}_{n\times(p+1)}$ is the augmented matrix constructed from the realized values predictor matrix $\boldsymbol{X}_{p\times n}$ in eq, (4.25) together with a column of ones,

$$\widetilde{\boldsymbol{X}}_{n\times(p+1)} = [\boldsymbol{X}^{\mathrm{T}}_{p\times n} \quad \mathbf{1}_n] = \begin{bmatrix} \boldsymbol{X}^{\mathrm{T}}_{\cdot,1} & 1 \\ \vdots & \vdots \\ \boldsymbol{X}^{\mathrm{T}}_{\cdot,n} & 1 \end{bmatrix} \tag{C.53}$$

Let from eq. (1.9)

$$w(\boldsymbol{Y}_{\cdot,j}) = 1/||\mu(\boldsymbol{Y}_{\cdot,j})\mathbf{1}_d - \boldsymbol{Y}_{\cdot,j}||_2^2 \tag{C.54}$$

Let the diagonal matrices $\boldsymbol{W}_{n\times n}(\boldsymbol{Y}_{d\times n})$ and $\boldsymbol{W}^{1/2}_{n\times n}(\boldsymbol{Y}_{d\times n})$ be (as in eq. (A.10)):

$$\boldsymbol{W}_{n\times n}(\boldsymbol{Y}_{d\times n}) := \mathrm{diag}(w(\boldsymbol{Y}_{\cdot,1}), \dots, w(\boldsymbol{Y}_{\cdot,n})) \tag{C.55}$$

$$\boldsymbol{W}^{1/2}_{n\times n}(\boldsymbol{Y}_{d\times n}) := \mathrm{diag}((w(\boldsymbol{Y}_{\cdot,1}))^{1/2}, \dots, (w(\boldsymbol{Y}_{\cdot,n}))^{1/2}) \tag{C.56}$$

The product $\widetilde{\boldsymbol{X}}^{\mathrm{T}}_{n\times(p+1)}\boldsymbol{W}_{n\times n}(\boldsymbol{Y}_{d\times n})\,\widetilde{\boldsymbol{X}}_{n\times(p+1)}$ is a symmetric square matrix (Gentle 2024, p.137). $\boldsymbol{W}_{n\times n}(\boldsymbol{Y}_{d\times n})$ is a positive definite matrix. Under the assumption that $\widetilde{\boldsymbol{X}}_{n\times(p+1)}$ has full column rank, $\widetilde{\boldsymbol{X}}^{\mathrm{T}}_{n\times(p+1)}\boldsymbol{W}_{n\times n}(\boldsymbol{Y}_{d\times n})\,\widetilde{\boldsymbol{X}}_{n\times(p+1)}$ is positive definite (Gentle 2024, p.134). Consequently, it is nonsingular (Gentle 2024, p.122) and therefore invertible (Gentle 2024, p.129).

Let $\boldsymbol{Z}_{d\times n}$ be the matrix of predictions in eq, (4.25), with columns $\boldsymbol{Z}_{\cdot,j} = \boldsymbol{z}_j$. Then

$$\boldsymbol{Z}_{d\times n} = \boldsymbol{\theta}_{d\times(p+1)}\widetilde{\boldsymbol{X}}^{\mathrm{T}}_{n\times(p+1)} \tag{C.57}$$

The estimate in eq, (4.25) becomes:

$$\widehat{\boldsymbol{\theta}}_{d\times(p+1)}(\boldsymbol{X}_{p\times n}, \boldsymbol{Y}_{d\times n}) := \underset{\boldsymbol{\theta}_{d\times(p+1)}\in\boldsymbol{\Theta}}{\arg\min}\ (1/n)\textstyle\sum_{j=1}^{n} L_{\mathrm{NS}}(\boldsymbol{\theta}_{d\times(p+1)}\widetilde{\boldsymbol{X}}^{\mathrm{T}}_{j,\cdot}, \boldsymbol{Y}_{\cdot,j}) \tag{C.58}$$

From eq. (1.7), this can be rewritten as:

$$\widehat{\boldsymbol{\theta}}_{d\times(p+1)}(\boldsymbol{X}_{p\times n}, \boldsymbol{Y}_{d\times n}) := \underset{\boldsymbol{\theta}_{d\times(p+1)}\in\boldsymbol{\Theta}}{\arg\min}\ (1/n)\textstyle\sum_{j=1}^{n} w(\boldsymbol{Y}_{\cdot,j})||\boldsymbol{\theta}_{d\times(p+1)}\widetilde{\boldsymbol{X}}^{\mathrm{T}}_{j,\cdot} - \boldsymbol{Y}_{\cdot,j}||_2^2 \tag{C.59}$$

Define the objective function

$$f(\boldsymbol{\theta}_{d\times(p+1)}) = \textstyle\sum_{j=1}^{n} w(\boldsymbol{Y}_{\cdot,j})||\boldsymbol{\theta}_{d\times(p+1)}\widetilde{\boldsymbol{X}}^{\mathrm{T}}_{j,\cdot} - \boldsymbol{Y}_{\cdot,j}||_2^2 = ||(\boldsymbol{\theta}_{d\times(p+1)}\widetilde{\boldsymbol{X}}^{\mathrm{T}}_{n\times(p+1)} - \boldsymbol{Y}_{d\times n})\boldsymbol{W}^{1/2}_{n\times n}(\boldsymbol{Y}_{d\times n})||_F^2 = ||(\boldsymbol{Z}_{d\times n} - \boldsymbol{Y}_{d\times n})\boldsymbol{W}^{1/2}_{n\times n}(\boldsymbol{Y}_{d\times n})||_F^2 \tag{C.60}$$

where $||\cdot||_{\mathrm{F}}^2$ is the squared Frobenius norm (see eq. (A.14)).

To facilitate the analysis, we vectorize the parameter matrix $\boldsymbol{\theta}_{d\times(p+1)}$, the realization matrix $\boldsymbol{Y}_{d\times n}$ and the prediction matrix $\boldsymbol{Z}_{d\times n}$. Define respectively the $d(p+1)$-, $dn$- and $dn$–dimensional vectors (see eqs. (A.19) and (A.20))

$$\boldsymbol{\beta}_{d(p+1)} = \mathrm{vec}(\boldsymbol{\theta}_{d\times(p+1)}) \tag{C.61}$$

$$\boldsymbol{y}_{dn} = \mathrm{vec}(\boldsymbol{Y}_{d\times n}) \tag{C.62}$$

$$\mathrm{vec}(\boldsymbol{Z}_{d\times n}) = \mathrm{vec}(\boldsymbol{I}_d \boldsymbol{\theta}_{d\times(p+1)} \widetilde{\boldsymbol{X}}_{n\times(p+1)}^{\mathrm{T}}) = (\widetilde{\boldsymbol{X}}_{n\times(p+1)} \otimes \boldsymbol{I}_d)\boldsymbol{\beta}_{d(p+1)} \tag{C.63}$$

The objective function then becomes

$$f(\boldsymbol{\beta}_{d(p+1)}) = ||(\boldsymbol{W}_{n\times n}^{1/2}(\boldsymbol{Y}_{d\times n}) \otimes \boldsymbol{I}_d)((\widetilde{\boldsymbol{X}}_{n\times(p+1)} \otimes \boldsymbol{I}_d)\boldsymbol{\beta}_{d(p+1)} - \boldsymbol{y}_{dn})||_2^2 \tag{C.64}$$

By eqs. (A.16) and (A.17)

$$(\boldsymbol{W}_{n\times n}^{1/2}(\boldsymbol{Y}_{d\times n}) \otimes \boldsymbol{I}_d)^{\mathrm{T}} = \boldsymbol{W}_{n\times n}^{1/2}(\boldsymbol{Y}_{d\times n}) \otimes \boldsymbol{I}_d \tag{C.65}$$

$$(\boldsymbol{W}_{n\times n}^{1/2}(\boldsymbol{Y}_{d\times n}) \otimes \boldsymbol{I}_d)^{\mathrm{T}}(\boldsymbol{W}_{n\times n}^{1/2}(\boldsymbol{Y}_{d\times n}) \otimes \boldsymbol{I}_d) = \boldsymbol{W}_{n\times n}(\boldsymbol{Y}_{d\times n}) \otimes \boldsymbol{I}_d \tag{C.66}$$

Expanding $f$ gives

$$f(\boldsymbol{\beta}_{d(p+1)}) = (\boldsymbol{\beta}_{d(p+1)}^{\mathrm{T}}(\widetilde{\boldsymbol{X}}_{n\times(p+1)} \otimes \boldsymbol{I}_d)^{\mathrm{T}} - \boldsymbol{y}_{dn}^{\mathrm{T}})(\boldsymbol{W}_{n\times n}(\boldsymbol{Y}_{d\times n}) \otimes \boldsymbol{I}_d)((\widetilde{\boldsymbol{X}}_{n\times(p+1)} \otimes \boldsymbol{I}_d)\boldsymbol{\beta}_{d(p+1)} - \boldsymbol{y}_{dn}) \tag{C.67}$$

Further expanding $f$ gives

$$f(\boldsymbol{\beta}_{d(p+1)}) = \boldsymbol{\beta}_{d(p+1)}^{\mathrm{T}}(\widetilde{\boldsymbol{X}}_{n\times(p+1)} \otimes \boldsymbol{I}_d)^{\mathrm{T}}(\boldsymbol{W}_{n\times n}(\boldsymbol{Y}_{d\times n}) \otimes \boldsymbol{I}_d)(\widetilde{\boldsymbol{X}}_{n\times(p+1)} \otimes \boldsymbol{I}_d)\boldsymbol{\beta}_{d(p+1)} - \boldsymbol{y}_{dn}^{\mathrm{T}}(\boldsymbol{W}_{n\times n}(\boldsymbol{Y}_{d\times n}) \otimes \boldsymbol{I}_d)(\widetilde{\boldsymbol{X}}_{n\times(p+1)} \otimes \boldsymbol{I}_d)\boldsymbol{\beta}_{d(p+1)} - \boldsymbol{\beta}_{d(p+1)}^{\mathrm{T}}(\widetilde{\boldsymbol{X}}_{n\times(p+1)} \otimes \boldsymbol{I}_d)^{\mathrm{T}}(\boldsymbol{W}_{n\times n}(\boldsymbol{Y}_{d\times n}) \otimes \boldsymbol{I}_d)\boldsymbol{y}_{dn} + \boldsymbol{y}_{dn}^{\mathrm{T}}(\boldsymbol{W}_{n\times n}(\boldsymbol{Y}_{d\times n}) \otimes \boldsymbol{I}_d)\boldsymbol{y}_{dn} \tag{C.68}$$

$\boldsymbol{\beta}_{d(p+1)}^{\mathrm{T}}(\widetilde{\boldsymbol{X}}_{n\times(p+1)} \otimes \boldsymbol{I}_d)^{\mathrm{T}}(\boldsymbol{W}_{n\times n}(\boldsymbol{Y}_{d\times n}) \otimes \boldsymbol{I}_d)\boldsymbol{y}_{dn}$ is a scalar and therefore equals its transpose:

$$\boldsymbol{\beta}_{d(p+1)}^{\mathrm{T}}(\widetilde{\boldsymbol{X}}_{n\times(p+1)} \otimes \boldsymbol{I}_d)^{\mathrm{T}}(\boldsymbol{W}_{n\times n}(\boldsymbol{Y}_{d\times n}) \otimes \boldsymbol{I}_d)\boldsymbol{y}_{dn} = (\boldsymbol{\beta}_{d(p+1)}^{\mathrm{T}}(\widetilde{\boldsymbol{X}}_{n\times(p+1)} \otimes \boldsymbol{I}_d)^{\mathrm{T}}(\boldsymbol{W}_{n\times n}(\boldsymbol{Y}_{d\times n}) \otimes \boldsymbol{I}_d)\boldsymbol{y}_{dn})^{\mathrm{T}} = \boldsymbol{y}_{dn}^{\mathrm{T}}(\boldsymbol{W}_{n\times n}(\boldsymbol{Y}_{d\times n}) \otimes \boldsymbol{I}_d)(\widetilde{\boldsymbol{X}}_{n\times(p+1)} \otimes \boldsymbol{I}_d)\boldsymbol{\beta}_{d(p+1)} \tag{C.69}$$

Thus:

$$f(\boldsymbol{\beta}_{d(p+1)}) = \boldsymbol{\beta}_{d(p+1)}^{\mathrm{T}}(\widetilde{\boldsymbol{X}}_{n\times(p+1)} \otimes \boldsymbol{I}_d)^{\mathrm{T}}(\boldsymbol{W}_{n\times n}(\boldsymbol{Y}_{d\times n}) \otimes \boldsymbol{I}_d)(\widetilde{\boldsymbol{X}}_{n\times(p+1)} \otimes \boldsymbol{I}_d)\boldsymbol{\beta}_{d(p+1)} - 2\boldsymbol{y}_{dn}^{\mathrm{T}}(\boldsymbol{W}_{n\times n}(\boldsymbol{Y}_{d\times n}) \otimes \boldsymbol{I}_d)(\widetilde{\boldsymbol{X}}_{n\times(p+1)} \otimes \boldsymbol{I}_d)\boldsymbol{\beta}_{d(p+1)} + \boldsymbol{y}_{dn}^{\mathrm{T}}(\boldsymbol{W}_{n\times n}(\boldsymbol{Y}_{d\times n}) \otimes \boldsymbol{I}_d)\boldsymbol{y}_{dn} \tag{C.70}$$

The gradient of $f$ with respect to $\boldsymbol{\beta}_{d(p+1)}$ is (applying eqs. (A.26) and (A.27))

$$\frac{\partial f(\boldsymbol{\beta}_{d(p+1)})}{\partial \boldsymbol{\beta}_{d(p+1)}} = 2(\boldsymbol{\beta}_{d(p+1)}^{\mathrm{T}}((\widetilde{\boldsymbol{X}}_{n\times(p+1)} \otimes \boldsymbol{I}_d)^{\mathrm{T}}(\boldsymbol{W}_{n\times n}(\boldsymbol{Y}_{d\times n}) \otimes \boldsymbol{I}_d)(\widetilde{\boldsymbol{X}}_{n\times(p+1)} \otimes \boldsymbol{I}_d)) - \boldsymbol{y}_{dn}^{\mathrm{T}}(\boldsymbol{W}_{n\times n}(\boldsymbol{Y}_{d\times n}) \otimes \boldsymbol{I}_d)(\widetilde{\boldsymbol{X}}_{n\times(p+1)} \otimes \boldsymbol{I}_d)) \tag{C.71}$$

Applying eqs. (A.16) and (A.24), the gradient vector becomes

$$\nabla_{\boldsymbol{\beta}_{d(p+1)}} f(\boldsymbol{\beta}_{d(p+1)}) = \left(\frac{\partial f(\boldsymbol{\beta}_{d(p+1)})}{\partial \boldsymbol{\beta}_{d(p+1)}}\right)^{\mathrm{T}} = 2(((\widetilde{\boldsymbol{X}}_{n\times(p+1)} \otimes \boldsymbol{I}_d)^{\mathrm{T}}(\boldsymbol{W}_{n\times n}(\boldsymbol{Y}_{d\times n}) \otimes$$

$$\boldsymbol{I}_d)(\widetilde{\boldsymbol{X}}_{n\times(p+1)} \otimes \boldsymbol{I}_d))\boldsymbol{\beta}_{d(p+1)} - (\widetilde{\boldsymbol{X}}_{n\times(p+1)} \otimes \boldsymbol{I}_d)^{\mathrm{T}}(\boldsymbol{W}_{n\times n}(\boldsymbol{Y}_{d\times n}) \otimes \boldsymbol{I}_d)\boldsymbol{y}_{dn}) \tag{C.72}$$

Setting the gradient to $\boldsymbol{0}_{d(p+1)}$:

$$((\widetilde{\boldsymbol{X}}_{n\times(p+1)} \otimes \boldsymbol{I}_d)^{\mathrm{T}}(\boldsymbol{W}_{n\times n}(\boldsymbol{Y}_{d\times n}) \otimes \boldsymbol{I}_d)(\widetilde{\boldsymbol{X}}_{n\times(p+1)} \otimes \boldsymbol{I}_d))\boldsymbol{\beta}_{d(p+1)} = (\widetilde{\boldsymbol{X}}_{n\times(p+1)} \otimes \boldsymbol{I}_d)^{\mathrm{T}}(\boldsymbol{W}_{n\times n}(\boldsymbol{Y}_{d\times n}) \otimes \boldsymbol{I}_d)\boldsymbol{y}_{dn} \tag{C.73}$$

By eqs. (A.16) and (A.17),

$$((\widetilde{\boldsymbol{X}}_{n\times(p+1)} \otimes \boldsymbol{I}_d)^{\mathrm{T}}(\boldsymbol{W}_{n\times n}(\boldsymbol{Y}_{d\times n}) \otimes \boldsymbol{I}_d)(\widetilde{\boldsymbol{X}}_{n\times(p+1)} \otimes \boldsymbol{I}_d)) = (\widetilde{\boldsymbol{X}}^{\mathrm{T}}_{n\times(p+1)}\boldsymbol{W}_{n\times n}(\boldsymbol{Y}_{d\times n})\widetilde{\boldsymbol{X}}_{n\times(p+1)}) \otimes \boldsymbol{I}_d \tag{C.74}$$

$$(\widetilde{\boldsymbol{X}}_{n\times(p+1)} \otimes \boldsymbol{I}_d)^{\mathrm{T}}(\boldsymbol{W}_{n\times n}(\boldsymbol{Y}_{d\times n}) \otimes \boldsymbol{I}_d) = (\widetilde{\boldsymbol{X}}^{\mathrm{T}}_{n\times(p+1)}\boldsymbol{W}_{n\times n}(\boldsymbol{Y}_{d\times n})) \otimes \boldsymbol{I}_d \tag{C.75}$$

$\widetilde{\boldsymbol{X}}^{\mathrm{T}}_{n\times(p+1)}\boldsymbol{W}_{n\times n}(\boldsymbol{Y}_{d\times n})\widetilde{\boldsymbol{X}}_{n\times(p+1)}$ is invertible. Applying eq. (A.18) to invert the Kronecker product $(\widetilde{\boldsymbol{X}}^{\mathrm{T}}_{n\times(p+1)}\boldsymbol{W}_{n\times n}(\boldsymbol{Y}_{d\times n})\widetilde{\boldsymbol{X}}_{n\times(p+1)}) \otimes \boldsymbol{I}_d$:

$$\boldsymbol{\beta}_{d(p+1)} = ((\widetilde{\boldsymbol{X}}^{\mathrm{T}}_{n\times(p+1)}\boldsymbol{W}_{n\times n}(\boldsymbol{Y}_{d\times n})\widetilde{\boldsymbol{X}}_{n\times(p+1)})^{-1} \otimes \boldsymbol{I}_d)((\widetilde{\boldsymbol{X}}^{\mathrm{T}}_{n\times(p+1)}\boldsymbol{W}_{n\times n}(\boldsymbol{Y}_{d\times n})) \otimes \boldsymbol{I}_d)\boldsymbol{y}_{dn} \tag{C.76}$$

Simplifying the Kronecker products with eq. (A.17) gives

$$\boldsymbol{\beta}_{d(p+1)} = (((\widetilde{\boldsymbol{X}}^{\mathrm{T}}_{n\times(p+1)}\boldsymbol{W}_{n\times n}(\boldsymbol{Y}_{d\times n})\widetilde{\boldsymbol{X}}_{n\times(p+1)})^{-1}\widetilde{\boldsymbol{X}}^{\mathrm{T}}_{n\times(p+1)}\boldsymbol{W}_{n\times n}(\boldsymbol{Y}_{d\times n})) \otimes \boldsymbol{I}_d)\boldsymbol{y}_{dn} = ((\boldsymbol{W}_{n\times n}(\boldsymbol{Y}_{d\times n})\widetilde{\boldsymbol{X}}_{n\times(p+1)}(\widetilde{\boldsymbol{X}}^{\mathrm{T}}_{n\times(p+1)}\boldsymbol{W}_{n\times n}(\boldsymbol{Y}_{d\times n})\widetilde{\boldsymbol{X}}_{n\times(p+1)})^{-1})^{\mathrm{T}} \otimes \boldsymbol{I}_d)\mathrm{vec}(\boldsymbol{Y}_{d\times n}) \tag{C.77}$$

By eq. (A.20):

$$\boldsymbol{\beta}_{d(p+1)} = \mathrm{vec}(\boldsymbol{I}_d\boldsymbol{Y}_{d\times n}\boldsymbol{W}_{n\times n}(\boldsymbol{Y}_{d\times n})\widetilde{\boldsymbol{X}}_{n\times(p+1)}(\widetilde{\boldsymbol{X}}^{\mathrm{T}}_{n\times(p+1)}\boldsymbol{W}_{n\times n}(\boldsymbol{Y}_{d\times n})\widetilde{\boldsymbol{X}}_{n\times(p+1)})^{-1}) \tag{C.78}$$

Applying eq. (C.61) and because the vectorization operator is bijective, it follows that

$$\widehat{\boldsymbol{\theta}}_{d\times(p+1)}(\boldsymbol{X}_{p\times n}, \boldsymbol{Y}_{d\times n}) = \boldsymbol{Y}_{d\times n}\boldsymbol{W}_{n\times n}(\boldsymbol{Y}_{d\times n})\widetilde{\boldsymbol{X}}_{n\times(p+1)}(\widetilde{\boldsymbol{X}}^{\mathrm{T}}_{n\times(p+1)}\boldsymbol{W}_{n\times n}(\boldsymbol{Y}_{d\times n})\widetilde{\boldsymbol{X}}_{n\times(p+1)})^{-1} \tag{C.79}$$

which is exactly eq. (4.26).

To verify that this critical point is a minimum, we examine the second derivative. The Hessian matrix of $f(\boldsymbol{\beta}_{d(p+1)})$ (using eq. (A.28)) is

$$\boldsymbol{H}_{\boldsymbol{\beta}_{d(p+1)}} f(\boldsymbol{\beta}_{d(p+1)}) = \frac{\partial \nabla_{\boldsymbol{\beta}_{d(p+1)}} f(\boldsymbol{\beta}_{d(p+1)})}{\partial \boldsymbol{\beta}_{d(p+1)}} = 2((\widetilde{\boldsymbol{X}}^{\mathrm{T}}_{n\times(p+1)}\boldsymbol{W}_{n\times n}(\boldsymbol{Y}_{d\times n})\widetilde{\boldsymbol{X}}_{n\times(p+1)}) \otimes \boldsymbol{I}_d) \tag{C.80}$$

The matrices $\boldsymbol{I}_d$ and $\widetilde{\boldsymbol{X}}^{\mathrm{T}}_{n\times(p+1)}\boldsymbol{W}_{n\times n}(\boldsymbol{Y}_{d\times n})\widetilde{\boldsymbol{X}}_{n\times(p+1)}$ are symmetric and positive definite.

Therefore, their eigenvalues are positive (Gentle 2024, p.186). Their Kronecker product is a symmetric matrix (Gentle 2024, p.118). The eigenvalues of the Kronecker product are all products of eigenvalues of the individual matrices, hence positive (Horn and Johnson 1991, p.245). Consequently, $\boldsymbol{H}$ is symmetric with all eigenvalues positive, therefore, it is positive definite for all $\boldsymbol{\beta}$ (Gentle 2024, p.186), confirming that $f(\boldsymbol{\beta}_{d(p+1)})$ is strictly convex (Boyd and Vandenberghe 2004, p.71). Therefore, the critical point is the unique global minimizer (Boyd and Vandenberghe 2004, p.69). This confirms that the estimate for the multi-dimensional linear regression model is given by eq. (4.26).■

## Appendix D Concise equation reference

We present in Table D.1 a structured summary of the key equations presented in the manuscript, organized to facilitate quick reference and comparison. It covers the three principal modeling frameworks (one-dimensional, multi-dimensional, and Nash-Sutcliffe) in both the $d \times n$ (Sections 3 and 4) and $n \times d$ (Section 5) data settings.

Table D.1. Summary of key formulas for the one-dimensional, multi-dimensional, and Nash–Sutcliffe cases in both $d \times n$ and $n \times d$ settings.

| Category ($d \times n$ setting) | One-dimensional case | Multi-dimensional case | Nash-Sutcliffe case |
|---|---|---|---|
| Functional | (3.24) | (3.29) | (4.3) |
| Loss function | (1.2) | (1.8), (3.32) | (4.2) |
| Identification function | (3.26) | (3.31) | (4.4) |
| Climatology | (3.28) | (3.36) | (4.6), (4.7) |
| Realized loss | (1.1) | (3.33), (3.34) | (1.10), (4.1) |
| Empirical identification | (3.27) | (3.35) | (4.5) |
| Linear regression model | (4.14) | (4.19) | (4.24) |
| Regression parameter estimate | (4.16) | (4.21) | (4.26) |
| Linear predictive model | (4.18) | (4.23) | (4.29) |
| Category ($n \times d$ setting) | | | |
| Functional | (3.24) | (3.29) | (4.3) |
| Loss function | (1.2) | (1.8), (3.32) | (4.2) |
| Identification function | (3.26) | (3.31) | (4.4) |
| Climatology | (3.28) | (5.4) | (5.6), (5.7) |
| Realized loss | (1.1) | (5.1) | (5.2) |
| Empirical identification | (3.27) | (5.3) | (5.5) |
| Linear regression model | (5.10) | (5.15) | (5.20) |
| Regression parameter estimate | (5.12) | (5.17) | (5.22) |
| Linear predictive model | (5.14) | (5.19) | (5.25) |

## Appendix E Statistical software

All analyses were conducted using the `R` programming language, version 4.5.2 (`R` Core Team 2025) within the `RStudio` integrated development environment (version

2026.01.0+392).

Data processing and visualization were performed using the `data.table` (Barrett et al. 2026) and `tidyverse` (Wickham et al. 2019; Wickham 2023) packages.

Simulations were performed using the `MASS` (Ripley and Venables 2025; Venables and Ripley 2002) and `simstudy` (Goldfeld and Wujciak-Jens 2020; 2025) packages.

The required data were retrieved from the `airGRdatasets` package (Delaigue et al. 2025).

Finally, all reports were generated using the `devtools` (Wickham et al. 2025), `knitr` (Xie 2014; 2015; 2025) and `rmarkdown` (Allaire et al. 2025; Xie et al. 2018; 2020) packages.

**Supplementary material**: The supplementary files for this paper are available at https://doi.org/10.5281/zenodo.21168825 and support full reproducibility of the reported results.

Section 6.1: `01_Simulation_example_01.Rmd`, `01_Simulation_example_01.html`, `01_Simulation_example_01.RData`, `01_Figures.Rmd`, `01_Figures.html`.

Section 6.2: `02_Simulation_example_02.Rmd`, `02_Simulation_example_02.html`, `02_Simulation_example_02.RData`, `02_Figures.Rmd`, `02_Figures.html`.

Section 6.3: `03_Simulation_example_03.Rmd`, `03_Simulation_example_03.html`.

Section 6.4: `04_data.Rmd`, `04_data.html`, `04_data.RData`.

Section 6.4.1: `05a_Streamflow_real_data_regression.Rmd`, `05a_Streamflow_real_data_regression.html`, `05a_Streamflow_real_data_regression.RData`, `05a_Figures.Rmd`, `05a_Figures.html`.

Section 6.4.2: `05b_Temperature_real_data_regression.Rmd`, `05b_Temperature_real_data_regression.html`, `05b_Temperature_real_data_regression.RData`, `05b_Figures.Rmd`, `05b_Figures.html`.

Functions for computing climatologies, loss functions, realized loss functions and regression models (as listed in Table D.1) are available in the `R_functions.R` file.

**Conflicts of interest:** The authors declare no conflict of interest.

**Statement:** During the preparation of this work, the authors used `DeepSeek-V.3.2` to assist with language polishing and to enhance readability. After using this service, the authors reviewed and edited the content as needed and take full responsibility for the content of the published article.